%% file: main.tex
\documentclass[11pt]{article}

\usepackage[noend]{algpseudocode}
\usepackage[linesnumbered,noend, ruled]{algorithm2e}
\usepackage[letterpaper, left=1in, right=1in, top=1in,bottom=1in]{geometry}
\usepackage[round]{natbib}

\usepackage{parskip,verbatim}

\usepackage{booktabs} % For formal tables

\usepackage[utf8]{inputenc}

\usepackage{geometry}

\usepackage{tikz}
\usetikzlibrary{automata, positioning}
\usepackage{makecell}
\usepackage{graphpap,amscd,mathrsfs,graphicx,lscape,enumitem,dsfont,bm,url,subfigure}
\usepackage{epsfig,amstext,xspace}

\usepackage{thmtools,thm-restate}

\usepackage{auxiliary}
\usepackage[utf8]{inputenc} % allow utf-8 input
\usepackage[T1]{fontenc}    % use 8-bit T1 fonts
\usepackage{hyperref}       % hyperlinks
\usepackage{url}            % simple URL typesetting
\usepackage{booktabs}       % professional-quality tables
\usepackage{amsfonts}       % blackboard math symbols
\usepackage{nicefrac}       % compact symbols for 1/2, etc.
\usepackage{microtype}      % microtypography
\usepackage{bbm}
\usepackage{cleveref}
\usepackage[misc]{ifsym}
\usepackage{pifont}

\usepackage{bbm}
\usepackage{amsthm}
\usepackage{amsfonts,amssymb}
\usepackage{amsmath}
\newtheorem{assumption}{Assumption}
\newtheorem{theorem}{Theorem}
\newtheorem{proposition}{Proposition}
\newtheorem{discussion}{Discussion}
\newtheorem{definition}{Definition}

\newtheorem{lemma}{Lemma}
\newtheorem{remark}{Remark}
\newtheorem{question}{Question}
\newtheorem{example}{Example}
\usepackage{paralist}
\usepackage{color}
\usepackage{subfigure}
\usepackage{hyperref}
\usepackage{url}
\usepackage{makecell}
\usepackage{booktabs}
\usepackage{caption}
\usepackage{colortbl}
\usepackage{xcolor}
\usepackage{multirow}
\usepackage{pifont}
\usepackage{bm}
\usepackage{tikz}
\usetikzlibrary{automata, positioning}
\usepackage{extarrows}
\usepackage{sidecap}
\usepackage{framed}
\usepackage{makecell}
\usepackage{tikz}

\newcommand{\Ex}{\mathbb{E}}

\usepackage{color}              % Need the color package
\usepackage[suppress]{color-edits}
\addauthor{tl}{cyan}
\addauthor{ws}{red}
\addauthor{ms}{purple}
\addauthor{as}{brown}

\usepackage{authblk}

\begin{document}
\title{Sample Complexity of Policy Gradient Finding Second-Order Stationary Points}

 \author[1]{Long Yang}
\author[2]{Qian Zheng}
\author[1]{Gang Pan}
\affil[1]{Zhejiang University}
\affil[2]{Nanyang Technological University}
\affil[ ]{\textsuperscript{1}\texttt{\{yanglong,gpan\}@zju.edu.cn};\quad \textsuperscript{2}\texttt{zhengqian@ntu.edu.sg}}
\date{\today}
%\date{November 2019}
\maketitle
\begin{abstract}
\input{abstract}
\end{abstract}

\input{background}

\input{direct_policy}

\input{related-work}

\clearpage
\appendix
\input{table}

\input{app-lemma}

\input{proof-pro-2}
\input{proof-pro-1}

\input{theorem_1_proof}

\clearpage
\bibliographystyle{plainnat}
\bibliography{reference}

\end{document}

%% file: abstract.tex
%!TEX root = main.tex
The goal of policy-based reinforcement learning (RL) is to search the maximal point of its objective.
However, due to the inherent non-concavity of its objective, convergence to a first-order stationary point (FOSP) can not guarantee the policy gradient methods finding a maximal point.
A FOSP can be a minimal or even a saddle point, which is undesirable for RL.
Fortunately, if all the saddle points are \emph{strict}, all the second-order stationary points (SOSP) are exactly equivalent to local maxima.
Instead of FOSP, we consider SOSP as the convergence criteria to character the sample complexity of policy gradient.
Our result shows that policy gradient converges to an $(\epsilon,\sqrt{\epsilon\chi})$-SOSP with probability at least $1-\widetilde{\mathcal{O}}(\delta)$
after the total cost of 
$\mathcal{O}\left(\dfrac{\epsilon^{-\frac{9}{2}}}{(1-\gamma)\sqrt\chi}\log\dfrac{1}{\delta}\right)$, where $\gamma\in(0,1)$.
Our result improves the state-of-the-art result significantly where it requires $\mathcal{O}\left(\dfrac{\epsilon^{-9}\chi^{\frac{3}{2}}}{\delta}\log\dfrac{1}{\epsilon\chi}\right)$.
Our analysis is based on the key idea that decomposes the parameter space $\mathbb{R}^p$ into three non-intersected regions: non-stationary point, saddle point, and local optimal region, then making a local improvement of the objective of RL in each region.
This technique can be potentially generalized to extensive policy gradient methods.

%% file: background.tex
\section{Introduction}
\label{sec:intro}

Policy gradient method \citep{williams1992simple,sutton2000policy} is widely used to search the optimal policy in modern reinforcement learning (RL).
Such method (or its variants) searches over a differentiable parameterized class of polices by performing a stochastic gradient on a cumulative expected reward function.
Due to its merits such as the simplicity of implementation in the simulated environment;
it requires low memory; it can be applied to any differentiable parameterized classes \citep{agarwal2019optimality},
policy gradient method has achieved significant successes in challenging fields such as robotics \citep{deisenroth2013survey,duan2016benchmarking}, playing Go \citep{silver2016mastering,silver2017mastering}, neural architecture search \citep{zoph2017neural},
NLP \citep{kurita2019multi,whiteson2019survey}, computer vision \citep{sarmad2019rl},
and recommendation system \citep{pan2019policy}.

Despite it has tremendous successful applications, suffering from high sample complexity is still a critical challenge for the policy gradient \citep{haarnoja2018soft,lee2019sample,xu2020sample}.
Thus, for policy gradient, theory analysis of its sample complexity plays an important role in RL since the sample complexity not only provides an understanding of the policy gradient but also gives insights on how to improve the sample efficiency of the existing RL algorithms.

Investigation of the sample complexity of policy gradient algorithm (or its variant) can be traced back to the pioneer works of \cite{kearns2000approximate,kakade2003sample}.
Recently, to improve sample efficiency, \cite{Matteo2018Stochastic,shen2019hessian,xu2020sample} introduce stochastic variance reduced gradient techniques \citep{johnson2013accelerating,nguyen2017sarah} to policy optimization, and they have studied the sample complexity of policy gradient methods to achieve a first-order stationary point (FOSP) (i.e., $\theta$ such that $\|\nabla J(\theta)\|_2\leq \epsilon$).
However, since the objective of RL is a non-concave function with respect to the standard policy parameterizations \citep{Matteo2018Stochastic,agarwal2019optimality}, 
a FOSP could be a maximal point, a minimal point, and even a saddle point.
Both minimal points and saddle points are undesirable for policy gradient since its goal is to search a maximal point, 
which implies within the numbers of samples provided by \cite{Matteo2018Stochastic,shen2019hessian,xu2020sample}, 
we can not guarantee the output of their policy gradient algorithm is a maximal point.
This motivates a fundamental question as follows,
\begin{question}
\label{def:question}
How many samples does an agent need to collect to guarantee the policy gradient methods converge to a maximal point certainly?
\end{question}
\subsection{Our Work}
In this paper, we consider the second-order stationary point (SOSP) to answer Question \ref{def:question}.
More specifically, inspired by the previous works from non-convex optimization \citep{jin2017escape,daneshmand2018escaping}, we investigate the sample complexity of policy gradient methods finding an $(\epsilon,\sqrt{\epsilon\chi})$-SOSP, see Definition \ref{def:sosp}, i.e., the convergent point $\theta$ satisfies
\[\|\nabla J(\theta)\|_2 \leq \epsilon,~~\text{and}~~\lambda_{\max}(\nabla^{2} J(\theta))\leq \sqrt{\chi\epsilon}.\]
The criterion of $(\epsilon,\sqrt{\epsilon\chi})$-SOSP requires the convergent point with a small gradient and with almost a negative semi-definite Hessian matrix.
This criterion not only ensures a convergent point is a FOSP but also rules out both saddle points (whose Hessian are indefinite) and minimal points (whose Hessian are positive definite).
Therefore, convergence to a $(\epsilon,\sqrt{\epsilon\chi})$-SOSP guarantees the policy gradient methods converge to a local maximal point clearly.
Our result shows that within a cost of 
\[
\mathcal{O}\left(
\dfrac{\epsilon^{-\frac{9}{2}}}{(1-\gamma)\sqrt\chi}\log\dfrac{1}{\delta}
\right)=
\widetilde{\mathcal{O}}\left(\epsilon^{-\frac{9}{2}}\right)
,\]
policy gradient converges to an $(\epsilon,\sqrt{\epsilon\chi})$-SOSP 
with probability at least $1-\widetilde{\mathcal{O}}(\delta)$.
Our result improves the state-of-the-art result of \citep{zhang2019global}  significantly, where they require $\widetilde{\mathcal{O}}(\epsilon^{-9})$ samples to achieve an $(\epsilon,\sqrt{\epsilon\chi})$-SOSP.

Notably, we provide a novel analysis that can be potentially generalized to extensive policy gradient methods.
Concretely, we decompose the parameter space $\mathbb{R}^p$ into three different regions: non-stationary point, saddle point, and local optimal region, then making a local improvement in each region.
The main challenge occurs on the saddle point region, where we utilize a technique called correlated negative curvature (CNC) \citep{daneshmand2018escaping} to make a local improvement.

\subsection{Paper Organization}
In Section \ref{Background and Notations}, we introduce some necessary conceptions of policy gradient and some standard assumptions in policy optimization.
In Section \ref{sec:assumption}, we formally define $(\epsilon,\sqrt{\epsilon\chi})$-SOSP.
Our main contribution lies in Section \ref{sec:mian-result}, where we provide the main result that presents the sample complexity of policy gradient finding an $(\epsilon,\sqrt{\epsilon\chi})$-SOSP, and we provide an overview of the proof technique.
Related works and future works are discussed in Section \ref{sec:Related Work}.

\subsection{Notations} 
Let $\|\cdot\|_{2}$ be the Euclidean norm of a vector in $\mathbb{R}^{p}$.
For a symmetric matrix $A$, we use $\lambda_{\min}(A)$ and $\lambda_{\max}(A)$ as its minimum and maximum eigenvalue correspondingly. 
Let $\|A\|_{op}$ denote the operator norm of the matrix $A$; furthermore, according to \cite{van1983matrix}, if $A\in\mathbb{R}^{p\times p}$ is a symmetric matrix, then $\|A\|_{op}=\max_{1\leq i\leq p}\{|\lambda_i|\}$, where $\{\lambda_i\}_{i=1}^{p}$ is the set of the eigenvalues of $A$.
We use $A\succ 0$ to denote a positive definite matrix $A$.
For a function $J(\cdot):\mathbb{R}^{p}\rightarrow \mathbb{R}$, let $\nabla J$ and $\nabla^{2}J$ denote its gradient vector and Hessian matrix correspondingly. 
Let $\mathbb{B}_{2}(o,r)$ be a $p$-dimensional $\ell_{2}$ ball with the centre $o$ and radius $r$, i.e., $\mathbb{B}_{2}(o,r)=\{x\in\mathbb{R}^{p};\|x-o\|_2\leq r\}$.
For any real number $x$, $\lceil x \rceil $and $\lfloor x\rfloor$ denote the nearest integer to $x$ from above and below.
We use $\widetilde{\mathcal{O}}$ to hide polylogarithmic factors in the input parameters, i.e., $\widetilde{\mathcal{O}}(f(x))=\mathcal{O}(f(x)\log(f(x))^{\mathcal{O}(1)})$.

\section{Policy Gradient Methods and Some Standard Assumptions}
\label{Background and Notations}
In this section, we introduce some necessary concepts of reinforcement learning, policy gradient and some standard assumptions in policy optimization.

\subsection{Reinforcement Learning}
Reinforcement learning \citep{sutton2018reinforcement} is often formulated as \emph{Markov decision processes} (MDP) $\mathcal{M}=(\mathcal{S},\mathcal{A},{P},{R},\rho_0,\gamma)$, 
where $\mathcal{S}$ is the state space, $\mathcal{A}$ is the action space;
$P(s^{'}|s,a)$ is the probability of state transition from $s$ to $s^{'}$ under playing the action $a$;
$R(\cdot,\cdot):\mathcal{S}\times\mathcal{A}\rightarrow [R_{\min},R_{\max}]$ is a bounded reward function, where $R_{\min},R_{\max}$ two positive scalars.
$\rho_{0}(\cdot):\mathcal{S}\rightarrow[0,1]$ is the initial state distribution and the discount factor $\gamma\in(0,1)$. 

The parametric \emph{policy} $\pi_{\theta}$ is a probability distribution over $\mathcal{S}\times\mathcal{A}$ with a parameter $\theta\in\mathbb{R}^p$, and we use $\pi_{\theta}(a|s)$ to denote the probability of playing $a$ in state $s$.
%In this paper, we mainly consider 
%two different policy classes: \emph{direct} and \emph{softmax} parameterization, which are
%\emph{complete} in the sense that any stochastic policy can be
%represented in this class \citep{agarwal2019optimality}. 
%The \emph{direct}  policies are parameterized by
%\begin{equation}
%    \label{eq:direct}
%    \pi_\theta (a| s) = \theta_{s,a}, ~\forall~(s,a)\in\mathcal{S}\times\mathcal{A}
%    \end{equation}
%where $\theta_{s,a} \geq 0$, $\sum_{a \in\mathcal{A}} \theta_{s,a} = 1$.
%the \emph{softmax} parameterization:
% \begin{equation}
%    \pi_\theta(a| s) = \frac{\exp(\theta_{s,a})}{\sum_{a'\in\mathcal{A}} \exp(\theta_{s,a'})},
%    \label{eq:softmax}
% \end{equation}
%where, $\theta_{s,a}=[\Theta]_{s,a}\in\mathbb{R}^{|\mathcal{S}|\times|\mathcal{A|}}$.
Let $\tau=\{s_{t}, a_{t}, r_{t+1}\}_{t\ge0}\sim\pi_{\theta}$ be a trajectory generated by the policy $\pi_{\theta}$, 
where $s_{0}\sim\rho_{0}(\cdot)$, $a_{t}\sim\pi_{\theta}(\cdot|s_{t})$, $r_{t+1}=R(s_t,a_t)$ and $s_{t+1}\sim P(\cdot|s_{t},a_{t})$. 
The \emph{state value function} of $\pi_\theta$ is defined as follows,
\[V^{\pi_{\theta}}(s) = \mathbb{E}_{\pi_{\theta}}\left[\sum_{t=0}^{\infty}\gamma^{t}r_{t+1}|s_{0} = s\right],\]
where $\mathbb{E}_{\pi_{\theta}}[\cdot|\cdot]$ denotes a conditional expectation on actions which are selected according to the policy $\pi_{\theta}$.
The \emph{advantage function} of the policy $\pi_{\theta}$ is defined as follows,
\[A^{\pi_{\theta}}(s,a)=Q^{\pi_{\theta}}(s,a)-V^{\pi_{\theta}}(s),\]
where
$Q^{\pi_{\theta}}(s,a)$ is the \emph{state-action value function}: \[Q^{\pi_{\theta}}(s,a) = \mathbb{E}_{\pi_{\theta}}\left[\sum_{t=0}^{\infty}\gamma^{t}r_{t+1}|s_{0} = s,a_{0}=a\right].\]
We use $P^{\pi_{\theta}}(s_t=s|s_0)$ to denote the probability of visiting the state $s$ after $t$
time steps from the initial state $s_0$ by executing $\pi_{\theta}$,
and \[d^{\pi_{\theta}}_{s_0}(s)=\sum_{t=0}^{\infty}\gamma^{t}P^{\pi_{\theta}}(s_t=s|s_0)\] is the (unnormalized) discounted stationary state distribution of the Markov chain (starting at $s_0$) induced by $\pi_{\theta}$.
Furthermore, since $s_0\sim\rho_{0}(\cdot)$, we define
\[
d^{\pi_{\theta}}_{\rho_0}(s)=\mathbb{E}_{s_0\sim\rho_{0}(\cdot)}[d^{\pi_{\theta}}_{s_0}(s)]
\]
as the discounted state visitation distribution over the initial distribution $\rho_0$.
Recall $\tau=\{s_{t}, a_{t}, r_{t+1}\}_{t\ge0}\sim\pi_\theta$,
we define
\[J(\pi_\theta|s_0)=\mathbb{E}_{\tau\sim\pi_{\theta},s_0\sim\rho_{0}(\cdot)}[R(\tau)]=\mathbb{E}_{s\sim d^{\pi_{\theta}}_{s_0}(\cdot),a\sim\pi_{\theta}(\cdot|s)}[R(s,a)],\]
where $R(\tau)=\sum_{t\ge0}\gamma^{t}r_{t+1}$, and $J(\pi_\theta|s_0)$ is ``conditional'' on $s_0$ since we emphasize the trajectory $\tau$ starting from $s_0$.
Furthermore,
we define the expected return $ J(\theta)=:\mathbb{E}_{s_0\sim\rho_{0}(
        \cdot)}[J(\pi_\theta|s_0)]$ as follows,
\begin{flalign}
\label{J-objectiove}
    J(\theta)=\mathbb{E}_{s\sim d^{\pi_{\theta}}_{\rho_0}(\cdot),a\sim\pi_{\theta}(\cdot|s)}[R(s,a)]=\int_{s\in\mathcal{S}} d_{\rho_0}^{\pi_{\theta}}(s)\int_{a\in\mathcal{A}}\pi_{\theta}(a|s) R(s,a)\text{d}a\text{d}s.
\end{flalign}
The goal of policy-based reinforcement learning is to solve the following policy optimization problem:
\begin{flalign}
\label{Eq:thata-optimal}
    \max_{\theta\in\mathbb{R}^{p}} J(\theta).
\end{flalign}
\subsection{Policy Gradient Methods}
The basic idea of policy gradient \citep{williams1992simple,sutton2000policy} is to update the parameter according to the direction with respect to the gradient of $J(\theta)$, i.e.,
\begin{flalign}
\label{update-policy-gradient}
\theta_{k+1}=\theta_{k}+\alpha \widehat{\nabla J(\theta_{k})},
\end{flalign}
where $\alpha>0$ is step-size, $\widehat{\nabla J(\theta_k)}$ is a stochastic estimator of policy gradient $\nabla J(\theta_k)$.
According to \cite{sutton2000policy}, we present the well-known \emph{policy gradient theorem} as follows,
  \begin{flalign}
    \nonumber
     \nabla J(\theta)=\int_{s\in\mathcal{S}} d_{\rho_0}^{\pi_{\theta}}(s)\int_{a\in\mathcal{A}}Q^{\pi_{\theta}}(s,a)\nabla \pi_{\theta}(s,a) \text{d}a\text{d}s=\mathbb{E}_{s\sim d^{\pi_{\theta}}_{\rho_0}(\cdot),a\sim\pi_{\theta}(\cdot|s)}\left[Q^{\pi_{\theta}}(s,a)\nabla \log\pi_{\theta}(a|s)\right],
 \end{flalign}
which provides a possible way to find the estimator of $\nabla J(\theta)$. 
One issue that we should address is how to estimate $Q^{\pi_{\theta}}(s, a)$ appears in the policy gradient theorem. 
A simple approach is to use a sample return $R(\tau)$ to estimate $Q^{\pi_{\theta}}(s, a) $, i.e., we calculate the policy gradient estimator as follows,
\begin{flalign}
\label{g-tau}
g(\tau|\theta)=\sum_{t\ge0}\nabla\log\pi_{\theta}(a_{t}|s_{t})R(\tau).
\end{flalign}
Replace $\widehat{\nabla J(\tau|\theta_k)}$ of (\ref{update-policy-gradient}) with $g(\tau|\theta_k)$, we achieve the update rule of $\mathtt{REINFORCE}$ \citep{williams1992simple}:
\begin{flalign}
\label{vpg}
\theta_{k+1}=\theta_{k}+\alpha g(\tau|\theta_k).
\end{flalign}
\subsection{Fisher Information Matrix}
For the policy optimization (\ref{J-objectiove}), we learn the parameter from the samples that come from an unknown probability distribution. 
Fisher information matrix \citep{fisher1920amath,kakade2002natural,ly2017tutorial} provides the information that a sample of data provides about the unknown parameter.
According to \cite{kakade2002natural,bhatnagar2008incremental},
the Fisher information matrix $F(\theta)$ is positive definite, i.e., there exists a constant $\omega>0$ s.t.,
\begin{flalign}
\label{fisher-information}
F(\theta)=:
\int_{s\in\mathcal{S}}d_{\rho_0}^{\pi_{\theta}}(s)\int_{a\in\mathcal{A}}\nabla\log \pi_{\theta}(a|s) [\nabla\log \pi_{\theta}(a|s)]^{\top}\text{d}s\text{d}a\succ \omega I_p,~~\forall~\theta\in\mathbb{R}^{p},
\end{flalign}
where $I_p\in\mathbb{R}^{p\times p}$ is the identity matrix.

%% file: direct_policy.tex
\subsection{Standard Assumptions}
\begin{assumption}
	\label{ass:On-policy-derivative}
	For each pair $(s, a)\in\mathcal{S}\times\mathcal{A}$, for any $\theta\in\mathbb{R}^p$, and all components $i$, $j$, there exists positive two constants $0\leq G,L,U<\infty$ such that 
	\begin{flalign}
	\label{def:F-G}
	\emph{\textbf{(a)}}:\big|\nabla_{\theta_i}\log\pi_{\theta}(a|s)\big|\leq G;~~\emph{\textbf{(b)}}:\Big|\frac{\partial^{2}}{\partial\theta_i\partial\theta_j}\log\pi_{\theta}(a|s)\Big|\leq L; 
	~~\emph{\textbf{(c)}}: \big|\nabla_{\theta_i}\pi_{\theta}(a|s)\big|\leq U.
	\end{flalign}
\end{assumption}
Assumptions \ref{ass:On-policy-derivative} is a standard condition in policy optimization, and it has be applied to several recent policy gradient literatures
\citep{castro2010convergent,pirotta2015policy,Matteo2018Stochastic,shen2019hessian,xu2020sample}.
Assumption \ref{ass:On-policy-derivative} is reasonable a condition since the widely used policy classes such as Gaussian, softmax \citep{konda1999actor}, and relative entropy policy \citep{peters2010relative} all satisfy (\ref{def:F-G}). Recently, \citet{zhang2019global,papini2019smoothing,wang-yue-finite-uai2020} have provided the details to check above policies satisfy Assumptions \ref{ass:On-policy-derivative}.

According to the Lemma B.2 of \citep{Matteo2018Stochastic}, 
Assumption \ref{ass:On-policy-derivative} implies the expected return $J(\theta)$ is $\ell$-Lipschitz smooth, i.e., for any $\theta,\theta^{'}\in\mathbb{R}^p$, we have
\begin{flalign}
\label{def:H}
\|\nabla J(\theta)-\nabla J(\theta^{'})\|_{2}\leq \ell\|\theta-\theta^{'}\|_2, 
\end{flalign} 
where $\ell=\frac{R_{\max}h(hG^{2}+L)}{(1-\gamma)}$, $h$ is a positive scalar that denotes the horizon of the trajectory $\tau$.
The property (\ref{def:H}) has been given as the Lipschitz assumption in previous works \citep{kumar2019sample,wang2020neural}, and it has been also verified by lots of recent works with some other regularity conditions \citep{zhang2019global,agarwal2019optimality,xu2020improvingsa}.

Furthermore, according to the Lemma 4.1 of \citep{shen2019hessian}, Assumption \ref{ass:On-policy-derivative} implies a property of the policy gradient estimator as follows,
for each $\tau\sim\pi_\theta$, we have
\begin{flalign}
	\label{def:sigma}
	\|g(\tau|\theta)-\nabla J(\theta)\|_{2}\leq\dfrac{GR_{\max}}{(1-\gamma)^2}=:\sigma.
\end{flalign}
The result of (\ref{def:sigma}) implies the boundedness of the variance of the policy gradient estimator $g(\tau|\theta)$, i.e., 
$\mathbb{V}\text{ar}(g(\tau|\theta))=\mathbb{E}[\|g(\tau|\theta)-\nabla J(\theta)\|^2_{2}]\leq\sigma^2$.
The boundedness of $\mathbb{V}\text{ar}(g(\tau|\theta))$ are also proposed as an assumption in the previous works \citep{Matteo2018Stochastic,xu2019improved,xu2020sample,wang2020neural}.
%The boundedness of the variance of the gradient estimator has been adopted in non-convex optimization, e.g., see \citep{ghadimi2016mini}.

\begin{assumption}[Smoothness of Policy Hessian]
\label{ass:hessian-lipschitz}
The the expected return function $J(\theta)$ is $\chi$-Hessian-Lipschitz, i.e., there exists a constant $0\leq\chi<\infty$ such that for all $\theta,\theta^{'}\in\mathbb{R}^{p}$:
\begin{flalign}
\label{def:hessian-lipschitz}
\|\nabla^{2} J(\theta)-\nabla^{2} J(\theta^{'})\|_{op}\leq  \chi\|\theta-\theta^{'}\|_2.
\end{flalign}
\end{assumption}
Assumption \ref{ass:hessian-lipschitz} requires that for the two near points, the Hessian matrix $\nabla^{2}J(\cdot)$ can not change dramatically in the terms of operator norm.
For RL, the parameter $\chi$ can be deduced by some other regularity conditions, e.g., \cite{zhang2019global} provides an estimation of $\chi$, see Appendix \ref{app-lemmas-01}.

\section{Second-Order Stationary Point}
\label{sec:assumption}

Due to the non-concavity of $J(\theta)$, finding global maxima is NP-hard in the worst case. 
The best one can hope is to convergence to stationary points. In this section, we formally define second-order stationary point (SOSP).
Furthermore, with the second-order information, we present Assumption \ref{local-concave} to make clear the maximal point that we mainly concern for policy optimization.

\begin{definition}
[Second-Order Stationary Point  \citep{nesterov2006cubic}
\label{def:sosp}
{\footnote{Recall problem (\ref{Eq:thata-optimal}) is a maximization problem, thus this definition of SOSP is slightly different from the minimization problem $\min_{x} f(x)$, where it requires $\|\nabla f(x)\|_2 \leq \epsilon~\text{and}~ \lambda_{\min}(\nabla^{2} f(x)) \ge0$. Similarly,
its $(\epsilon,\sqrt{\chi\epsilon})$-SOSP requires $\|\nabla f(x)\|_2 \leq \epsilon~\text{and}~ \lambda_{\min}(\nabla^{2} f(x)) \ge -\sqrt{\chi\epsilon}.$
}}]
For the $\chi$-Hessian-Lipschitz function $J(\cdot)$, we say that $\theta$ is a second-order stationary point if 
\begin{flalign}
\|\nabla J(\theta)\|_2 = 0~~~\emph{and}~~~ \lambda_{\max}(\nabla^{2} J(\theta)) \leq 0;
\end{flalign}
 we say $\theta$ is an $(\epsilon,\sqrt{\chi\epsilon})$-second-order stationary point if
 \begin{flalign}
 \label{condi-ep-sepo}
\|\nabla J(\theta)\|_2 \leq \epsilon~~~\emph{and}~~~ \lambda_{\max}(\nabla^{2} J(\theta)) \leq \sqrt{\chi\epsilon}.
\end{flalign}
\end{definition}
The SOSP is a very important concept for the policy optimization (\ref{Eq:thata-optimal}) because it rules the saddle points (whose Hessian are indefinite) and minimal points (whose Hessian are positive definite), which is usually more desirable than convergence to a first-order stationary point (FOSP).
Recently,
\cite{shen2019hessian,xu2020sample} introduce FOSP to measure the convergence of policy gradient methods.
As mentioned in Section \ref{sec:intro}, for policy optimization (\ref{Eq:thata-optimal}), an algorithm converges to a FOSP is not sufficient to ensure that algorithm outputs a maximal point. 
While SOSP overcomes above shortcomings, which is our main motivation to consider SOSP as a convergence criterion.

\begin{figure}[h]
\begin{center}
\begin{tikzpicture}
\tikzset{node style/.style={state, 
                                    minimum width=0.1cm,
                                    line width=0.3mm,
                                    fill=gray!80!white}}

        % Draw the states
        \node[node style] at (0, 0)     (s1)     {$s_0$};
        \node[node style] at (3, 0)     (s2)     {$s_1$};
         \node[node style] at (-3, 0)     (s3)     {$s_2$};
        %\node[node style] at (3, -5.196) (stagnant) {Stagnant};

        % Connect the states with arrows
        \draw[every loop,
              %auto=right,
              line width=0.3mm,
              %>=latex,
              draw=black,
              fill=black]
            (s3)     edge[loop above] node {$\mathtt{left}$} (s3)
            (s2)     edge[loop above] node {$\mathtt{right}$} (s2)
            %(s1)     edge[bend right=20]            node {$0.5$} (s2)
            (s1)     edge[ auto=right]            node {$p_1,\mathtt{right}$} (s2)
             (s1)     edge[auto=left]            node {$R_1=\mathtt{1}$} (s2)
            (s1)     edge[auto=left]             node {$p_2,\mathtt{left}$} (s3)
            (s1)     edge[ auto=right]             node {$R_2=\mathtt{1}$} (s3)
            (s1)     edge[loop above] node {$\mathtt{up},p_3,R_3=\mathtt{0}$} (s1)
            ;
    \end{tikzpicture}
 \end{center}
 \caption{Three-States MDP.}
 \label{figure-example}
 \end{figure}
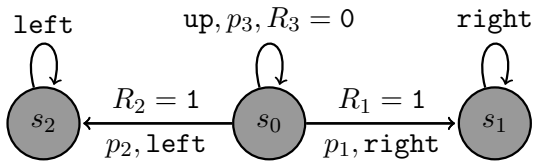

\begin{assumption}[Structure of $J(\theta)$]
\label{local-concave}
For any $\theta\in\mathbb{R}^{p}$, at least one of the following holds:
(i) $\|\nabla J(\theta)\|\ge\epsilon$; (ii) $\lambda_{\max} (\nabla^2J(\theta))\ge\sqrt{\epsilon\chi}$;
(iii) $\theta$ nears a local maximal point $\theta_{\star}$: there exists a positive scalar $\varrho $ such that $\theta$ falls in to the ball $\mathbb{B}_{2}(\theta_{\star},\varrho )$, 
and $J(\theta)$ is $\zeta$-strongly concave on $\mathbb{B}_{2}(\theta_{\star},\varrho )$.
\end{assumption}
In the standard non-convex literature such as \citep{ge2015escaping,jin2017escape}, the condition (ii) of Assumption \ref{local-concave} is often called $(\epsilon,\chi,\varrho)$-\emph{strict saddle}.
In this case, all the SOSP are local maxima and hence convergence to second-order stationary points is equivalent to convergence to local maxima.
In the following three-states MDP (see Figure \ref{figure-example}), we verify that the Assumption \ref{local-concave} holds on policy optimization.

 \begin{example}
 \label{example-strctly-saddle-mdp}
% We provide its dynamic transformation in Figure \ref{figure-example}.
In this deterministic MDP, the states $s_1$ and $s_2$ equip the action space $\mathcal{A}=\{\mathtt{right},\mathtt{left}\}$, while $s_0$ equips an additional action $\mathtt{up}$.
The policy $\pi_{\theta}(\cdot|\cdot)$ with a parameter $\theta=(\theta_1,\theta_2)^{\top}\in\mathbb{R}^{2}$, Let $ C_{o}=:[0,1]\times[0,1],$
if $\theta\in C_{o}$, we define
$
\pi_{\theta}(\mathtt{right}|s_0)=:p_1=\frac{1}{\sqrt{2\pi}}(1-\theta_1^2+\theta_2^2);
$
if $\theta\notin C_{o}$, we define Gaussian policy
$
\pi_{\theta}(\mathtt{left}|s_0)=:p_2=\frac{1}{\sqrt{2\pi}}{\exp\big\{-\frac{(2-\|\theta\|^2_2)}{2}\big\}};
$
otherwise, $\pi_{\theta}(\mathtt{up}|s_0)=:p_3=1-p_1-p_2$.
Then
$J(\theta)=p_1R_1+p_2R_2+p_3R_3$, i.e.,
\begin{flalign}
\label{example-objec}
J(\theta)=
 \begin{cases}
\frac{1}{\sqrt{2\pi}}(1-\theta_1^2+\theta_2^2),&\theta\in C_{o}\\
\frac{1}{\sqrt{2\pi}}{\exp\big\{-\frac{(2-\|\theta\|^2_2)}{2}\big\}},&\theta\notin C_{o}.
\end{cases}
\end{flalign}
The function $J(\theta)$ (\ref{example-objec}) satisfies Assumption \ref{local-concave}. Since the origin $(0,0)\in C_{o}$ is a saddle point of $J(\theta)$, and $\lambda_{\max}(\nabla^{2} J(\theta)|_{(0,0))}=1$, thus the point $(0,0)$ satisfies strict saddle point property.
Besides, on the complementary space of $C_{o}$, i.e., $\mathbb{R}^2-C_{o}$, $J(\theta)$ is a strongly concave function.
\end{example}
%to make clear the maximal point that we mainly concern for policy optimization.

%However, since the expected return $J(\theta)$ (\ref{J-objectiove}) is a typical non-concave function \citep{Matteo2018Stochastic,agarwal2019optimality}, an algorithm converges to a FOSP is not sufficient to ensure that algorithm generates a global or local maximum. 
%In fact, for the non-concave optimization problem (\ref{Eq:thata-optimal}), a FOSP can be a global maximum, local maximum, saddle point or even local minimum. 
%Clearly,
%for the maximum problem (\ref{Eq:thata-optimal}), both local minimum and saddle point are undesirable, which requires us to introduce new criterion to character the convergence of policy optimization (\ref{Eq:thata-optimal}).

\section{Main Result and Technique Overview}
\label{sec:mian-result}

Our contribution lies in this section.
Theorem \ref{complex-analyisis} presents the sample complexity of policy gradient algorithm (\ref{vpg}) finding an $(\epsilon,\sqrt{\chi\epsilon})$-SOSP.
Section \ref{sec:mian-result-02} provides an overview of the proof technique;
Section \ref{sec:local-improve} provides all the key steps, Section \ref{sec:Sketch} provides a sketch of the proof of Theorem \ref{complex-analyisis}.

\begin{theorem}
\label{complex-analyisis}
Under Assumption \ref{ass:On-policy-derivative}-\ref{local-concave}, 
consider $\{\theta_{k}\}_{k\ge0}$ generated by (\ref{vpg}), and $\iota$ is defined in (\ref{cnc}).
For a small enough step-size $\alpha$ such that
$\alpha\leq\min\left\{\dfrac{\epsilon^{2}}{2\sqrt{\chi\epsilon}R_{\min}^{2}\omega^2},\dfrac{2\epsilon^2}{(\epsilon^2+\sigma^2)\ell}\right\}=\mathcal{O}(\epsilon^{2}),$
the iteration (\ref{vpg}) returns an $(\epsilon,\sqrt{\chi\epsilon})$-SOSP with probability at least $1-\delta-\delta\log(\frac{1}{\delta})=1-\widetilde{O}(\delta)$ after the times of 
\[
K=
\left\lceil \dfrac{6R_{\max}}{\alpha^2(1-\gamma)\iota^2\sqrt{\chi\epsilon}}\log\dfrac{1}{\delta}\right\rceil+1
=
\mathcal{O}\left(
\dfrac{\epsilon^{-\frac{9}{2}}}{(1-\gamma)\sqrt\chi}\log\dfrac{1}{\delta}
\right).
\]
\end{theorem}

\begin{remark}
\label{remark-zhang-yang}
\emph
{
Theorem \ref{complex-analyisis} illustrates that policy gradient algorithm (\ref{vpg}) needs a cost of $\widetilde{\mathcal {O}}(\epsilon^{-\frac{9}{2}})$ to find an $(\epsilon,\sqrt{\chi\epsilon})$-SOSP.
To the best of our knowledge, \cite{zhang2019global} firstly consider to introduce SOSP to measure the sample complexity of policy-based RL.
\cite{zhang2019global} propose a \emph{modified random-horizon policy gradient} ($\mathtt{MRPG}$) algorithm, and they show that $\mathtt{MRPG}$ needs at least a cost of 
$\mathcal{O}
\big(
\epsilon^{-9}\chi^{\frac{3}{2}}\frac{1}{\delta}\log\frac{1}{\epsilon\chi}\big)=\widetilde{\mathcal{O}}(\epsilon^{-9})
$
to find an $(\epsilon,\sqrt{\chi\epsilon})$-SOSP.
Clearly, result of Theorem \ref{complex-analyisis} improves the sample complexity of \cite{zhang2019global} significantly from $\widetilde{\mathcal {O}}(\epsilon^{-{9}})$ to $\widetilde{\mathcal {O}}(\epsilon^{-\frac{9}{2}})$.
Additionally, compared to \cite{zhang2019global}, our analysis does \emph{not} invoke a geometric distribution restriction on the horizon. 
In the real work, the horizon of a trajectory only depends on the simulated environment, 
it is not necessary to draw a horizon from a geometric distribution, i.e., our result is more practical.
}
\end{remark}

\subsection{Technique Overview}
\label{sec:mian-result-02}

Recall that an $(\epsilon,\sqrt{\chi\epsilon})$-SOSP requires a point has a small gradient, and whose Hessian matrix does not have a significantly positive eigenvalue,
%If the current $\theta_{k}$ is \textbf{not} an $(\epsilon,\sqrt{\chi\epsilon})$-SOSP, i.e., it does not satisfy the condition of (\ref{condi-ep-sepo}),
which inspires us to consider an idea that decomposes the parameter space $\mathbb{R}^p$ into three non-intersected regions, and then analyzing them separately.
%: non-stationary region (\ref{region-L-1}), around saddle point (\ref{region-L-2}) and local optimal region (\ref{region-L-3}).
%It is noteworthy that the local optimal region, i.e., $\mathcal{L}_3$ is the desirable region where we expect policy gradient algorithm converges to.
%We formally define those regions as follows:
\begin{compactenum}[\ding{182}]
\item  \textbf{Case I: Non-Stationary Region.} In this case, we consider the region with large gradient, i.e.,
\begin{flalign}
\label{region-L-1}
\mathcal{L}_{1}=\{\theta\in\mathbb{R}^{p}: \|\nabla J(\theta)\|_2\ge\epsilon\};
\end{flalign}
\end{compactenum}
\begin{compactenum}[\ding{183}]
\item  \textbf{Case II: Around Saddle Point.} We consider the region where the norm of the policy gradient is small , while the maximum eigenvalue of the Hessian matrix $\nabla^{2} J(\theta)$ is larger than zero:
\begin{flalign}
\label{region-L-2}
\mathcal{L}_{2}=\{\theta\in\mathbb{R}^{p}: \|\nabla J(\theta)\|_2\leq\epsilon\}
\cap\{\theta\in\mathbb{R}^{p}: \lambda_{\max}(\nabla^{2} J(\theta)) \ge \sqrt{\chi\epsilon}\};
\end{flalign}
\end{compactenum}
\begin{compactenum}[\ding{184}]
\item  \textbf{Case III: Local Optimal Region.} In this case, we consider the region $
\mathcal{L}_3=\mathbb{R}^{p}-(\mathcal{L}_1 \cup \mathcal{L}_2)
$:
\begin{flalign}
\label{region-L-3}
\mathcal{L}_{3}=\{\theta\in\mathbb{R}^{p}: \|\nabla J(\theta)\|_2\leq\epsilon\}\cap\{\theta\in\mathbb{R}^{p}: \lambda_{\max}(\nabla^{2} J(\theta)) \leq \sqrt{\chi\epsilon}\}.
\end{flalign}
\end{compactenum}
It is noteworthy that the local optimal region, i.e., $\mathcal{L}_3$ is the desirable region where we expect policy gradient algorithm converges to it with high probability.
Before we provide the formal proof, in Section \ref{sec:local-improve}, we present three separate propositions to make local improvement on above three regions correspondingly.
The main challenge occurs on region $\mathcal{L}_2$, where we utilize a technique called correlated negative curvature (CNC) \citep{daneshmand2018escaping} to make a local improvement.
%Finally, in Section \ref{sec:Sketch}, we prove Theorem \ref{complex-analyisis}.
%All techniques can be potentially generalized to extensive policy optimization algorithms.

\subsubsection{Local Improvement on Each Case}
\label{sec:local-improve}
%Now, we present the results of local improvement on each region, and the details are in appendix.
%For case I, the standard stochastic gradient descent \citep{nesterov2013introductory} is powerful enough to ensure an improvement on $J(\theta)$ (\ref{J-objectiove}) after a single update, please see the next Proposition \ref{propo-2}.
\begin{proposition}[Local Improvement on $\mathcal{L}_1$]
\label{propo-2}
Under Assumption \ref{ass:On-policy-derivative}-\ref{ass:hessian-lipschitz}.
The sequence $\{\theta_{k}\}_{k\ge0}$ generated according to (\ref{vpg}).
If a point $\theta_k\in\mathcal{L}_1$, let $\alpha<\min\left\{\dfrac{2\epsilon^2}{(\epsilon^2+\sigma^2)\ell},\frac{2}{\ell}\right\}=\mathcal{O}(\epsilon^{-2})$, then we have
\begin{flalign}
\label{pro2:eq-2}
\mathbb{E}\big [J(\theta_{k+1})\big ]-J(\theta_{k})\overset{(\emph{\textbf{a}})}\ge\big(\alpha-\frac{\ell\alpha^{2}}{2}\big)\|\nabla J(\theta_{k})\|_{2}^{2}-
\frac{\ell\alpha^{2}\sigma^2}{2}\overset{(\emph{\textbf{b}})}\ge\frac{1}{2}\alpha\epsilon^2.
\end{flalign}
\end{proposition}
We provide its proof in Appendix \ref{app-proof-propo-2}.
Proposition \ref{propo-2} shows that when the gradient is large, the expected return $J(\theta)$ increases in one step.
It is noteworthy that the step-size plays an important role in achieving the result of (\ref{pro2:eq-2}).
Concretely, for a positive scalar  $\alpha-{\ell\alpha^{2}}/{2}$ (i.e., which requires $\alpha<{2}/{\ell}$), Eq.(\textbf{a})
of (\ref{pro2:eq-2}) guarantees the desired increase whenever the norm of the policy gradient is large enough.
At the same time, when considering the lower threshold value $\epsilon$ of the norm of the policy gradient in the region $\mathcal{L}_1$,
the second term of (\ref{pro2:eq-2}) achieves at least $\alpha\epsilon^2-\ell\alpha^2(\epsilon^2+\sigma^2)/2$.
Thus,
to make a clear improvement, the condition (\textbf{b}) requires step-size $\alpha$ should satisfy $\alpha<{2\epsilon^2}/{(\epsilon^2+\sigma^2)\ell}$.
\begin{proposition}[Local Improvement on $\mathcal{L}_{2}$]
\label{saddle-point-case}
Under Assumption \ref{ass:On-policy-derivative}-\ref{ass:hessian-lipschitz},
consider the sequence $\{\theta_{k}\}_{k\ge0}$ generated by (\ref{vpg}). If a point $\theta_{k}$ falls in to $\mathcal{L}_{2}$,
there exists a positive scalar $\iota$ (\ref{cnc}), and $\widehat{\kappa}_{0}$ such that
\begin{flalign}
\label{def:main-body-kappppo-1}
\widehat{\kappa}_{0}
=:
\left\lfloor \dfrac{\log\big({1}/({1-\sqrt{\alpha}\sigma_{H_0}})\big)}{\log(1+\alpha \sqrt{\chi\epsilon})}\right\rfloor=\mathcal{O}(\epsilon^{-\frac{1}{2}}), 
\end{flalign}
where $\sigma_{H_0}= \frac{2p\sqrt{p}hR_{\max}(hG^2+L)}{1-\gamma}$, then after at most $j\leq\widehat{\kappa}_{0}$ steps, we have
\begin{flalign}
\label{result-of-improvement-saddle-point}
\mathbb{E}[J(\theta_{k+j})]-J(\theta_{k})
\ge\alpha^{2} \iota^2\sqrt{\chi\epsilon}.
\end{flalign}
\end{proposition}
Proposition \ref{saddle-point-case} illustrates that even a point gets stuck in the region thar nears a saddle point, policy gradient method will ensure an increase in the value of $J(\theta)$ within at most $\mathcal{O}(\epsilon^{-\frac{1}{2}})$ steps.
We provide proof of Proposition \ref{saddle-point-case} in Appendix \ref{app-sec:proof-pro-saddle-point}.
The proof is very technical, the following correlated negative curvature (CNC) condition \citep{daneshmand2018escaping} plays a crucial role in achieving the result of (\ref{result-of-improvement-saddle-point}).
Concretely, let $u_p$ be the unit eigenvector corresponding to the maximum eigenvalue of $\nabla^2 J(\theta)$, 
CNC ensures the second moment of the projection of policy gradient estimator $g(\tau|\theta)$ along the direction $u_p$ is uniformly bounded away from zero, i.e., there exists a positive scalar $\iota$ s.t.,
\begin{flalign}
\label{cnc}
\textbf{CNC}:~~
\mathbb{E}[\langle g(\tau|\theta),u_p \rangle^2]\ge \iota^2,~~~\forall \theta \in \mathbb{R}^p.
\end{flalign}
We provide (\ref{cnc}) in the Discussion \ref{app-discussion} of Appendix \ref{app-proof-pro-sada-impro}. 
CNC shows that the perturbation caused by a stochastic policy gradient estimator $g(\tau|\theta)$ is guaranteed to take an increase in the value $J(\theta)$.
%Finally, it is noteworthy that \cite{zhang2019global} firstly check the CNC condition in policy optimization.
\begin{proposition} 
\label{propo-1}
Under Assumption \ref{ass:On-policy-derivative}-\ref{local-concave},
consider the sequence $\{\theta_{k}\}_{k\ge0}$ generated by (\ref{vpg}). For any $\delta\in(0,1)$, $\theta_{\star}$ satisfies Assumption \ref{local-concave}, let the step-size $\alpha$ and the stopping time $\kappa_0$ statisfy 
\[\alpha\leq\min\left\{\delta,\frac{1}{\zeta},\frac{\zeta}{\ell^{2}},\frac{\zeta\varrho^2}{3\sigma^2}\right\}, \alpha\log\frac{1}{\alpha}\leq\dfrac{2\zeta\varrho^4}{27\big({G^{2}R_{\max}^{2}}/{(1-\gamma)^{2}}+\zeta\varrho^{2}+\sigma^2\big)^2},\kappa_{0}
=\Big\lfloor\frac{1}{\alpha^2}\log\frac{1}{\delta}\Big\rfloor,
\]
and if some iteration $\theta_{k}$ falls into the ball $\mathbb{B}_{2}(\theta_{\star},\frac{\sqrt{3}}{3}\varrho)\subset \mathbb{B}_{2}(\theta_{\star},\varrho) $, i.e., $\|\theta_{k}-\theta_{\star}\|_{2}^{2}\leq\frac{1}{3}\varrho^2$. Then,%with probability at least $1-\delta\log\dfrac{1}{\delta}$, for all $j\in[0,\kappa_0-1]$, we have
%$\big\|\theta_{k+j}-\theta_{\star}\big\|_{2}\leq\varrho$, i.e.,
\[
\mathbb{P}\big(\big\|\theta_{k+j}-\theta_{\star}\big\|_{2}\leq\varrho\big)\ge 1-\delta\log\frac{1}{\delta},~\forall~j\in[0,\kappa_0-1].
\]
\end{proposition}
We provide its proof in Appendix \ref{sec:app-proof-pro1-sec}.
Proposition \ref{propo-1} illustrates that once an iteration gets sufficiently close to a local optimum $\theta_{\star}$, it can get trapped in the neighborhood of $\theta_{\star}$ for a really long time. 
%Although the Assumption \ref{local-concave} ensures the local $\zeta$-strongly concavity of the 
%expected return function $J(\theta)$ (\ref{J-objectiove}) in the neighborhood $\mathbb{B}_{2}(\theta_{\star},\varrho)$, 
%it is still non-trivial to achieve the result of (\ref{high-pro-near-optimal-starr}) since 
%Finally, it is noteworthy that since $\lim_{\alpha\rightarrow 0^{+}}\alpha\log\frac{1}{\alpha}=0$, then for an enough small $\alpha$, the condition of the step-size in Proposition \ref{propo-1} alway exists.

\subsubsection{Proof Sketch of Theorem \ref{complex-analyisis}}
\label{sec:Sketch}

\begin{proof}
Our proof contains three steps.
\begin{compactenum}
\item
 \textbf{Firstly}, we will prove that within $\left\lceil\dfrac{6R_{\max}}{\alpha^2(1-\gamma)\iota^2\sqrt{\chi\epsilon}}\right\rceil$ steps, with probability at least $\dfrac{1}{2}$, one iteration falls into $\mathcal{L}_3$.
Let above procedure lasts $\left\lceil \log\dfrac{1}{\delta}\right\rceil $ steps, according to the \emph{inclusion-exclusion} formula of probability: after $K_{o}=\left\lceil\dfrac{6R_{\max}}{\alpha^2(1-\gamma)\iota^2\sqrt{\chi\epsilon}}\log\dfrac{1}{\delta}\right\rceil$ steps, with probability $1-\delta$, one of $\{\theta_{k}\}_{k\ge0}$ falls into $\mathcal{L}_3$.
\item\textbf{Secondly},
Proposition \ref{propo-1} shows that once an iteration enters the region $\mathcal{L}_3$, the iteration gets trapped there for at least $\kappa_0$ steps with probability $1-\delta\log\dfrac{1}{\delta}$.
\item\textbf{Finally}, let $K\in(K_{o}+1,K_{o}+\kappa_0)$, combining above two results, the output $\theta_{K}$ falls into the region $\mathcal{L}_3$ with probability at least $1-(\delta+\delta\log\dfrac{1}{\delta})$, which concludes the result of Theorem \ref{complex-analyisis}.
\end{compactenum}

Now, we only need to prove: starting from any point, within  $\left\lceil\dfrac{6R_{\max}}{\alpha^2(1-\gamma)\iota^2\sqrt{\chi\epsilon}}\right\rceil$ steps, with probability at least $\dfrac{1}{2}$, one of $\{\theta_{k}\}_{k\ge0}$ falls into $\mathcal{L}_3$.

We define a stochastic process $\{\varsigma_k\}_{k\ge0}$ ($\varsigma_0=0$) to trace the numbers of samples,
\[
\varsigma_{k+1}= 
\begin{cases}
\varsigma_{k}+1~&\text{if}~\theta_{\varsigma_{k}} \in\mathcal{L}_{1}\cup\mathcal{L}_{3}\\
\varsigma_{k}+\widehat{\kappa}_{0}~&\text{if}~\theta_{\varsigma_{k}} \in\mathcal{L}_{2},
\end{cases}
\]
where $\widehat{\kappa}_{0}$ is defined in (\ref{def:main-body-kappppo-1}). Let $\beta=\iota^{2}\sqrt{\chi\epsilon}$, we can rewrite the results of Proposition \ref{propo-2}-\ref{saddle-point-case} as follows,
\begin{flalign}
\label{proof-inter01}
\Ex[J(\theta_{\varsigma_{k+1}})-J(\theta_{\varsigma_{k}})|\theta_{\varsigma_{k}}\in\mathcal{L}_1]\overset{(\ref{pro2:eq-2})}\ge\frac{1}{2}\alpha\epsilon^{2},~~~~
\Ex[J(\theta_{\varsigma_{k+1}})-J(\theta_{\varsigma_{k}})|\theta_{\varsigma_{k}}\in\mathcal{L}_2]\overset{(\ref{result-of-improvement-saddle-point})}\ge\alpha^{2}\beta.
\end{flalign}
Putting the results of (\ref{proof-inter01}) together, let $\alpha^{2} \beta\leq\frac{1}{2}\alpha\epsilon^{2}$, i.e., $\alpha\leq\frac{\epsilon^{2}}{2\beta}$ ,we have
\begin{flalign}
\label{proof-inter002}
\Ex[J(\theta_{\varsigma_{k+1}})-J(\theta_{\varsigma_{k}})|\theta_{\varsigma_{k}}\not\in\mathcal{L}_3]\ge\alpha^2\beta\Ex[(\varsigma_{k+1}-\varsigma_{k})|\theta_{\varsigma_{k}}\not\in\mathcal{L}_3].
\end{flalign}
We define the event $\mathcal{E}_{k}$
\[
\mathcal{E}_{k}=\bigcap_{j=0}^{k}\big\{j:\theta_{\varsigma_j}\not\in\mathcal{L}_3\big\}.
\]
Let $\bm{1}_{A}$ denote indicator function, where if event $A$ happens, $\bm{1}_{A}=1$, otherwise $\bm{1}_{A}=0$, then
%We consider the term $J(\theta_{\varsigma_{k}})\bm{1}_{\mathcal{E}_{{k}}}$ as follows,
\begin{flalign}
\nonumber
\mathbb{E}[J(\theta_{\varsigma_{k+1}})\bm{1}_{\mathcal{E}_{{k+1}}}-J(\theta_{\varsigma_{k}})\bm{1}_{\mathcal{E}_{{k}}}]
 =&\mathbb{E}[J(\theta_{\varsigma_{k+1}})(\bm{1}_{\mathcal{E}_{{k+1}}}-\bm{1}_{\mathcal{E}_{{k}}})]+\mathbb{E}[(J(\theta_{\varsigma_{k+1}})-J(\theta_{\varsigma_{k}}))
\bm{1}_{\mathcal{E}_{{k}}}]\\
\label{prof-them-1-int}
\overset{(\ref{proof-inter01})}\ge&-\frac{R_{\max}}{1-\gamma}(\mathbb{P}(\mathcal{E}_{{k+1}}-\mathcal{E}_{{k}}))+
\alpha^2\beta\Ex[\varsigma_{k+1}-\varsigma_{k}|\bm{1}_{\mathcal{E}_{{k}}}]\mathbb{P}(\mathcal{E}_{{k}}),
\end{flalign}
where we use the boundedness of $J(\theta)\ge-\dfrac{R_{\max}}{1-\gamma}$.

Summing the above expectation (\ref{prof-them-1-int}) over $k$, then
\begin{flalign}
\nonumber
\mathbb{E}[J(\theta_{\varsigma_{k+1}})\bm{1}_{\mathcal{E}_{{k+1}}}]-J(\theta_0)
&=
-\frac{R_{\max}}{1-\gamma}(\mathbb{P}(\mathcal{E}_{{k+1}})-\mathbb{P}(\mathcal{E}_{{0}}))+\alpha^2\beta\sum_{j=0}^{k}\big(\Ex[\varsigma_{j+1}]\mathbb{P}(\mathcal{E}_{{j}})-\Ex[\varsigma_{j}]\mathbb{P}(\mathcal{E}_{{j}})\big)\\
\label{prof-1-101}
&\ge
-\frac{R_{\max}}{1-\gamma}+\alpha^2\beta\sum_{j=0}^{k}\big(\Ex[\varsigma_{j+1}]\mathbb{P}(\mathcal{E}_{{j+1}})-\Ex[\varsigma_{j}]\mathbb{P}(\mathcal{E}_{{j}})\big)\\
\label{prof-1-102}
&=
-\frac{R_{\max}}{1-\gamma}+\alpha^2\beta\Ex[\varsigma_{k+1}]\mathbb{P}(\mathcal{E}_{{k+1}}),
\end{flalign}
where Eq.(\ref{prof-1-101}) holds since $\mathbb{P}(\mathcal{E}_{{k+1}})-\mathbb{P}(\mathcal{E}_{{0}})\leq1$; $\mathcal{E}_{j+1}\subset\mathcal{E}_{j}$ implies $\mathbb{P}(\mathcal{E}_{j+1})\leq \mathbb{P}(\mathcal{E}_{j})$; and Eq.(\ref{prof-1-102}) holds since $\varsigma_0 =0$.

Finally, since
$\mathbb{E}[J(\theta_{\varsigma_{k+1}})\bm{1}_{\mathcal{E}_{{k+1}}}-J(\theta_{\varsigma_{k}})\bm{1}_{\mathcal{E}_{{k}}}]\leq\dfrac{2R_{\max}}{1-\gamma}$, 
from the results of (\ref{prof-1-102}), 
if \[\mathbb{E}[\varsigma_{k+1}]\ge\dfrac{6R_{\max}}{\alpha^2(1-\gamma)\beta}=\dfrac{6R_{\max}}{\alpha^2(1-\gamma)\iota^2\sqrt{\chi\epsilon}},\] 
then $\mathbb{P}[\mathcal{E}_{k+1}]\leq\dfrac{1}{2}$. This concludes the proof.
\end{proof}

%% file: related-work.tex
\section{Related Work and Future Work}
\label{sec:Related Work}

Compared to the tremendous empirical works, theoretical results of policy gradient methods are relatively scarce.
In this section, we compare our result with current works in the following discussion.
For clarity, we have presented the complexity comparison to some results in Table \ref{table:comparison}.
Furthermore, we discuss future works to extend our proof technique to other policy gradient methods.

\subsection{First-Order Measurement}
According to \citet{shen2019hessian}, $\mathtt{REINFORCE}$ needs $\mathcal{O}(\epsilon^{-4})$ random trajectories to achieve the $\epsilon$-FOSP, and no provable improvement on its complexity has been made so far.
\cite{xu2019improved} also notice 
the order of sample complexity of $\mathtt{REINFORCE}$ and $\mathtt{GPOMDP}$ \citep{baxter2001infinite} reaches $\mathcal{O}(\epsilon^{-4})$.
With an additional assumption
$\mathbb{V}{\text{ar}}\Big[\prod_{i\ge 0}\dfrac{\pi_{\theta_0}(a_i|s_i)}{\pi_{\theta_t}(a_i|s_i)}\Big]$,$\mathbb{V}{\text{ar}}[g(\tau|\theta)]<+\infty$, 
\cite{Matteo2018Stochastic} show that the $\mathtt{SVRPG}$ needs sample complexity of $\mathcal{O}(\epsilon^{-4})$ to 
achieve the $\epsilon$-FOSP.
Later, under the same assumption as \citep{Matteo2018Stochastic}, \cite{xu2019improved} reduce the sample complexity of $\mathtt{SVRPG}$ to $\mathcal{O}(\epsilon^{-\frac{10}{3}})$.
Recently, \cite{shen2019hessian}, \cite{yang2019policy} and \cite{xu2020sample} introduce stochastic variance reduced gradient (SVRG) techniques \citep{johnson2013accelerating,nguyen2017sarah,fang2018spider} to policy optimization, their new methods improve sample complexity to $\mathcal{O}(\epsilon^{-3})$ to 
achieve an $\epsilon$-FOSP.
\cite{pham2020hybrid} propose $\mathtt{ProxHSPGA}$ that is a hybrid stochastic policy gradient estimator by combining existing $\mathtt{REINFORCE}$ estimator with the adapted $\mathtt{SARAH}$ \citep{nguyen2017sarah} estimator. 
\cite{pham2020hybrid} show  $\mathtt{ProxHSPGA}$ also need $\mathcal{O}(\epsilon^{-3})$ trajectories to achieve the $\epsilon$-FOSP.
To compare clearly, we summarize more details of the comparison in Table \ref{table:comparison}.

\subsection{Second-Order Measurement} As mentioned in the previous section, for RL, an algorithm converges to a FOSP is not sufficient to ensure that algorithm outputs a maximal point, which is our main motivation to consider SOSP to measure the convergence of policy gradient method.
To the best of our knowledge, \cite{zhang2019global} firstly introduce SOSP to RL to measure the sample complexity of policy gradient methods.
\cite{zhang2019global} propose $\mathtt{MRPG}$ that needs at least $\widetilde{\mathcal{O}}(\epsilon^{-9})$ samples, which is worse than our result $\widetilde{\mathcal{O}}(\epsilon^{-\frac{9}{2}})$.
We have discussed this comparison in the previous Remark \ref{remark-zhang-yang}.

Additionally, it is noteworthy that although we are all adopting the CNC technique to ensure the local improvement on saddle point region, 
our technique is different from \cite{zhang2019global} at least from two aspects: Firstly, our CNC condition is more general since we consider the fundamental policy gradient estimator (\ref{vpg}) and our analysis can be extended to generalized to extensive policy optimization algorithms; while the CNC result of \cite{zhang2019global} is limited in their proposed algorithm $\mathtt{MRPG}$;
Secondly, on the region $\mathcal{L}_2$, our result shows that within at most $\mathcal{O}(\epsilon^{-\frac{1}{2}})$ steps, policy gradient ensures an increase in the value of $J(\theta)$.
While, \cite{zhang2019global} require $\Omega(\epsilon^{-5}\log{\frac{1}{\epsilon}})$, which is the main reason why our analysis to achieve a better sample complexity.

\begin{table}
\centering
\begin{tabular}{|c|c|c|c|}
\hline
Algorithm & Conditions & Guarantee & Complexity \\
\hline
$\mathtt{REINFORCE}$~\citep{williams1992simple}&Assumption \ref{ass:On-policy-derivative} & First-Order&$\mathcal{O}(\epsilon^{-4})$\\
\hline
%\makecell{SVRG~\citep{reddi2016stochastic}\\
%\citep{allen2016variance}} & $O(\frac{n^{2/3}}{\epsilon^2}+n)$ & 1st-Order & \checkmark \\
%\hline
\makecell{$\mathtt{GPOMDP}$\\\citep{baxter2001infinite}}&Assumption \ref{ass:On-policy-derivative} &First-Order&$\mathcal{O}(\epsilon^{-4})$ \\
\hline
\makecell{$\mathtt{SVRPG}$\\\citep{Matteo2018Stochastic} }& 
\makecell{Assumption \ref{ass:On-policy-derivative}\\
$\mathbb{V}\text{ar}\Big[\prod_{i\ge0}\frac{\pi_{\theta_0}(a_i|s_i)}{\pi_{\theta_t}(a_i|s_i)}\Big]<+\infty$} &First-Order & $\mathcal{O}(\epsilon^{-4})$ \\
\hline
\makecell{$\mathtt{SVRPG}$\\\citep{xu2019improved} }& 
\makecell{Assumption \ref{ass:On-policy-derivative}\\
$\mathbb{V}\text{ar}\Big[\prod_{i\ge0}\frac{\pi_{\theta_0}(a_i|s_i)}{\pi_{\theta_t}(a_i|s_i)}\Big]<+\infty$} & First-Order & $\mathcal{O}(\epsilon^{-\frac{10}{3}})$ \\
\hline
$\mathtt{HAPG}$~\citep{shen2019hessian}&Assumption \ref{ass:On-policy-derivative} &First-Order&$\mathcal{O}(\epsilon^{-3})$\\
\hline
$\mathtt{VRMPO}$~\citep{yang2019policy}&Assumption \ref{ass:On-policy-derivative} &First-Order&$\mathcal{O}(\epsilon^{-3})$\\
\hline
\makecell{$\mathtt{SRVR}$-$\mathtt{PG}$\\\citep{xu2020sample}}&\makecell{Assumption \ref{ass:On-policy-derivative}\\
$\mathbb{V}\text{ar}\Big[\prod_{i\ge0}\frac{\pi_{\theta_0}(a_i|s_i)}{\pi_{\theta_t}(a_i|s_i)}\Big]<+\infty$} &First-Order&$\mathcal{O}(\epsilon^{-3})$\\
\hline
\makecell{$\mathtt{ProxHSPGA}$\\\citep{pham2020hybrid} }& 
\makecell{Assumption \ref{ass:On-policy-derivative}\\
$\mathbb{V}\text{ar}\Big[\prod_{i\ge0}\frac{\pi_{\theta_0}(a_i|s_i)}{\pi_{\theta_t}(a_i|s_i)}\Big]<+\infty$} & First-Order & $\mathcal{O}(\epsilon^{-{3}})$ \\
 \rowcolor{blue!20}
\hline
$\mathtt{MRPG}$~\citep{zhang2019global}&Assumption \ref{ass:On-policy-derivative} and Eq.(\ref{fisher-information})&Second-Order&$\widetilde{\mathcal{O}}\big(\epsilon^{-{9}}\big)$\\
 \rowcolor{blue!20}
\hline
Our work &Assumption \ref{ass:On-policy-derivative}-\ref{local-concave} &Second-Order&$\widetilde{\mathcal{O}}\big(\epsilon^{-\frac{9}{2}}\big)$\\
\hline
\end{tabular}
 \vspace{8pt}
\caption{
Complexity comparison, where the result of first-order requires $\|\nabla J(\theta)\|_2\leq\epsilon$, section-order requires an additional condition $\lambda_{\max}(\nabla^{2} J(\theta)) \leq \sqrt{\chi\epsilon}$.
}
\label{table:comparison}
\end{table}

\subsection{Future Work} 

 In this paper, we mainly consider  Monte Carlo gradient estimator (\ref{g-tau}), the technique of proof can be generalized to extensive policy gradient methods such as replacing $R(\tau)$ with state-action value function $Q^{\pi}(s_t,a_t)$, advantage function $A^{\pi}(s_t,a_t)$, baseline function $R(\tau)-V^{\pi}(s_t,a_t)$, and temporal difference error $r_{t+1}+\gamma V^{\pi}(s_{t+1},a_{t+1})-V^{\pi}(s_{t},a_{t})$.

Our result of $\widetilde{\mathcal{O}}(\epsilon^{-\frac{9}{2}})$ to achieve $(\epsilon,\sqrt{\epsilon\chi})$-SOSP is still far from the best-known $\epsilon$-FOSP result $\mathcal{O}(\epsilon^{-3})$.
In theory, \cite{allen2018neon2} and \cite{xu2018first} independently show that finding a SOSP is not much harder than FOSP.
Recently, in non-convex optimization, \cite{ge2019stabilized} show that with a simple variant of SVRG, we can find a SOSP that almost matches the known the first-order stationary points.

This provides a motivation that we can introduce some latest developments such as \citep{du2017gradient,daneshmand2018escaping,jin2018accelerated,zhou2018finding,zhou2018stochastic,ge2019stabilized,fang2019sharp} to give some fresh understanding to RL algorithms.
Besides, it will be also interesting to rethink the sample complexity of SOSP of the works \citep{Matteo2018Stochastic,shen2019hessian,yang2019policy,pham2020hybrid}, where they have proposed SVRG version of policy gradient methods.

 It is noteworthy that we don't consider the actor-critic type algorithms.
Recently \cite{yang2019provably,kumar2019sample,agarwal2019optimality,xu2020improvingsa,wang2020neural} have analyzed the complexity of actor-critic or natural actor-critic algorithms, and it will be interesting to rethink the sample complexity of SOSP of actor-critic or natural actor-critic algorithms .

%Additionally, also due to the non-concavity of its objective, until recently, searching the global maximum of policy-based RL is very challenging.
%\cite{hillar2013most} show that searching the global optimal solution of a very simple non-concave function likes a degree 4 polynomial can be NP-hard.
%Thus, a practical direction is to consider policy gradient converges to the local optimality.

\section{Conclusion}

In this paper, we provide the sample complexity of the policy gradient method finding second-order stationary points.
Our result shows that policy gradient methods converge to an $(\epsilon,\sqrt{\epsilon\chi})$-SOSP  at a cost of $\widetilde{\mathcal{O}}(\epsilon^{-\frac{9}{2}})$,
which improves the the best-known result of by a factor of  $\widetilde{\mathcal{O}}(\epsilon^{-\frac{9}{2}})$.
Besides, we think the technique of proof can be potentially generalized to extensive policy optimization algorithms, and give some fresh understanding to the existing algorithms.

%% file: table.tex
\section{Table of Notations}

For convenience of reference, we list key notations that have be used in this paper.

\begin{tabular}{r c p{11cm}}
    $\mathcal{S}$ &: & The set of states.\\   
    $\mathcal{A}$ &: & The set of actions.\\ 
    $P(s^{'}|s,a)$  &: &The probability of state transition from $s$ to $s^{'}$ under playing the action $a$. \\
    $\rho_{0}$&: & $\rho_{0}(\cdot):\mathcal{S}\rightarrow[0,1]$ is the initial state distribution.\\
     $\gamma$ &: &The discount factor, and $\gamma\in(0,1)$.\\
     $\tau$ &: &The trajectory generated according to $\pi_{\theta}$, i.e.,$\tau=\{s_{t}, a_{t}, r_{t+1}\}_{t\ge0}\sim\pi_{\theta}$.\\
     $P^{\pi_{\theta}}(s_t|s_0)$  &: & The probability of visiting the state $s_t$ after $t$
time steps from the initial state $s_0$ by executing $\pi_{\theta}$.\\
$d^{\pi_{\theta}}_{s_0}(s)$&: & The (unnormalized) discounted stationary state distribution of the Markov chain (starting at $s_0$) induced by $\pi_{\theta}$,
$d^{\pi_{\theta}}_{s_0}(s)=\sum_{t=0}^{\infty}\gamma^{t}P^{\pi_{\theta}}(s_t=s|s_0)$.\\
$d^{\pi_{\theta}}_{\rho_0}(s)$&: & $d^{\pi_{\theta}}_{\rho_0}(s)=\mathbb{E}_{s_0\sim\rho_{0}(\cdot)}[d^{\pi_{\theta}}_{s_0}(s)]$\\
$J(\theta)$&:& The performance objective, defined in (\ref{J-objectiove}).\\
$g(\tau|\theta)$&:& An policy gradient estimator $g(\tau|\theta)=\sum_{t=0}^{\infty}\nabla\log\pi_{\theta}(a_{t}|s_{t})R(\tau)$, defined in (\ref{g-tau}).\\
$F(\theta)$&:& Fisher information matrix defined in (\ref{fisher-information}).\\
$\omega$&:& A positive scaler defined in (\ref{fisher-information}).\\
$G$,$L$,$U$&:& Two positive scalers defined in (\ref{def:F-G}).\\
$\ell$&:& Lipschitz parameter of $\nabla J(\theta)$, and it is defined in (\ref{def:H}).\\
$\sigma$&:&An upper-bound of $\mathbb{V}\text{ar}(g(\tau|\theta))$, and it is defined in (\ref{def:sigma}).\\
$\chi$&:&Hessian-Lipschitz parameter $\chi$ presented in (\ref{def:hessian-lipschitz}).
\end{tabular}

%% file: app-lemma.tex
\section{Some Lemmas}
\label{app-lemmas-01}
\begin{lemma}[Azuma's Inequality]
\label{lem:Azuma-Inequality}
Let $\{Z_t\}_{t\in\mathbb{N}}$ be a martingale with respect to the filtration: $\mathcal{F}_0\subset\mathcal{F}_1\cdots\subset\mathcal{F}_{t}\subset\cdots$. Assume that there are predictable processes $\{A_t\}_{t\in\mathbb{N}}$ and $\{B_t\}_{t\in\mathbb{N}}$ (i.e., $A_t,B_t \in \mathcal{F}_{t-1}$) and constants $0 < c_t < +\infty$ such that: for all $t \ge 1$, almost surely,
\[A_t \leq Z_t -Z_{t-1} \leq B_t ~~~~\text{and} ~~~~ B_t -A_t \leq c_t.\]
Then, for all $\delta>0$, 
\[
\mathbb{P}[Z_t-Z_0\ge\delta]\leq\exp
\bigg(
-\dfrac{2\delta^2}{\sum_{i= 0}^{t} c_{i}^{2}}
\bigg)
.
\]
\end{lemma}
For the proof of Azuma's inequality, please refer to 
\url{http://www.math.wisc.edu/~roch/grad-prob/gradprob-notes20.pdf}.

The following Lemma \ref{value-difference} illustrates the difference of the performance between two policies and it is helpful throughout this paper.
\begin{lemma}
\label{value-difference}
\emph{(\citep{kakade2002approximately})}
For any policy $\pi$, $\tilde{\pi}$, and any initial state $s_0\in\mathcal{S}$, we have
\begin{flalign}
\label{kakade-2002}
V^{\pi}(s_0)-V^{\tilde{\pi}}(s_0)=\mathbb{E}_{s\sim d^{\pi}_{s_0}(\cdot),a\sim\pi(\cdot|s)}[A^{\tilde{\pi}}(s,a)].
\end{flalign}
\end{lemma}

\begin{lemma}[\citep{zhang2019global}]
\label{app-lemma-chipaper}
Under Assumption \ref{ass:On-policy-derivative}. With an additional condition as follows,
\[
\|
\nabla^2 \log \pi_{\theta_1}(a|s)-\nabla^2 \log \pi_{\theta_2}(a|s)\|_2
\leq W\|\theta_1-\theta_2\|_2,
\]
then we have estimate of $\chi$ such that 
\[
\|\nabla^{2} J(\theta)-\nabla^{2} J(\theta^{'})\|_{op}\leq  \chi\|\theta-\theta^{'}\|_2,
\]
where
\[
\chi=:\dfrac{R_{\max}GL}{(1-\gamma)^2}
+
\dfrac{R_{\max}G^{3}(1+\gamma)}{(1-\gamma)^3}
+
\dfrac{R_{\max}G}{1-\gamma}
\max\bigg\{
L,\dfrac{\gamma G^2}{1-\gamma},
\dfrac{W}{G},
\dfrac{L\gamma}{1-\gamma},
\dfrac{G(1+\gamma)+L\gamma(1-\gamma)}{1-\gamma^2}
\bigg\}.
\]
\end{lemma}

\begin{remark}
\emph{
Lemma \ref{app-lemma-chipaper} illustrates, with some other regularity conditions, we can give a concrete estimate of the parameter $\chi$ that satisfies Assumption \ref{ass:hessian-lipschitz}.
This Assumption \ref{ass:hessian-lipschitz} significantly simplifies the theory analysis, but it could be removed by other regularity conditions.
}
\end{remark}

%% file: proof-pro-2.tex
\clearpage
\section{Proof of Proposition \ref{propo-2}}
\label{app-proof-propo-2}

\textbf{Proposition} \ref{propo-2}
\emph{
Under Assumption \ref{ass:On-policy-derivative}-\ref{local-concave}.
The sequence $\{\theta_{k}\}_{k\ge0}$ generated according to (\ref{vpg}).
If a point $\theta_k\in\mathcal{L}_1$, i.e., $\Big\|\nabla J(\theta_k)\Big\|_2\ge \epsilon$, and let $\alpha<\Big\{\dfrac{2\epsilon^2}{(\epsilon^2+\sigma^2)\ell},\dfrac{2}{\ell}\Big\}$, then after one update, the following holds
\begin{flalign}
\nonumber
\mathbb{E}\Big [J(\theta_{k+1})\Big ]-J(\theta_{k})&\ge\Big(\alpha-\dfrac{\ell\alpha^{2}}{2}\Big)\Big\|\nabla J(\theta_{k})\Big\|_{2}^{2}-
\dfrac{\ell\alpha^{2}\sigma^2}{2}\\
\nonumber
&\ge\alpha\epsilon^2-\frac{1}{2}\ell\alpha^2(\epsilon^2+\sigma^2)>0.
\end{flalign}
}

\begin{proof}(of Proposition \ref{propo-2})

Recall for each $k\in\mathbb{N}$, $\tau_{k}\sim\pi_{\theta_k}$, to simplify expression, we introduce a notation as follows,
\[g(\theta_k)=:g(\theta_k|\tau_k)=\sum_{t\ge0}\nabla_{\theta}\log\pi_{{\theta}}(a_{t}|s_{t})R(\tau_k)|_{\theta=\theta_k}\]
Then, the update (\ref{vpg}) can be rewritten as follows,
\[
\theta_{k+1}=\theta_{k}+\alpha g(\theta_k).
\]

Let
\begin{flalign}
\label{def:xi_k}
\xi_{k}=g(\theta_k)-\nabla J(\theta_k),
\end{flalign}
By the fact $\nabla J(\theta_k)=\mathbb{E}[g(\theta_k)]$, we have $\mathbb{E}[\xi_k]=0.$

By the result of (\ref{def:H}), i.e., for each $\theta,\theta^{'}$, $\big\|\nabla J(\theta)-\nabla J(\theta^{'})\big\|_{2}\leq \ell\big\|\theta-\theta^{'}\big\|_2$, which implies
\begin{flalign}
|J(\theta^{'})-J(\theta)-\langle \nabla J(\theta),\theta^{'}-\theta\rangle|\leq\dfrac{\ell}{2}\big\|\theta-\theta^{'}\big\|_{2}^{2}.
\end{flalign}
Then, we have
\begin{flalign}
\nonumber
J(\theta_{k+1})-J(\theta_{k})\ge\langle \nabla J(\theta_{k}),\theta_{k+1}-\theta_{k}\rangle-\dfrac{\ell}{2}\big\|\theta_{k}-\theta_{k+1}\big\|_{2}^{2},
\end{flalign}
which implies 
\begin{flalign}
\nonumber
\mathbb{E}\big [J(\theta_{k+1})\big ]-J(\theta_{k})&\ge \nabla J(\theta_{k})^{\top}\mathbb{E}\big [\theta_{k+1}-\theta_{k}\big ]-\dfrac{\ell}{2}\mathbb{E}\big [\big\|\theta_{k}-\theta_{k+1}\big\|_{2}^{2}\big ]\\
\nonumber
&=\nabla J(\theta_{k})^{\top}\mathbb{E}\big [\alpha g(\theta_k)\big ]-\dfrac{\ell\alpha^{2}}{2}\mathbb{E}\big [\big\|\xi_{0}+\nabla J(\theta_k)\big\|_{2}^{2}\big ]\\
\nonumber
&=\nabla J(\theta_{k})^{\top}\mathbb{E}\big [\alpha g(\theta_k)\big ]-
\dfrac{\ell\alpha^{2}}{2}\mathbb{E}\big [\big\|\xi_{0}^{\top}\xi_{0}+2\xi_{0}^{\top}\nabla J(\theta_k)+\nabla J(\theta_k)^{\top}\nabla J(\theta_k)\big ]\\
\nonumber
&=\big(\alpha-\dfrac{\ell\alpha^{2}}{2}\big)\big\|\nabla J(\theta_{k})\big\|_{2}^{2}-
\dfrac{\ell\alpha^{2}}{2}\mathbb{E}\big [\underbrace{\xi_{0}^{\top}\xi_{0}}_{=\big\|\nabla J(\theta_{k})-g(\theta_k)\big\|_{2}^{2}}\big ]\\
\nonumber
&\overset{\eqref{def:sigma}}\ge\big(\alpha-\dfrac{\ell\alpha^{2}}{2}\big)\big\|\nabla J(\theta_{k})\big\|_{2}^{2}-
\dfrac{\ell\alpha^{2}\sigma^2}{2}\\
\nonumber
&\ge\big(\alpha-\dfrac{\ell\alpha^{2}}{2}\big)\epsilon^2-
\dfrac{\ell\alpha^{2}\sigma^2}{2}=\alpha\epsilon^2-\dfrac{1}{2}\ell\alpha^2\epsilon^2-\dfrac{1}{2}\ell\alpha^2\sigma^2,
\end{flalign}
Solving $\dfrac{1}{2}\alpha\epsilon^2-\dfrac{1}{2}\ell\alpha^2\epsilon^2-\dfrac{1}{2}\ell\alpha^2\sigma^2>0$, we have $\alpha<\dfrac{\epsilon^2}{(\epsilon^2+\sigma^2)\ell}$, then 
we have
\[
\mathbb{E}\big [J(\theta_{k+1})\big ]-J(\theta_{k})\ge\dfrac{1}{2}\alpha \epsilon^2.
\]
This concludes the proof.
\end{proof}

%% file: proof-pro-1.tex
\clearpage
\section{Proof of Proposition \ref{propo-1}}
\label{sec:app-proof-pro1-sec}

In this section we provide the details of the proof of Proposition \ref{propo-1}.
We need the following Lemma \ref{lemma:Boundedness of the near iteration}, Lemma \ref{boundedness-of-z-j-lemma} and Lemma \ref{lemma-bound-the-kapp-gp-0} to show the result of Proposition \ref{propo-1}.

\textbf{Proposition \ref{propo-1}}\emph{
Under Assumption \ref{ass:On-policy-derivative}-\ref{local-concave},
Consider the sequence $\{\theta_{k}\}_{k\ge0}$ generated by (\ref{vpg}). For any $\delta\in(0,1)$, let the step-size $\alpha$ and the stopping time $\kappa_0$ statisfy 
\[\alpha\leq\min\Big\{\delta,\dfrac{1}{\zeta},\dfrac{\zeta}{\ell^{2}},\dfrac{\zeta\varrho^2}{3\sigma^2}\Big\}, ~~~\alpha\log\dfrac{1}{\alpha}\leq\dfrac{2\zeta\varrho^4}{27\Big(\dfrac{G^{2}R_{\max}^{2}}{(1-\gamma)^{2}}+\zeta\varrho^{2}+\sigma^2\Big)^2},~~~\kappa_{0}=\Big\lfloor
\dfrac{1}{\alpha^2}\log\dfrac{1}{\delta}
\Big\rfloor,
\]
and if some iteration $\theta_{k}$ falls into the ball $\mathbb{B}_{2}(\theta_{\star},\dfrac{\sqrt{3}}{3}\varrho)\subset \mathbb{B}_{2}(\theta_{\star},\varrho) $, i.e., $\big\|\theta_{k}-\theta_{\star}\big\|_{2}^{2}\leq\dfrac{1}{3}\varrho^2$. Then, with probability at least $1-\delta\log\dfrac{1}{\delta}$, for all $j\in[0,\kappa_0-1]$, we have
$\big\|\theta_{k+j}-\theta_{\star}\big\|_{2}\leq\varrho$, i.e.,
\[
\mathbb{P}\Big(\big\|\theta_{k+j}-\theta_{\star}\big\|_{2}\leq\varrho\Big)\ge 1-\delta\log\dfrac{1}{\delta}.
\]}

Recall for each $k\in\mathbb{N}$, $\tau_{k}\sim\pi_{\theta_k}$, to simplify expression, we introduce a notation as follows,
\[g(\theta_k)=:g(\theta_k|\tau_k)=\sum_{t=0}^{h}\nabla_{\theta}\log\pi_{{\theta}}(a_{t}|s_{t})R(\tau_k)|_{\theta=\theta_k}\]
Then, the update (\ref{vpg}) can be rewritten as follows,
\[
\theta_{k+1}=\theta_{k}+\alpha g(\theta_k).
\]

\begin{lemma}[Boundedness of the near iteration]
\label{lemma:Boundedness of the near iteration}
Consider the sequence $\{\theta_{k}\}_{k\in\mathbb{N}}$ generated according to (\ref{vpg}),i.e.,
$
\theta_{k+1}=\theta_{k}+\alpha g(\theta_k).
$
Then, 
\begin{flalign}
\label{boundedness-of-near-iteration}
\big\|\theta_{k+1}-\theta_{k}\big\|_{2}\leq\alpha\big\|g(\theta_{k+1})-g(\theta_{k})\big\|_{2}.
\end{flalign}
\end{lemma}
Lemma \ref{lemma:Boundedness of the near iteration} is a direct result of Lemma 2 in \citep{ghadimi2016mini}. We omit its proof.

We use $\mathcal{E}_{[k:k+t]}$ to denote the event that the element of the sequence $\{\theta_{j}\}_{j=k}^{k+t}$ falls into the ball $\mathbb{B}_{2}(\theta_{\star},\varrho)$,
and use $\kappa_0$ to denote its stopping time, i.e.,
\begin{flalign}
\nonumber
\mathcal{E}_{[k:k+t]}&=\bigcap_{j=0}^{t}\Big\{\theta_{k+j}:\big\|\theta_{k+j}-\theta_{\star}\big\|_2\leq \varrho\Big\},\\
\label{def:e-t-kappa-0}
\kappa_{0}&=\inf_{j\ge0}
\big\{j:\big\|\theta_{k+j}-\theta_{\star}\big\|_2> \varrho\big\}.
\end{flalign}
From the definition of (\ref{def:e-t-kappa-0}), we notice two following basic facts: 

\ding{182} If $t>\kappa_0$, the event $\mathcal{E}_{[k:k+t]}$ never happens, i.e., $\mathcal{E}_{[k:k+t]}=\varnothing$ after the time $\kappa_0$; For each $0\leq j<\kappa_0$, the point $\theta_{k+j}$ falls into the ball $\mathbb{B}_{2}(\theta_{\star},\varrho)$.

\ding{183} For each $t$, we have $\mathcal{E}_{[k:k+t+1]}\subset\mathcal{E}_{[k:k+t]}$.

\begin{lemma}
\label{boundedness-of-z-j-lemma}
We consider the term
$Z_j =\max\Big\{(1-\alpha\zeta)^{-j} \big(\big\|\theta_{k+j}-\theta_{\star}\big\|_{2}^{2}-\frac{\alpha\sigma^2}{\zeta}\big),0 \Big\},$
 then for each $j\in[0,\kappa_0-1]$, the following holds
\begin{flalign}
\label{app-proof-pro1-3}
|Z_{j+1}-Z_{j}|\leq \underbrace{(1-\alpha\zeta)^{-(j+1)}\Big(\alpha^2{2G^{2}R_{\max}^{2}}/{(1-\gamma)^{2}}+\alpha\zeta\varrho^{2}+\alpha^{2}\sigma^2\Big)}_{=:c_{j+1}},
\end{flalign}
i.e., we have
$
|Z_{j+1}-Z_{j}|\leq c_{j+1}.
$
\end{lemma}
Lemma \ref{boundedness-of-z-j-lemma} illustrates the boundedness of $Z_{j+1}-Z_{j}$, 
which is helpful for us to use Azuma's inequality to achieve more refined results later.

\begin{lemma}
\label{lemma-bound-the-kapp-gp-0}
Consider the sequence $\{\theta_{k}\}_{k\in\mathbb{N}}$ generated according to (\ref{vpg}),i.e.,
$
\theta_{k+1}=\theta_{k}+\alpha g(\theta_k).
$ 
Then,
for the term $\theta_{k+\kappa_0}-\theta_{\star}$, the following holds
\[\mathbb{P}\Big(\big\|\theta_{k+\kappa_0-1}-\theta_{\star}\big\|_{2}^{2}\ge\varrho^2\Big)\leq\alpha^2\delta.\]
\end{lemma}

\subsection{Some Preliminary Results}
Before we give the details of the proof of Lemma \ref{boundedness-of-z-j-lemma}, Lemma \ref{lemma-bound-the-kapp-gp-0} and Proposition \ref{propo-1},  we present some preliminary results.

Let
\begin{flalign}
\label{def:xi_k}
\xi_{k}=g(\theta_k)-\nabla J(\theta_k),
\end{flalign}
By the fact $\nabla J(\theta_k)=\mathbb{E}[g(\theta_k)]$, we have $\mathbb{E}[\xi_k]=0.$

We use $\mathcal{F}_{[k,k+t]}=\sigma(\xi_k,\cdots,\xi_{k+t})$ to denote the $\sigma$-field generated by the information from time $k$ to $k+t$: $\{\xi_k,\cdot\cdot\cdot,\xi_{k+t}\}$.

 %and $\bm{1}_(\mathcal{E}_{[k:k+t+1]})\leq\bm{1}_(\mathcal{E}_{[k:k+t]})$,
%where $\bm{1}_(\mathcal{E})$ is indicator function for the event $\mathcal{E}$, i.e., $\bm{1}_(\mathcal{E})=1$ if $\mathcal{E}$ occurs and $\bm{1}_(\mathcal{E})=0$  otherwise.

Recall the update rule (\ref{vpg}), let $\alpha_{k}=\alpha\leq\dfrac{\zeta}{\ell^2}$, we analyze the $(k+\kappa_0)$-th iteration as follows,
\begin{flalign}
\nonumber
&\theta_{k+\kappa_0}= \theta_{k+\kappa_0-1}+\alpha g(\theta_{k+\kappa_0-1})\\
\nonumber
=&\theta_{k+\kappa_0-1}+\alpha\Big(\underbrace{g(\theta_{k+\kappa_0-1})-\nabla J(\theta_{k+\kappa_0-1})}_{=\xi_{k+\kappa_0-1}; \text{Eq.(\ref{def:xi_k})}}+\nabla J(\theta_{k+\kappa_0-1})\Big),\\
\label{app:def-udate-theta-k}
=&\theta_{k+\kappa_0-1}+\alpha\Big(\nabla J(\theta_{k+\kappa_0-1})+\xi_{k+\kappa_0-1}\Big).
\end{flalign}

Under Assumption \ref{local-concave},
$J(\theta)$ is $\zeta$-strongly concave in $\mathbb{B}(\theta_{\star},\varrho)$, i.e.,
\begin{flalign}
\label{app-a-strong-cove}
\nabla J(\theta^{'})\leq J(\theta)+\nabla J(\theta)^{\top}(\theta^{'}-\theta)-\frac{\zeta}{2}\|\theta^{'}-\theta\|_{2}^{2}, ~~\forall~\theta,\theta^{'}\in\mathbb{B}_{2}(\theta_{\star},\varrho ).
\end{flalign}

Since $\theta_{k}$ falls into the ball $\mathbb{B}_{2}(\theta_{\star},\dfrac{\sqrt{3}}{3}\varrho)\subset \mathbb{B}_{2}(\theta_{\star},\varrho) $,
recall $\kappa_{0}=\inf_{j\ge0}
\big\{j:\big\|\theta_{k+j}-\theta_{\star}\big\|_2> \varrho\big\}$, thus the point $\theta_{k+\kappa_0-1}$ also falls into the ball $\mathbb{B}(\theta_{\star},\varrho)$, according to (\ref{app-a-strong-cove}), we have
\begin{flalign}
\nonumber
J(\theta_{\star})-J(\theta_{k+\kappa_0-1})&\leq \nabla J(\theta_{k+\kappa_0-1})^{\top} (\theta_{\star}-\theta_{k+\kappa_0-1})-\dfrac{\zeta}{2}\big\|\theta_{k+\kappa_0-1}-\theta_{\star}\big\|_{2}^{2},\\
\nonumber
J(\theta_{k+\kappa_0-1})-J(\theta_{\star})&\leq \underbrace{\nabla J(\theta_{\star})^{\top}}_{=0} (\theta_{k+\kappa_0-1}-\theta_{\star})-\dfrac{\zeta}{2}\big\|\theta_{\star}-\theta_{k+\kappa_0-1}\big\|_{2}^{2},
\end{flalign}
which implies
\begin{flalign}
\label{gap-theta-*}
\zeta\big\|\theta_{k+\kappa_0-1}-\theta_{\star}\big\|_{2}^{2}\leq\nabla J\big(\theta_{k+\kappa_0-1}\big)^{\top} \big(\theta_{\star}-\theta_{k+\kappa_0-1}\big).
\end{flalign}
Under Assumption \ref{ass:On-policy-derivative}, recall the result of (\ref{def:H}), we have
\begin{flalign}
\label{def:lipstiz-nabla-J}
\big\|\nabla J(\theta_{k+\kappa_0-1})-\nabla J(\theta_{\star})\big\|_2=\big\|\nabla J(\theta_{k+\kappa_0-1})\big\|_2\leq \ell \big\|\theta_{k+\kappa_0-1}-\theta_{\star}\big\|_2.
\end{flalign}

\subsection{Proof of Lemma \ref{boundedness-of-z-j-lemma}}
\begin{proof}(of Lemma \ref{boundedness-of-z-j-lemma}).

Recall $\mathcal{F}_{[k,k+t]}=\sigma(\xi_k,\cdots,\xi_{k+t})$ is the $\sigma$-field generated by the information from time $k$ to $k+t$.
Firstly, we turn to analyze the expected gap between $\theta_{k+\kappa_0}$ and $\theta_{\star}$ as follows,
\begin{flalign}
\nonumber
&\mathbb{E}\Big[\big\|\theta_{k+\kappa_0}-\theta_{\star}\big\|_{2}^{2}\Big|\mathcal{F}_{[k:k+\kappa_0-1]}\Big]\\
\nonumber
\overset{(\ref{app:def-udate-theta-k})}=&\mathbb{E}\Big[\Big\|\theta_{k+\kappa_0-1}+\alpha\Big(\nabla J(\theta_{k+\kappa_0-1})+\xi_{k+\kappa_0-1}\Big)-\theta_{\star}\Big\|_{2}^{2}\Big|\mathcal{F}_{[k:k+\kappa_0-1]}\Big]\\
\nonumber
=&\big\|\theta_{k+\kappa_0-1}-\theta_{\star}\big\|_{2}^{2}+\alpha^{2}\underbrace{\mathbb{E}\Big[\Big\|\xi_{k+\kappa_0-1}\Big\|_{2}^{2}\Big|\mathcal{F}_{[k:k+\kappa_0-1]}\Big]}_{\leq \sigma^{2};\text{Eq.(\ref{def:sigma})}}\\
\nonumber
&~~~~~~~+\alpha^{2}\underbrace{\big\|\nabla J(\theta_{k+\kappa_0-1})\big\|_{2}^{2}}_{\overset{(\ref{def:lipstiz-nabla-J})}\leq \ell^{2} \big\|\theta_{k+\kappa_0-1}-\theta_{\star}\big\|^{2}_2}
+2\alpha
\underbrace{\langle\nabla J(\theta_{k+\kappa_0-1}),\theta_{k+\kappa_0-1}-\theta_{\star}\rangle}_{\overset{(\ref{gap-theta-*})}\leq -\zeta\big\|\theta_{k+\kappa_0-1}-\theta_{\star}\big\|_{2}^{2}}\\
\label{error-gap-up}
\leq&(1+\alpha^{2}\ell^2-2\alpha\zeta)\big\|\theta_{k+\kappa_0-1}-\theta_{\star}\big\|_{2}^{2}+\alpha^{2}\sigma^2\\
\label{iteration-gap}
\leq&(1-\alpha\zeta)\big\|\theta_{k+\kappa_0-1}-\theta_{\star}\big\|_{2}^{2}+\alpha^{2}\sigma^2,
\end{flalign}
Eq.(\ref{iteration-gap}) holds since $\alpha\leq\dfrac{\zeta}{\ell^2}$. 

Furthermore, rearranging Eq.(\ref{iteration-gap}), we have 
\begin{flalign}
\label{app-proof-pro1-1}
\mathbb{E}\Big[\Big\|\theta_{k+\kappa_0}-\theta_{\star}\Big\|_{2}^{2}\Big|\mathcal{F}_{[k:k+\kappa_0-1]}\Big]-\dfrac{\alpha\sigma^2}{\zeta}\leq(1-\alpha\zeta) \Big(\big\|\theta_{k+\kappa_0-1}-\theta_{\star}\big\|_{2}^{2}-\dfrac{\alpha\sigma^2}{\zeta}\Big).
\end{flalign}
Now, , we define a sequence $\{Z_{j}\}^{\kappa_0}_{j=0}$ with its element as follows,
\begin{flalign}
\label{def:Z_t}
Z_j =\max\bigg\{(1-\alpha\zeta)^{-j} \Big(\big\|\theta_{k+j}-\theta_{\star}\big\|_{2}^{2}-\dfrac{\alpha\sigma^2}{\zeta}\Big),0 \bigg\};~~~~0 \leq j\leq \kappa_0.
\end{flalign}

Since $\kappa_{0}=\inf_{j\ge0}
\big\{j:\big\|\theta_{k+j}-\theta_{\star}\big\|_2> \varrho\big\}$, recall $\mathcal{E}_{[k:k+t]}=\cap_{j=0}^{t}\{\theta_{k+j}:\big\|\theta_{k+j}-\theta_{\star}\big\|_2\leq \varrho\}$, which implies 
$\bm{1}_{\mathcal{E}_{[k:k+j]}}=1$, for each $j\in[0,\kappa_0-1]$. Thus,
\begin{flalign}
\label{app-proof-pro1-2}
Z_{j+1}\bm{1}_{\mathcal{E}_{[k:k+j]}}=Z_{j+1}.
\end{flalign}

Furthermore, from the result of (\ref{app-proof-pro1-1}), $Z_{j+1}\bm{1}_{\mathcal{E}_{[k:k+j]}}$ is a a super-martingale, i.e., 
\begin{flalign}
\label{app-proof-pro1-4}
\mathbb{E}\Big[Z_{j+1}\bm{1}_{\mathcal{E}_{[k:k+j]}}\Big]\leq Z_{j}\bm{1}_{\mathcal{E}_{[k:k+j-1]}}.
\end{flalign}

Let's bound on the term $|Z_{j+1}-Z_{j}|\overset{(\ref{app-proof-pro1-2})}=|Z_{j+1}\bm{1}_{\mathcal{E}_{[k:k+j]}}-Z_{j}\bm{1}_{\mathcal{E}_{[k:k+j-1]}}|$, for each $0\leq j\leq \kappa_0 -1$, the following holds
\begin{flalign}
\nonumber
&~~~~~|Z_{j+1}-Z_{j}|\\
\nonumber
&\leq\Bigg|(1-\alpha\zeta)^{-(j+1)}
\bigg(\Big(\big\|\theta_{k+j+1}-\theta_{\star}\big\|_{2}^{2}-\dfrac{\alpha\sigma^2}{\zeta}\Big)-(1-\alpha\zeta)\Big(\big\|\theta_{k+j}-\theta_{\star}\big\|_{2}^{2}-\dfrac{\alpha\sigma^2}{\zeta}\Big)\Bigg)\Bigg|\\
\nonumber
&=(1-\alpha\zeta)^{-(j+1)}\Big| \big\|\theta_{k+j+1}-\theta_{\star}\big\|_{2}^{2}-\big\|\theta_{k+j}-\theta_{\star}\big\|_{2}^{2}+\alpha\zeta\big\|\theta_{k+j}-\theta_{\star}\big\|_{2}^{2}-\alpha^{2}\sigma^2\Big|
\\
\nonumber
&\leq(1-\alpha\zeta)^{-(j+1)}\bigg(\underbrace{\Big| \big\|\theta_{k+j+1}-\theta_{\star}\big\|_{2}^{2}-\big\|\theta_{k+j}-\theta_{\star}\big\|_{2}^{2}\Big|}_{\leq\big\|\theta_{k+j+1}-\theta_{k+j}\big\|_{2}^{2}}+\alpha\zeta\big\|\theta_{k+j}-\theta_{\star}\big\|_{2}^{2}+\alpha^{2}\sigma^2
\bigg)\\
\label{app-iter-gap-1}
&\leq(1-\alpha\zeta)^{-(j+1)}\Big( \underbrace{\big\|\theta_{k+j+1}-\theta_{k+j}\big\|_{2}^{2}}_{\overset{\eqref{boundedness-of-near-iteration}}\leq\alpha^2 \big\|g(\theta_{k+j+1})-g(\theta_{k+j})\big\|_{2}^{2}}+\alpha\zeta\underbrace{\big\|\theta_{k+j}-\theta_{\star}\big\|_{2}^{2}}_{\leq \varrho^2}+\alpha^{2}\sigma^2\Big)
\end{flalign}

Under Assumption \ref{ass:On-policy-derivative}, $\{g(\theta_k)\}_{k\ge0}$ is uniformly bounded, i.e., for each $k$, we have
\begin{flalign}
\label{bound-g-k}
\big\|g(\theta_k)\big\|_{2}=\bigg\|\sum_{t\ge0}\underbrace{\nabla_{\theta}\log\pi_{{\theta_k}}(a_{t}|s_{t})}_{\leq G; (\ref{def:F-G})}R(\tau_k)\bigg\|_{2}\leq G\sum_{t\ge0}\gamma^{t}r_{t+1}\leq\dfrac{GR_{\max}}{1-\gamma}.
\end{flalign} 
Thus, we can rewrite (\ref{app-iter-gap-1}) as follows,
\begin{flalign}
\nonumber
|Z_{j+1}-Z_{j}|&\leq (1-\alpha\zeta)^{-(j+1)}\Big(\alpha^2\big\|g(\theta_{k+j+1})\big\|_{2}^{2}+\alpha^2\big\|g(\theta_{k+j})\big\|_{2}^{2}+\alpha\zeta\varrho^{2}+\alpha^{2}\sigma^2\Big)\\
\label{c_j}
&\overset{(\ref{bound-g-k})}\leq (1-\alpha\zeta)^{-(j+1)}\Big(\alpha^2\dfrac{2G^{2}R_{\max}^{2}}{(1-\gamma)^{2}}+\alpha\zeta\varrho^{2}+\alpha^{2}\sigma^2\Big)=:c_{j+1}.
\end{flalign}
This concludes the proof.
\end{proof}

\subsection{
Proof of Lemma \ref{lemma-bound-the-kapp-gp-0}
}

\begin{proof}(of Lemma \ref{lemma-bound-the-kapp-gp-0}).
Recall the results of (\ref{app-proof-pro1-4}) and (\ref{app-proof-pro1-3}).
By Azuma's inequality, consider the sequence $\{Z_{j}\}^{\kappa_0-1}_{j=0}$ defined in Eq.(\ref{def:Z_t}),
recall $c_{j}$ is defined in (\ref{c_j}),
for any $\delta>0$, we have
\[
\mathbb{P}\Big(Z_{\kappa_0-1}-Z_0\ge\delta\Big)\leq\exp
\Big(
-\dfrac{2\delta^2}{\sum_{j= 0}^{\kappa_0-1} c_{j}^{2}}
\Big),
\]
which implies the following holds,
\begin{flalign}
\label{high-prob-1}
\mathbb{P}\Bigg(
Z_{\kappa_0-1}-Z_0\ge\sqrt{\dfrac{1}{2}\sum_{j= 0}^{\kappa_0-1} c_{j}^{2}\Big(\log\dfrac{1}{\delta}+2\log\dfrac{1}{\alpha}\Big)}
\Bigg)\leq\alpha^2\delta.
\end{flalign}
The result of (\ref{high-prob-1}) implies the following holds with probability less than $\alpha^2\delta$,
\begin{flalign}
\nonumber
\big\|\theta_{k+\kappa_0-1}-\theta_{\star}\big\|_{2}^{2}-\frac{\alpha\sigma^2}{\zeta}&\ge(1-\alpha\zeta)^{\kappa_0-1}\sqrt{\frac{1}{2}
\sum_{j= 0}^{\kappa_0-1} c_{j}^{2}
\Big(\log\frac{1}{\delta} 
+2\log\frac{1}{\alpha}\Big)}
\\
\label{app-iteration-gap}
&~~~~~~~~~~~+(1-\alpha\zeta)^{\kappa_0-1}
\Big(\big\|\theta_{k}-\theta_{\star}\big\|_{2}^{2}-\frac{\alpha\sigma^2}{\zeta}\Big).
\end{flalign}
Rearranging Eq.(\ref{app-iteration-gap}),we have 
\begin{flalign}
\nonumber
\big\|\theta_{k+{\kappa_0-1}}-\theta_{\star}\big\|_{2}^{2}\ge&
\underbrace{(1-\alpha\zeta)^{{\kappa_0-1}}\sqrt{\dfrac{1}{2}
\sum_{j= 0}^{{\kappa_0-1}} c_{j}^{2}
\Big(\log\dfrac{1}{\delta} +2\log\dfrac{1}{\alpha}\Big)} }_{I_1}
\\
\label{app-iteration-gap-rearranging}
&+\underbrace{\big(1-\alpha\zeta\big)^{{\kappa_0-1}}
\big\|\theta_{k}-\theta_{\star}\big\|_{2}^{2}}_{I_2}+\underbrace{\big(1-(1-\alpha\zeta)^{{\kappa_0-1}}\big)\dfrac{\alpha\sigma^2}{\zeta}}_{I_3}.
\end{flalign}
We need to give the conditions to achieve the boundedness of the terms $I_1,I_2,I_3$ defined in Eq.(\ref{app-iteration-gap-rearranging}).

\underline{Boundedness of $I_1$ in Eq.(\ref{app-iteration-gap-rearranging}).}
Consider $c_j$ in (\ref{c_j}), i.e., $c_j= (1-\alpha\zeta)^{-j}\Big(\alpha^2{2G^{2}R_{\max}^{2}}/{(1-\gamma)^{2}}+\alpha\zeta\varrho^{2}+\alpha^{2}\sigma^2\Big),$
and with the condition $\alpha\leq\delta$, we have
\begin{flalign}
\nonumber
I_1&=
(1-\alpha\zeta)^{{\kappa_0-1}}\sqrt{\dfrac{1}{2}
\sum_{j= 0}^{{\kappa_0-1}} c_{j}^{2}
\log\dfrac{1}{\delta}} \\
\nonumber
&=\Big(\alpha^2\dfrac{2G^{2}R_{\max}^{2}}{(1-\gamma)^{2}}+\alpha\zeta\varrho^{2}+\alpha^{2}\sigma^2\Big)
\sqrt{\dfrac{1}{2}
\sum_{j= 0}^{{\kappa_0-1}} (1-\alpha\zeta)^{2({\kappa_0-1})-2j}}\sqrt{
\log\dfrac{1}{\delta}+2\log\dfrac{1}{\alpha}} \\
\nonumber
&\leq\Big(\alpha^2\dfrac{2G^{2}R_{\max}^{2}}{(1-\gamma)^{2}}+\alpha\zeta\varrho^{2}+\alpha^{2}\sigma^2\Big)
\sqrt{\dfrac{1}{2}
\sum_{j= 0}^{{\kappa_0-1}} (1-\alpha\zeta)^{2j}}\sqrt{3
\log\dfrac{1}{\alpha}} 
\\
\nonumber
&\leq\Big(\alpha^2\dfrac{2G^{2}R_{\max}^{2}}{(1-\gamma)^{2}}+\alpha\zeta\varrho^{2}+\alpha^{2}\sigma^2\Big)
\sqrt{\dfrac{1}{2}
\sum_{j= 0}^{\infty} (1-\alpha\zeta)^{2j}}\sqrt{
3\log\dfrac{1}{\alpha}}
\\
\nonumber
&=\Big(\alpha^2\dfrac{2G^{2}R_{\max}^{2}}{(1-\gamma)^{2}}+\alpha\zeta\varrho^{2}+\alpha^{2}\sigma^2\Big)
\sqrt{
\dfrac{
3\log\dfrac{1}{\alpha}}
{2\underbrace{\Big(1-(1-\alpha\zeta)^2\Big)}_{\ge \alpha\zeta}}
} \\
\label{app-a-ne-1}
&\leq\dfrac{\Big(\alpha^2\dfrac{2G^{2}R_{\max}^{2}}{(1-\gamma)^{2}}+\alpha\zeta\varrho^{2}+\alpha^{2}\sigma^2\Big)}{\sqrt{2\alpha\zeta}}
\sqrt{3\log\dfrac{1}{\alpha}}
\\
\label{app-a-ne-2}
&\leq\dfrac{\Big(\alpha\dfrac{2G^{2}R_{\max}^{2}}{(1-\gamma)^{2}}+\alpha\zeta\varrho^{2}+\alpha\sigma^2\Big)}{\sqrt{2\alpha\zeta}}
\sqrt{3\log\dfrac{1}{\alpha}}=\dfrac{\sqrt{\alpha}\Big(\dfrac{2G^{2}R_{\max}^{2}}{(1-\gamma)^{2}}+\zeta\varrho^{2}+\sigma^2\Big)}{\sqrt{2\zeta}}
\sqrt{3\log\dfrac{1}{\alpha}},
\end{flalign}
where Eq.(\ref{app-a-ne-1}) holds since $\alpha\zeta\in(0,1)$, then
$1-(1-\alpha\zeta)^{2}=2\alpha\zeta-(\alpha\zeta)^2=\alpha\zeta(2-\alpha\zeta)\ge\alpha\zeta;$
Eq.(\ref{app-a-ne-2}) holds since $\alpha\in(0,1)$.

We turn to find the condition satisfies
\begin{flalign}
\label{app-proof-pro1-6}
I_1=
\dfrac{\sqrt{\alpha}\Big(\dfrac{G^{2}R_{\max}^{2}}{(1-\gamma)^{2}}+\zeta\varrho^{2}+\sigma^2\Big)}{\sqrt{2\zeta}}
\sqrt{3\log\dfrac{1}{\alpha}}\leq \dfrac{1}{3}\varrho^{2},
\end{flalign}
which requires the step-size should stisfy
\begin{flalign}
\label{step-size-1}
\alpha\log\dfrac{1}{\alpha}\leq\dfrac{2\zeta\varrho^4}{27\Big(\dfrac{G^{2}R_{\max}^{2}}{(1-\gamma)^{2}}+\zeta\varrho^{2}+\sigma^2\Big)^2}.
\end{flalign}
It is noteworthy that since $\lim_{\alpha\rightarrow 0^{+}}\alpha\log\frac{1}{\alpha}=0$, then for an enough small $\alpha$, the condition (\ref{step-size-1}) alway holds.

\underline{Boundedness of $I_2$ in Eq.(\ref{app-iteration-gap-rearranging}).}  In fact,
$
I_2=\big(1-\alpha\zeta\big)^{{\kappa_0-1}}
\big\|\theta_{k}-\theta_{\star}\big\|_{2}^{2}\leq\big\|\theta_{k}-\theta_{\star}\big\|_{2}^{2}\leq\dfrac{1}{3}\varrho^2.
$

\underline{Boundedness of $I_3$ in Eq.(\ref{app-iteration-gap-rearranging}).} In fact, let's consider
$
I_3=\big(1-(1-\alpha\zeta)^{{\kappa_0-1}}\big)\dfrac{\alpha\sigma^2}{\zeta}\leq \dfrac{\alpha\sigma^2}{\zeta}\leq\dfrac{1}{3}\varrho^2,
$
which implies we need the following condition of step-size:
$
\alpha\leq\dfrac{\zeta\varrho^2}{3\sigma^2}.
$

By the results of all the above boundedness of $I_1,I_2,I_3$, under the following condition of step-size
\[\alpha\leq\min\Big\{\delta,\dfrac{1}{\zeta},\dfrac{\zeta}{\ell^{2}},\dfrac{\zeta\varrho^2}{3\sigma^2}\Big\}, ~~~\alpha\log\dfrac{1}{\alpha}\leq\dfrac{2\zeta\varrho^4}{27\Big({G^{2}R_{\max}^{2}}/{(1-\gamma)^{2}}+\zeta\varrho^{2}+\sigma^2\Big)^2},\]
the result of (\ref{app-iteration-gap-rearranging}) implies
\[\mathbb{P}\Big(\big\|\theta_{k+{\kappa_0-1}}-\theta_{\star}\big\|_{2}^{2}\ge\varrho^2\Big)\leq\alpha^2\delta.\]
This concludes the proof.
\end{proof}

\subsection{Proof of Proposition \ref{propo-1}}

\begin{proof}(of Proposition \ref{propo-1})
Let $\overline{\mathcal{E}_{[k:k+t]}}$ be the complementary event of ${\mathcal{E}_{[k:k+t]}}$. 
Recall
\[\mathcal{E}_{[k:k+t]}=\bigcap_{j=0}^{t}\Big\{\theta_{k+j}:\big\|\theta_{k+j}-\theta_{\star}\big\|_2\leq \varrho\Big\},\]
according to the basic law of set theory, we have
\begin{flalign}
\nonumber
\overline{\mathcal{E}_{[k:k+\kappa_0]}}&=\overline{\bigcap_{j=0}^{\kappa_0}\Big\{\theta_{k+j}:\big\|\theta_{k+j}-\theta_{\star}\big\|_2\leq \varrho\Big\}}\\
\nonumber
&=\bigcup_{j=0}^{\kappa_0}\overline{\Big\{\big\|\theta_{k+j}-\theta_{\star}\big\|_2\leq \varrho\Big\}}=\bigcup_{j=0}^{\kappa_0}{\Big\{\big\|\theta_{k+j}-\theta_{\star}\big\|_2> \varrho\Big\}}\\
&=\bigcup_{j=0}^{\kappa_0-1}{\Big\{\big\|\theta_{k+j}-\theta_{\star}\big\|_2> \varrho\Big\}}\bigcup \Big\{\big\|\theta_{k+\kappa_0}-\theta_{\star}\big\|_2> \varrho\Big\}.
\end{flalign}
Then,
\begin{flalign}
\label{iter-recuure}
\mathbb{P}\Big(\overline{\mathcal{E}_{[k:k+\kappa_{0}]}}\Big)&\leq\mathbb{P}\Big(\bigcup_{j=0}^{\kappa_0-1}{\Big\{\big\|\theta_{k+j}-\theta_{\star}\big\|_2> \varrho\Big\}}\Big)+\mathbb{P}\Big(\big\|\theta_{k+\kappa_0}-\theta_{\star}\big\|_2> \varrho\Big)
\leq\mathbb{P}\Big(\overline{\mathcal{E}_{[k:k+\kappa_{0}-1]}}\Big)+\alpha^2\delta.
\end{flalign}
Furthermore, the the condition $\theta_{k}$ falls in the ball $\mathbb{B}_{2}(\theta_{\star},\dfrac{\sqrt{3}}{3}\varrho)$ implies \[\mathbb{P}\Big(\overline{\mathcal{E}_{[k:k]}}\Big)=0;\]
By the result of (\ref{iter-recuure}), we have
\begin{flalign}
\mathbb{P}\Big(\overline{\mathcal{E}_{[k:k+\kappa_{0}]}}\Big)=\sum_{j=1}^{\kappa_0}\bigg(
\mathbb{P}\Big(\overline{\mathcal{E}_{[k:k+j]}}\Big)-\mathbb{P}\Big(\overline{\mathcal{E}_{[k:k+j-1]}}\Big)\bigg)
+\mathbb{P}\Big(\overline{\mathcal{E}_{[k:k]}}\Big)
\leq\kappa_{0}\alpha^2\delta.
\end{flalign}
Let 
$
\kappa_{0}=\Big\lfloor
\dfrac{1}{\alpha^2}\log\dfrac{1}{\delta}
\Big\rfloor+1,
$
then we have 
$
\mathbb{P}\Big[\overline{\mathcal{E}_{[k:k+\kappa_{0}]}}\Big]\leq \delta\log\dfrac{1}{\delta},
$
which implies for each $0\leq j\leq\Big\lfloor
\dfrac{1}{\alpha^2}\log\dfrac{1}{\delta}
\Big\rfloor+1$,  the following holds with probability at least $1-\delta\log\dfrac{1}{\delta}$
\begin{flalign}
\big\|\theta_{k+j}-\theta_{\star}\big\|_{2}^{2}\leq\varrho^{2},
\end{flalign}
 i.e., for each $j\in[0,\kappa_0]$, we have
\[
\mathbb{P}\Big(\big\|\theta_{k+j}-\theta_{\star}\big\|_{2}\leq\varrho\Big)\ge 1-\delta\log\dfrac{1}{\delta}.
\]
This concludes the proof.
\end{proof}

%% file: theorem_1_proof.tex
\clearpage
\section{Proof of Proposition \ref{saddle-point-case}}
\label{app-sec:proof-pro-saddle-point}

\textbf{Proposition} \ref{saddle-point-case}
\emph{
Under Assumption \ref{ass:On-policy-derivative}-\ref{local-concave},
consider the sequence $\{\theta_{k}\}_{k\ge0}$ generated by (\ref{vpg}). If a point $\theta_{k}$ satisfies
\[\lambda_{\max}(\nabla^{2}J(\theta_{k}))\ge\sqrt{\chi\epsilon},\|\nabla J(\theta_k)\|_{2}\leq\epsilon,\]
there exists a positive scalar $\widehat{\kappa}_{0}$:
\[
\widehat{\kappa}_{0}
=
\bigg\lfloor \frac{\log\Big(\frac{1}{1-\sqrt{\alpha}\sigma_{H_0}}\Big)}{\log(1+\alpha \sqrt{\chi\epsilon})}\bigg\rfloor,
~\text{where}~
\sigma_{H_0}= \dfrac{2p\sqrt{p}hR_{\max}(hG^2+L)}{1-\gamma}, 
\]
after at most $j\leq\widehat{\kappa}_{0}$ steps, we have
\begin{flalign}
\nonumber
\mathbb{E}[J(\theta_{k+j})]-J(\theta_{k})
\ge\alpha^{2} \iota^2\sqrt{\chi\epsilon},
\end{flalign}
where $\iota$ is a positive constant defined in (\ref{app-e-it-constant}).
}

\textbf{Notations}~There are many notations in this section
for convenience of reference, we list key notations and constants in the following table.

\begin{tabular}{r c p{11cm}}
$H_0$ &: & $H_0=:\nabla^2 J(\theta_0)$.\\   
$\{\lambda_i\}_{i=1}^{p}$&: & $\lambda_1\leq\lambda_2\leq\cdots\leq\lambda_p$ are the eigenvalues of the matrix $\nabla^{2}J(\theta_0)$.\\
$\Lambda$ &: &  The operator norm of matrix $H_0$, i.e., $\Lambda=\max_{1\leq i\leq p}\{|\lambda_i|\}$.\\
$\widehat{J}(\theta)$ &: &The second order approximation of $J(\theta)$, and it is defined in (\ref{def:hat-J}).\\
$\widehat{H}_0$&: & An estimator of $H_0$, it is defined in (\ref{def:hessin-h-0})\\
$\widehat{\theta}_{k}$&: & The iteration defined in (\ref{sec-update}) that is generated according to problem $\max_{\theta}\widehat{J}(\theta)$.\\
$\xi_{k}$&: &$ \xi_{k}=g(\theta_k)-\nabla J(\theta_k)$ that is defined in (\ref{def:xi_k}).\\
$\widehat{\xi}_{k}$&: &$\widehat{\xi}_{k}=(\widehat{H}_0-H_0)(\widehat{\theta}_{k}-\widehat{\theta}_{0})$.\\
$C_1$&: &$C_1$ is defined in (\ref{app-boundness-xi-1}).\\
$C_2,C_3$&: &$C_2,C_3$ are defined in (\ref{c-2}).\\
$I_1,I_2$&: &They are defined in (\ref{iteration-the-j}).\\
$\widehat{\kappa}_{0}$&: &A positive integer defined in (\ref{hat-kapp-0}).\\
$\Delta H_{k}$&: &$\Delta H_{k}=\nabla^{2}J(\theta_{k})-H_0=\nabla^{2}J(\theta_{k})-\nabla^{2}J(\theta_{0}).$\\
$\Delta_{k}$&: &$\Delta_{k}=\nabla J(\theta_{k})-\nabla\widehat{J}(\widehat{\theta}_{k}).$ \\
$\widehat{\phi},\phi$&: &$\widehat{\phi}=\widehat{\theta}_{k+1}-\theta_{0}=\widehat{\theta}_{k+1}-\widehat{\theta}_{0}$ and $\phi=\theta_{k+1}-\widehat{\theta}_{k+1}$.\\
$\Big\|\frac{{d_{\rho_0}^{\pi_{\theta_*}}}}{\rho_0}\Big\|_{\infty}$&: &$\Big\|\frac{{d_{\rho_0}^{\pi_{\theta_*}}}}{\rho_0}\Big\|_{\infty}=\sup_{s\in\mathcal{S}}\frac{{d_{\rho_0}^{\pi_{\theta_*}}}(s)}{\rho_0(s)}$.\\
$\sigma_{H_0}$&: &It is defined in (\ref{def:boundedness-sigma-0}).\\
$\mathcal{B}_{k},\mathcal{C}_{k}$&: & Two events defined in (\ref{event-1-k}) and (\ref{event-2-k}).\\
$B_1$-$B_4$&: & They are defined in (\ref{app-a-bound-1})-(\ref{app-a-bound-4}).\\
$D_1,D_2$&: & They are defined in (\ref{def:d-1-2}).\\
$E_1,E_2$ &: &They are defined in (\ref{def:constant-E}).\\
$\beta$ &: &
$\beta=\frac{2\sqrt{\alpha\log\frac{1}{\alpha} }D_1B_1+ D_2}{\Lambda^2+2\Lambda}$
\end{tabular}

Without loss of generality, in the proof, we consider the initial $\theta_0$ falls into the region $\mathcal{L}_2$.

\textbf{Organization and Key Ideas in This Section}
It is very technical to achieve the result of Proposition \ref{saddle-point-case}, we outline some necessary intermediate results in the following Section \ref{app-pro3-pre-results}. 
Concretely, Lemma \ref{app-key-lem} and Lemma \ref{app-key-lem-second-1} play a key role in the proof of Proposition \ref{saddle-point-case}. 
We utilize the second order information of the expected return $J(\theta)$ as following two key steps:
\begin{compactenum}[\textbullet]
\item Firstly, let $\widehat{J}(\theta)$ (defined in (\ref{def:hat-J})) be the second order approximation of $J(\theta)$, we consider the following optimization problem
\begin{flalign}
\label{app-second-policy-op}
\max_{\theta}\widehat{J}(\theta),
\end{flalign}
and construct a sequence $\{\widehat{\theta}_{k}\}_{k\ge0}$ (defined in the below Eq.(\ref{sec-update})) to solve the problem (\ref{app-second-policy-op}).
Lemma \ref{app-key-lem} illustrates if an initial iteration $\widehat{\theta}_{0}(=\theta_0)$ falls into the region 
\[\mathcal{L}_{2}=\{\theta\in\mathbb{R}^{p}: \|\nabla J(\theta)\|_2\leq\epsilon\}
\cap\{\theta\in\mathbb{R}^{p}: \lambda_{\max}(\nabla^{2} J(\theta)) \ge \sqrt{\chi\epsilon}\},\]
then the following facts happen: $\widehat{\theta}_{k+1}$ closes to the initial point $\widehat{\theta}_{0}$ for a long time and $\|\nabla \widehat{J}(\widehat{\theta}_{k+1})\|_2$ can be small for  a long time.
Lemma \ref{lem-boundedness-of-h0}-\ref{high-pro-i-02-term-lem} provide some intermediate results for the proof of Lemma \ref{app-key-lem}, we will provide all the details of them step by step in the following Section \ref{proof-app-a-lemmapg-gap-ho-h} to \ref{proof-ap-a-lemmapg-gap-nabla-j-hat}. The proof of Lemma \ref{app-key-lem} lies in Section \ref{proof-of-emma-app-key-lem}.
\end{compactenum}

\begin{compactenum}[\textbullet]
\item Then, under above conditions, we provide Lemma \ref{app-key-lem-second-1}, 
which illustrates if an initial iteration $\widehat{\theta}_{0}$ falls into the region $\mathcal{L}_{2}$, for a proper step-size, the iteration generated according to the problem (\ref{app-second-policy-op}) can be closed to the solution of policy optimization (\ref{Eq:thata-optimal}) with high probability. We present the precise result in (\ref{precise-02}).
Similarly, the policy gradient estimator of the problem (\ref{Eq:thata-optimal}) can be closed to the policy gradient estimator of (\ref{app-second-policy-op}), which is presented in (\ref{precise-01}).
\end{compactenum}

\subsection{Summary of Preliminary Results}
\label{app-pro3-pre-results}

Let $\lambda_1\leq\lambda_2\leq\cdots\leq\lambda_p$ be the eigenvalues of the matrix $\nabla^{2}J(\theta_0)$.
We use $H_0$ to denote the matrix $\nabla^{2}J(\theta_0)$, in this section, we assume the initial point $\theta_{0}$ satisfies
\[\lambda_{\max}(\nabla ^{2}J(\theta_{0}))=\lambda_{\max}(H_0)=\lambda_p\ge\sqrt{\chi\epsilon},~~~~~~~\|\nabla J(\theta_0)\|_{2}\leq\epsilon.\]
Furthermore, by the definition of the operator norm of $H_0$, we have
\[
\|H_0\|_{op}=\max_{1\leq i\leq p}\{| \lambda_i |\}=\max\{{|\lambda_1|, \lambda_p}\}=:\Lambda\ge\sqrt{\chi\epsilon}.
\]
We use $\widehat{J}(\theta)$ to denote the second order approximation of $J(\theta)$, i.e., let $H_0=\nabla^{2}J(\theta_0)\in\mathbb{R}^{p\times p}$,
\begin{flalign}
\label{def:hat-J}
\widehat{J}(\theta)=J(\theta_0)+\nabla J(\theta_{0})^{\top}(\theta-\theta_0)+\dfrac{1}{2}(\theta-\theta_{0})^{\top}H_{0}(\theta-\theta_{0}).
\end{flalign}
Let $\widehat{\tau}_0=\{\widehat{s}^{0}_{t},\widehat{a}^{0}_{t},\widehat{r}^{0}_{t+1}\}_{t=0}^{h}\sim\pi_{\widehat{\theta}_0}$, and we introduce a notation $\widehat{\Phi}(\theta)$ as follows,
\[
\widehat{\Phi}(\theta)=\sum_{t=0}^{h}\sum_{i=t}^{h}\gamma^{i}\widehat{r}^{0}_{i+1}(\widehat{s}^{0}_{i},\widehat{a}^{0}_{i})\log \pi_{\theta}(\widehat{s}^{0}_{h},\widehat{a}^{0}_{h}).
\]
According to the section 7.2 of \citep{shen2019hessian}, we conduct an unbiased estimator of $H_0$ as:
\begin{flalign}
\label{def:hessin-h-0}
\widehat{H}_0=\nabla \widehat{\Phi}(\theta)\nabla\log p(\widehat{\tau}_{0};\pi_{\theta})^{\top}+\nabla^{2} \widehat{\Phi}(\theta)\Big|_{\theta={\widehat{\theta}_0}},
\end{flalign}
where $p(\widehat{\tau}_{0};\pi_{\widehat{\theta}_0})=\rho_{0}(\widehat{s}^{0}_{0})\prod_{t=0}^{h}P({\widehat{s}^{0}_{t+1}|\widehat{s}^{0}_{t},\widehat{a}^{0}_{t})}\pi_{\widehat{\theta}_{0}}(\widehat{a}^{0}_{t}|\widehat{s}^{0}_{t})$ is the probability of generating $\widehat{\tau}_{0}$ according to the policy $\pi_{\widehat{\theta}_0}$.

\begin{lemma}[Boundedness of $\|\widehat{H}_0-H_0\|_{op}$]
\label{lem-boundedness-of-h0}
Let $H_{0}=\nabla^2 J(\theta_0)$, $\widehat{H}_0$ defined in (\ref{def:hessin-h-0}) is an estimator of $H_{0}$.
Under Assumption \ref{ass:On-policy-derivative}-\ref{local-concave}, the following holds
\[
\|\widehat{H}_0-H_0\|_{op}\leq 2\dfrac{p\sqrt{p}hR_{\max}}{1-\gamma}\Big(hG^2+L\Big)=:\sigma_{H_0}.
\]
\end{lemma}

Now, we consider a coupled sequence $\{\widehat{\theta}_{k}\}_{k\ge0}$ is the iteration of policy gradient solution on the second order approximation function $\widehat{J}(\theta)$ defined in (\ref{def:hat-J}), i.e., $\{\widehat{\theta}_{k}\}_{k\ge0}$ solves the problem $\max_{\theta}\widehat{J}(\theta)$ along the direction of a policy gradient estimator of $\nabla\widehat{J}(\theta)$.
Concretely,
we set the initial value of $\widehat{\theta}_{0}$ as the same consideration with policy optimization problem (\ref{Eq:thata-optimal}), i.e.,
$\widehat{\theta}_{0}=\theta_0.$
Then define the update rule of $\widehat{\theta}_{k}$ as follows,
\begin{flalign}
\label{sec-update}
\widehat{\theta}_{0}=\theta_0,~~~~~~~~~~
\widehat{\theta}_{k+1}=\widehat{\theta}_{k} +\alpha\widehat{\nabla\widehat{J}(\widehat{\theta}_{k})},
\end{flalign}
where $\widehat{\nabla\widehat{J}(\cdot)}$ is an estimator of the gradient function $\nabla\widehat{J}(\cdot)$, and $\alpha$ is step-size. 
Taking the gradient of the function $\widehat{J}(\theta)$ (\ref{def:hat-J}), we have
\begin{flalign}
\label{nab-j-re}
\nabla \widehat{J}(\theta)=\nabla J(\theta_0)+H_0 (\theta-\theta_0),
\end{flalign}
for the iteration (\ref{sec-update}), we define an estimator of $\nabla \widehat{J}(\widehat{\theta}_{k})$ as follows,
\begin{flalign}
\label{estimate-hat-gradient}
\widehat{\nabla\widehat{J}(\widehat{\theta}_{k})}=g(\widehat {\theta}_0)+\widehat{H}_{0}(\widehat{\theta}_{k}-\widehat{\theta}_{0}),
\end{flalign}
where $\widehat{H}_{0}$ is defined in (\ref{def:hessin-h-0}).
Recall $\xi_k$ defined in (\ref{def:xi_k}), since $\widehat{\theta}_{0}=\theta_0$,
then $\xi_0=g(\widehat {\theta}_0)-\nabla J(\widehat{\theta}_0)$. Let $\widehat{\xi}_{k}=(\widehat{H}_0-H_0)(\widehat{\theta}_{k}-\widehat{\theta}_{0})$, since $\nabla \widehat{J}(\widehat{\theta}_{k})=\nabla J(\widehat{\theta}_0)+H_0(\widehat{\theta}_{k}-\widehat{\theta}_{0})$,
we can rewrite (\ref{sec-update}): 
\begin{flalign}
\label{sec-update-2}
\widehat{\theta}_{k+1} &\xlongequal{(\ref{sec-update}),(\ref{estimate-hat-gradient})}\widehat{\theta}_{k} +\alpha\Big(g(\widehat {\theta}_0)+\widehat{H}_{0}(\widehat{\theta}_{k}-\widehat{\theta}_{0})\Big)\\
\nonumber
&=\widehat{\theta}_{k} +\alpha\Big(\underbrace{g(\widehat {\theta}_0)-\nabla J(\widehat{\theta}_0)}_{=\xi_0}+\underbrace{\nabla J(\widehat{\theta}_0)+H_0(\widehat{\theta}_{k}-\widehat{\theta}_{0})}_{=\nabla \widehat{J}(\widehat{\theta}_{k})}+\underbrace{(\widehat{H}_0-H_0)(\widehat{\theta}_{k}-\widehat{\theta}_{0})}_{=\widehat{\xi}_{k}}\Big)\\
\label{sec-update-1}
&=\widehat{\theta}_{k} +\alpha\Big(\nabla\widehat{J}(\widehat{\theta}_{k})+\widehat{\xi}_{k}+\xi_0\Big).
\end{flalign}
Staring from (\ref{sec-update-1}), after some careful calculations, we have
\begin{flalign}
\label{iteration-the-j}
\widehat{\theta}_{k+1}-\widehat{\theta}_{0}&=\alpha\underbrace{\sum_{j=0}^{k}(I+\alpha H_0)^{j}\nabla{J}(\widehat{\theta}_0)}_{=:I_2}+\alpha\underbrace{\sum_{j=0}^{k}(I+\alpha H_0)^{k-j}\big(\widehat{\xi}_{j}+\xi_0\big)}_{=:I_1},\\
\label{iteration-nabla-j}
\nabla \widehat{J}(\widehat{\theta}_{k+1})&=(I+\alpha H_0)^{k}\nabla J(\theta_0)+\alpha\underbrace{\sum_{j=0}^{k}(I+\alpha H_0)^{k-j}\big(\widehat{\xi}_{j}+\xi_0\big)}_{=:I_1}.
\end{flalign}

\begin{lemma}[Boundedness of Term $I_1$ with High Probability]
\label{high-pro-i-1-term-lem}
Under the conditions of Assumption \ref{ass:On-policy-derivative}-\ref{local-concave} and Proposition \ref{saddle-point-case}, consider the term $I_1=\sum_{j=0}^{k}(I+\alpha H_0)^{k-j}\big(\widehat{\xi}_{j}+\xi_0\big)$ defined in (\ref{iteration-the-j})-(\ref{iteration-nabla-j}).
Then, there exists a positive integer $\widehat{\kappa}_{0}$, such that
for each $k\in[0,\widehat{\kappa}_{0}]$, for any $\delta\in(0,1)$, we have
\[
\mathbb{P}\Bigg(
\Big\|\sum_{j=0}^{k}(I+\alpha H_0)^{k-j}\big(\widehat{\xi}_{j}+\xi_0\big)\Big\|_2\leq 2C_1\sqrt{\widehat{\kappa}_{0}\log{\dfrac{4}{\delta}}}\Bigg)\ge1-\delta,
\]
where $C_1$ is defined in (\ref{app-boundness-xi-1}).
\end{lemma}

\begin{lemma}[Boundedness of Term $I_2$ with High Probability]
\label{high-pro-i-02-term-lem}
Under the conditions of Assumption \ref{ass:On-policy-derivative}-\ref{local-concave} and Proposition \ref{saddle-point-case}, consider the term $I_2=\sum_{j=0}^{k}(I+\alpha H_0)^{j}\nabla{J}(\widehat{\theta}_0)$ defined in (\ref{iteration-the-j}).
Then, there exists a positive integer $\widehat{\kappa}_{0}$, such that
for each $k\in[0,\widehat{\kappa}_{0}]$, we have
\[
\big\|I_2\big\|_{2}=\Big\|\sum_{j=0}^{k}(I+\alpha H_0)^{j}\nabla{J}(\widehat{\theta}_0)\Big\|_{2}
\leq\dfrac{\epsilon}{1-\sqrt{\alpha}\sigma_{H_0}}.
\]
\end{lemma}
From the results of Lemma \ref{high-pro-i-1-term-lem} and Lemma \ref{high-pro-i-02-term-lem}, combine the result of (\ref{iteration-the-j})-(\ref{iteration-nabla-j}), we have the boundedness of $\|\widehat{\theta}_{k+1}-\widehat{\theta}_{0}\big\|_2$ and $\|\nabla \widehat{J}(\widehat{\theta}_{k+1})\|_2$ as following Lemma \ref{app-key-lem}.
\begin{lemma}
\label{app-key-lem}
Under Assumption \ref{ass:On-policy-derivative}-\ref{local-concave},
consider the sequence $\{\theta_{k}\}_{k\ge0}$ generated by (\ref{vpg}), $\{\widehat{\theta}_{k}\}_{k\ge0}$ generated by (\ref{sec-update}), and the initial point satisfies $\theta_{0}=\widehat{\theta}_{0}$.
Let $H_0=\nabla ^{2}J(\theta_{0})$,
the initial point $\theta_0$ also satisfies
$\lambda_{\max}(H_0)\ge\sqrt{\chi\epsilon},\|\nabla J(\theta_0)\|_{2}\leq\epsilon$.
Let $\widehat{J}(\theta)$ (\ref{def:hat-J}) be the second order approximation of the expected return $J(\theta)$.
Let 
\[\sigma_{H_0}= \frac{2p\sqrt{p}hR_{\max}(hG^2+L)}{1-\gamma}, 
\alpha<\min\Big\{\dfrac{1}{\sigma_{H_0}},\dfrac{1}{\sigma^{2}_{H_0}}\Big\},
\bigg\lfloor \frac{\log\big(\frac{1}{1-\sqrt{\alpha}\sigma_{H_0}}\big)}{\log(1+\alpha \sqrt{\chi\epsilon})}\bigg\rfloor
=:\widehat{\kappa}_{0}.\]
For each $k\in[0,\widehat{\kappa}_{0}]$, for any $\delta\in(0,1)$, we have
\[
\mathbb{P}\bigg(
\big\|\widehat{\theta}_{k+1}-\widehat{\theta}_{0}\big\|_2\leq\alpha\Big(2C_1\sqrt{ \widehat{\kappa}_{0}\log{\dfrac{4}{\delta}}}+
\dfrac{\epsilon}{1-\sqrt{\alpha}\sigma_{H_0}}
\Big)
\bigg)\ge1-\delta.
\]
\[
\mathbb{P}\bigg(
\big\|\nabla \widehat{J}(\widehat{\theta}_{k+1})\big\|_2\leq
 2\alpha C_1\sqrt{ \widehat{\kappa}_{0}\log{\dfrac{4}{\delta}}}+
\dfrac{\epsilon}{1-\sqrt \alpha \sigma_{H_0}}
\bigg)\ge1-\delta.
\]
where $C_1=\frac{1}{1-\sqrt{\alpha} \sigma_{H_0}}\Big(
  { \frac{\sigma_{H_0}G R_{\max}}{1-\gamma}}
  \big(
  \frac{1}{\sqrt{\chi\epsilon}(1-\sqrt{\alpha} \sigma_{H_0})}+\frac{\alpha}{1-\alpha \sigma_{H_0}}
  \big)+\sigma
  \Big)$.
\end{lemma}

Now, we turn to consider the term $\nabla J(\theta_{k})-\nabla\widehat{J}(\widehat{\theta}_{k})=:\Delta_{k}$ and $\theta_{k+1}-\widehat{\theta}_{k+1}$.
We give the partition of $\Delta_{k}$ and $\theta_{k+1}-\widehat{\theta}_{k+1}$ as below.

Let's calculate the policy gradient $\nabla J(\theta_{k+1})$ as follows,
\begin{flalign}
\nonumber
\nabla J(\theta_{k+1})&=\nabla J(\theta_{k})+\nabla^{2}J(\theta_{k})(\theta_{k+1} -\theta_{k})+e_{k}\\
\label{iteration-nabla-j-1}
&=(1+\alpha H_0)\nabla J(\theta_{k})+\alpha H_0(\widehat{\xi}_{k}+\xi_0)+\alpha\Delta H_{k}(\nabla J(\theta_{k})+\widehat{\xi}_{k}+\xi_0)+e_{k},
\end{flalign}
where 
\[e_{k}=\int_{0}^{1}\big[\nabla^{2}J(\theta_{k}+x(\theta_{k+1}-\theta_{k}))-\nabla^{2}J(\theta_{k})\big]\text{d}(\theta_{k+1}-\theta_{k})x,\]
\[\Delta H_{k}=\nabla^{2}J(\theta_{k})-H_0=\nabla^{2}J(\theta_{k})-\nabla^{2}J(\theta_{0}).\]
Furthermore, let $\Delta_{k}=\nabla J(\theta_{k})-\nabla\widehat{J}(\widehat{\theta}_{k})$, then according to (\ref{iteration-nabla-j}) and (\ref{iteration-nabla-j-1}), after some careful calculations, we have the following partition of the term $\nabla J(\theta_{k+1})-\nabla\widehat{J}(\widehat{\theta}_{k+1})$ and $\theta_{k+1}-\widehat{\theta}_{k+1}$
\begin{flalign}
\nonumber
\Delta_{k+1}&=\nabla J(\theta_{k+1})-\nabla\widehat{J}(\widehat{\theta}_{k+1})\\
\label{iterdelta}
&=(I+\alpha H_0)\Delta_{k}+\alpha\Delta H_{k}\big(\Delta_{k}+\nabla\widehat{J}(\widehat{\theta}_{k})\big)+\alpha\Delta H_{k}\big(\widehat{\xi}_{k}+\xi_0\big)+e_{k}\\
\label{itertheta}
\theta_{k+1}&=\widehat{\theta}_{k+1}+\alpha\sum_{j=0}^{k}\Delta_{j}
\end{flalign}
Finally, we need the boundedness of $\|\nabla J(\theta_k)\|_{2}$ to prove the Lemma \ref{app-key-lem-second-1}, we present it below.
\begin{lemma}
\label{boundedness-j-the-k}
The term $\|\nabla J(\theta_k)\|_{2}$ is upper-bounded as follows,
\[
\|\nabla J(\theta_k)\|_{2}\overset{(\emph{\textbf{a}})}=\|\nabla J(\theta_{\star})-\nabla J(\theta_k)\|_{2}
\leq
\Big\|\dfrac{{d_{\rho_0}^{\pi_{\theta_{\star}}}}}{\rho_0}\Big\|_{\infty}\dfrac{(1+\gamma |\mathcal{S}|)|\mathcal{A}|^2 |\mathcal{S}| R_{\max}}{(1-\gamma)^2},
\]
where $\Big\|\dfrac{{d_{\rho_0}^{\pi_{\theta_{\star}}}}}{\rho_0}\Big\|_{\infty}=\max_{s\in\mathcal{S}}\dfrac{d_{\rho_0}^{\pi_{\theta_{\star}}}(s)}{{\rho_0(s)}}$, Eq.\emph{(\textbf{a})} holds since $\nabla J(\theta_{\star})=0$.
\end{lemma}

\begin{lemma}
\label{app-key-lem-second-1}
Under Assumption \ref{ass:On-policy-derivative}-\ref{local-concave},
consider the sequence $\{\theta_{k}\}_{k\ge0}$ generated by (\ref{vpg}), $\{\widehat{\theta}_{k}\}_{k\ge0}$ generated by (\ref{sec-update}), and the initial point satisfies $\theta_{0}=\widehat{\theta}_{0}$.
Let $H_0=\nabla ^{2}J(\theta_{0})$,
the initial point $\theta_0$ also satisfies
$\lambda_{\max}(H_0)\ge\sqrt{\chi\epsilon},\|\nabla J(\theta_0)\|_{2}\leq\epsilon$.
Let $\widehat{J}(\theta)$ (\ref{def:hat-J}) be the second order approximation of the expected return $J(\theta)$.
Let $\sigma_{H_0}= \frac{2p\sqrt{p}hR_{\max}(hG^2+L)}{1-\gamma}$,
$\alpha<\frac{1}{\sigma_{H_0}},  \bigg\lfloor \frac{\log\big(\frac{1}{1-\sqrt{\alpha}\sigma_{H_0}}\big)}{\log(1+\alpha \sqrt{\chi\epsilon})}\bigg\rfloor+1
=:\widehat{\kappa}_{0}.$
For each $k\in[0,\widehat{\kappa}_{0}]$, the following holds,
\begin{flalign}
\label{precise-01}
\mathbb{P}\Bigg(\max \Big\{\|\Delta_k\|_2,
\|\Delta_k\|^2_2 \Big\}\ge C_4\sqrt{\alpha^3\log\dfrac{1}{\alpha}}\Bigg)\leq\dfrac{\sigma_{H_0}}{\sqrt{\chi\epsilon}}\alpha^{\frac{3}{2}}+o(\alpha^{\frac{3}{2}}),
\end{flalign}
\begin{flalign}
\label{precise-02}
\mathbb{P}\Bigg(
\big\|\theta_{k+1}-\widehat{\theta}_{k+1}\big\|_{2}\leq\alpha^2\sqrt{\log\dfrac{1}{\alpha}}\dfrac{\sigma_{H_0}}{\sqrt{\chi\epsilon}}C_4+o\Big(\alpha^2\sqrt{\log\dfrac{1}{\alpha}}\Big)
\Bigg)\leq\dfrac{\sigma_{H_0}}{\sqrt{\chi\epsilon}}\alpha^{\frac{3}{2}}+o(\alpha^{\frac{3}{2}}).
\end{flalign}
\end{lemma}

\subsection{Proof of Lemma \ref{lem-boundedness-of-h0}}
\label{proof-app-a-lemmapg-gap-ho-h}

\begin{proof}(of Lemma\ref{lem-boundedness-of-h0}).
We need two results of operator norms to show the boundedness of $\|\widehat{H}_0-H_0\|_{op}$. For any matrix $A=(a_{i,j})\in\mathbb{R}^{p\times p},B\in\mathbb{R}^{p\times p}$, the following holds,
\begin{flalign}
\label{app-op-ineq-01}
\|A+B\|_{op}&\leq \|A\|_{op}+\|B\|_{op}, \\
\label{app-op-ineq}
\max_{i,j}\{|a_{i,j}|\}&\leq\|A\|_{op}\leq p\sqrt{p}\max_{i,j}\{|a_{i,j}|\}.
\end{flalign}
For the proof of (\ref{app-op-ineq-01}) and (\ref{app-op-ineq}), please refer to Theorem 3.4 of a lecture provided in 
\begin{center}
\url{https://kconrad.math.uconn.edu/blurbs/linmultialg/matrixnorm.pdf}.
\end{center}

Using the result of of (\ref{app-op-ineq}), we have
\begin{flalign}
\nonumber
\|\widehat{H}_0-H_0\|_{op}\leq  \|\widehat{H}_0\|_{op}+\|H_0\|_{op},
\end{flalign}
to achieve the boundedness of $\|\widehat{H}_0-H_0\|_{op}$, we need to bound $\|\widehat{H}_0\|_{op}$, $\|H_0\|_{op}$ correspondingly.

Recall ${\tau}_0=\{{s}^{0}_{t},{a}^{0}_{t},{r}^{0}_{t+1}\}_{t=0}^{h}\sim\pi_{{\theta}_0}$, let
$
{\Phi}(\theta)=\sum_{t=0}^{h}\sum_{i=t}^{h}\gamma^{i}{r}^{0}_{i+1}({s}^{0}_{i},{a}^{0}_{i})\log \pi_{\theta}({s}^{0}_{h},{a}^{0}_{h}).
$
According to section 7.2 of \citep{shen2019hessian}, the second order derivative of $J(\theta)$ is
\begin{flalign}
\label{shen-SODJ}
\nabla^2 J(\theta)=\mathbb{E}_{\tau\sim\pi_{\theta_0}}\Big[\nabla {\Phi}(\theta)\nabla\log p({\tau}_{0};\pi_{\theta})^{\top}+\nabla^{2} {\Phi}(\theta),\Big]
\end{flalign}
where $p({\tau}_{0};\pi_{{\theta}_0})=\rho_{0}({s}^{0}_{0})\prod_{t=0}^{h}P({{s}^{0}_{t+1}|{s}^{0}_{t},{a}^{0}_{t})}\pi_{{\theta}_{0}}({a}^{0}_{t}|{s}^{0}_{t})$ is the probability of generating ${\tau}_{0}$ according to the policy $\pi_{{\theta}_0}$.
From the result of (\ref{shen-SODJ}), if we get the boundedness of $
\nabla {\Phi}(\theta)\nabla\log p({\tau}_{0};\pi_{\theta})^{\top}+\nabla^{2} {\Phi}(\theta)$, then the boundedness of $\big\|H_{0}\big\|_{op}=\big\|\nabla^2 J(\theta_0)\big\|_{op}$ is clear.

Now, we will bound the operator norm of the matrix $
\nabla {\Phi}(\theta)\nabla\log p({\tau}_{0};\pi_{\theta})^{\top}+\nabla^{2} {\Phi}(\theta)$. We need a simple fact: if $a=(a_1,a_2,\cdots,a_p)^{\top},b=(b_1,b_2,\cdots,b_p)\in\mathbb{R}^{p}$, and $A=ab^{\top}$, then we have
\begin{flalign}
\label{app-a-op-2}
\|A\|_{op}=\|ab^{\top}\|_{op}\overset{(\ref{app-op-ineq})}\leq p\sqrt{p}\max_{i,j}|a_i b_j|\leq p\sqrt{p}\max_{i}|a_i| \cdot\max_{j}| b_j|.
\end{flalign}

The results of (\ref{app-op-ineq}),(\ref{app-a-op-2}) imply that to achieve the boundedness of the operator norm of the matrix $
\nabla {\Phi}(\theta)\nabla\log p({\tau}_{0};\pi_{\theta})^{\top}+\nabla^{2} {\Phi}(\theta)$,
we need to bound the elements of the vectors: $\nabla {\Phi}(\theta), \nabla\log p({\tau}_{0};\pi_{\theta})^{\top}$, and bound each element of the Hessian matrix $\nabla^2 {\Phi}(\theta)$.

Recall Assumption \ref{ass:On-policy-derivative}, for each $1\leq j\leq p$, we have
\[
\Big|[\nabla {\Phi}(\theta)]_{j}\Big|
=\Big |\Big[\sum_{t=0}^{h}\sum_{i=t}^{h}\gamma^{i}{r}^{0}_{i+1}({s}^{0}_{i},{a}^{0}_{i})\underbrace{\nabla \log \pi_{\theta}({s}^{0}_{h},{a}^{0}_{h})}_{\overset{(\ref{def:F-G})}\leq G}\Big]_{j}\Big|\leq\dfrac{hGR_{\max}}{1-\gamma},
\]
\[
\Big|[\nabla\log p({\tau}_{0};\pi_{\theta})]_{j}\Big|
=\Big |\Big[\sum_{t=0}^{h}\rho_{0}({s}^{0}_{0})P({{s}^{0}_{t+1}|{s}^{0}_{t},{a}^{0}_{t})}\underbrace{\nabla \log \pi_{\theta}({s}^{0}_{h},{a}^{0}_{h})}_{\overset{(\ref{def:F-G})}\leq G}\Big]_{j}\Big|\leq hG,
\]
where $[\cdot]_{j}$ denotes the $j$-the coordinate component of a vector.
Combining above two results, we have 
\begin{flalign}
\label{app-PHi}
\|\nabla {\Phi}(\theta)\nabla\log p({\tau}_{0};\pi_{\theta})^{\top}\|_{op}\overset{(\ref{app-a-op-2})}\leq p\sqrt{p}\dfrac{hGR_{\max}}{1-\gamma}\cdot hG
=p\sqrt{p}\dfrac{h^2G^2R_{\max}}{1-\gamma}.
\end{flalign}

Now, we bound on the term $\|\nabla^{2} {\Phi}(\theta)\|_{op}$. Let $[\cdot]_{i,j}$ denotes the $(i,j)$-the coordinate component of a matrix, then for each element of $\|\nabla^{2} {\Phi}(\theta)\|_{op}$, we have
\begin{flalign}
\nonumber
\Big|[\nabla^{2} {\Phi}(\theta)]_{i,j}\big|
=\Big |\sum_{t=0}^{h}\sum_{i=t}^{h}\gamma^{i}{r}^{0}_{i+1}({s}^{0}_{i},{a}^{0}_{i})\underbrace{\dfrac{\partial^{2}}{\partial\theta_i\partial\theta_j}\log\pi_{\theta}({s}^{0}_{h},{a}^{0}_{h}))}_{\overset{(\ref{def:F-G})}\leq F}\Big|\leq\dfrac{hLR_{\max}}{1-\gamma},
\end{flalign}
which implies 
\begin{flalign}
\label{app-boun-H-0}
\|\nabla^{2} {\Phi}(\theta)\|_{op}\overset{(\ref{app-op-ineq})}\leq p\sqrt{p}\dfrac{hLR_{\max}}{1-\gamma}.
\end{flalign}
By the results of (\ref{app-PHi}) and (\ref{app-boun-H-0}), we have
\begin{flalign}
\nonumber
\|H_0\|_{op}=\|\nabla J(\theta_0)\|_{op}&\leq \|\nabla^{2} {\Phi}(\theta)\nabla\log p({\tau}_{0};\pi_{\theta})^{\top}\|_{op}+\|\nabla^{2} {\Phi}(\theta)\|_{op}\\
\nonumber
&\leq \dfrac{p\sqrt{p}hR_{\max}}{1-\gamma}\Big(hG^2+L\Big).
\end{flalign}
As the same analysis with $\|H_0\|_{op}$, we have
$
\|\widehat{H}_0\|_{op}\leq \dfrac{p\sqrt{p}hR_{\max}}{1-\gamma}\Big(hG^2+L\Big).
$
Thus, we have
\begin{flalign}
\label{def:boundedness-sigma-0}
\|\widehat{H}_0-H_0\|_{op}\leq  \|\widehat{H}_0\|_{op}+\|H_0\|_{op}\leq 2\dfrac{p\sqrt{p}hR_{\max}}{1-\gamma}\Big(hG^2+L\Big)=:\sigma_{H_0}.
\end{flalign}
This concludes the proof.
\end{proof}

\subsection{Proof of Lemma \ref{high-pro-i-1-term-lem}}

\begin{proof}(of Lemma \ref{high-pro-i-1-term-lem})
Recall 
$I_1=\sum_{j=0}^{k}(I+\alpha H_0)^{k-j}\big(\widehat{\xi}_{j}+\xi_0\big)$, which implies to bound $I_1$, we need to bound $\widehat{\xi}_{k}$.

Firstly, we decompose the term $\widehat{\xi}_{k}=(\widehat{H}_0-H_0)(\widehat{\theta}_{k}-\widehat{\theta}_{0})$.
Consider the update (\ref{sec-update-2}), we have
\begin{flalign}
\label{gap-hat-theta}
\widehat{\theta}_{k+1}-\widehat{\theta}_{k}&=(I+\alpha\widehat{H}_0)(\widehat{\theta}_{k}-\widehat{\theta}_{k-1})\\
\nonumber
&=(I+\alpha H_0)(\widehat{\theta}_{k}-\widehat{\theta}_{k-1})+\alpha(\widehat{H}_0-H_0)(\widehat{\theta}_{k}-\widehat{\theta}_{k-1})\\
\nonumber
&=\cdots\cdots\\
\label{theta_k-theta_k-1}
&=(I+\alpha H_0)^{k}(\widehat{\theta}_{1}-\widehat{\theta}_{0})+\alpha^{k}(\widehat{H}_0-H_0)^{k}(\widehat{\theta}_{1}-\widehat{\theta}_{0}).
\end{flalign}
Recall $\widehat{\xi}_{k}=(\widehat{H}_0-H_0)(\widehat{\theta}_{k}-\widehat{\theta}_{0})$, according to the result of (\ref{theta_k-theta_k-1}), we rewrite $\widehat{\xi}_{k}$ as follows,
\begin{flalign}
\label{theta_k-theta_k-01}
\widehat{\xi}_{k}=(\widehat{H}_0-H_0)
\Bigg(
\sum_{j=0}^{k-1}\bigg((I+\alpha H_0)^{j}(\widehat{\theta}_{1}-\widehat{\theta}_{0})+\alpha^{j}(\widehat{H}_0-H_0)^{j}(\widehat{\theta}_{1}-\widehat{\theta}_{0})\bigg)
\Bigg).
\end{flalign}

\underline{Boundedness of $\widehat{\xi}_{k}=(\widehat{H}_0-H_0)(\widehat{\theta}_{k}-\widehat{\theta}_{0})$.}
\begin{flalign}
\nonumber
\big\|\widehat{\xi}_{k}\big\|_{2}\overset{(\ref{theta_k-theta_k-01})}=&\bigg\|(\widehat{H}_0-H_0)
\bigg(
\sum_{j=0}^{k-1}\Big((I+\alpha H_0)^{j}(\widehat{\theta}_{1}-\widehat{\theta}_{0})+\alpha^{j}(\widehat{H}_0-H_0)^{j}(\widehat{\theta}_{1}-\widehat{\theta}_{0})\Big)
\bigg)\bigg\|_{2}\\
\nonumber
\leq&
\big\|(\widehat{H}_0-H_0)\big\|_{op}\underbrace{\|\widehat{\theta}_{1}-\widehat{\theta}_{0}\|_{2}}_{=\alpha \|g(\theta_0)\|_2}
\bigg(
\sum_{j=0}^{k-1}\Big((1+\alpha \Lambda)^{j}+\alpha^{j}\big\|\widehat{H}_0-H_0\big\|^{j}_{op}\Big)
\bigg)\\
\nonumber
\overset{(\ref{def:boundedness-sigma-0})}\leq& \alpha\sigma_{H_0} \|g(\theta_0)\|_2\bigg(
\sum_{j=0}^{k-1}\Big((1+\alpha \Lambda)^{j}+\alpha^{j}\sigma^{j}_{H_0}
\bigg)
\\
\label{app-a-last-1}
\leq& \alpha\sigma_{H_0} \|g(\theta_0)\|_2\bigg(
 \dfrac{(1+\alpha \Lambda)^{k-1}}{\alpha \Lambda}
 +
  \dfrac{1-\alpha^{k-1} \sigma^{k-1}_{H_0}}{1-\alpha \sigma_{H_0}}
\bigg).
\end{flalign}
Let 
\begin{flalign}
\label{app-condition-k-ieq}
{(1+\alpha \Lambda)^{k-1}}\leq\dfrac{1}{1-\sqrt{\alpha} \sigma_{H_0}},
\end{flalign} 
which requires 
$
k\leq\bigg\lfloor \frac{\log\Big(\frac{1}{1-\sqrt{\alpha}\sigma_{H_0}}\Big)}{\log(1+\alpha \Lambda)}\bigg\rfloor.
$
For a enough small $\alpha$, the term $
\bigg\lfloor \frac{\log\Big(\frac{1}{1-\sqrt{\alpha}\sigma_{H_0}}\Big)}{\log(1+\alpha \Lambda)}\bigg\rfloor
$ decreases as $\Lambda$ increases.
Recall $\Lambda \ge\sqrt{\chi\epsilon}$, then we have
\begin{flalign}
\label{hat-kapp-0}
k\leq\bigg\lfloor \dfrac{\log\Big(\dfrac{1}{1-\sqrt{\alpha}\sigma_{H_0}}\Big)}{\log(1+\alpha \Lambda)}\bigg\rfloor\leq \bigg\lfloor \dfrac{\log\Big(\dfrac{1}{1-\sqrt{\alpha}\sigma_{H_0}}\Big)}{\log(1+\alpha \sqrt{\chi\epsilon})}\bigg\rfloor
=:\widehat{\kappa}_{0}.
\end{flalign}
Thus, for all $k\leq \widehat{\kappa}_{0}$, from (\ref{app-a-last-1}), and $\Lambda \ge\sqrt{\chi\epsilon}$, we have
\begin{flalign}
\nonumber
\big\|\widehat{\xi}_{k}\big\|_{2}\leq&
  {\alpha\sigma_{H_0} \|g(\theta_0)\|_2}
  \Big(
  \dfrac{1}{\alpha\Lambda(1-\sqrt{\alpha} \sigma_{H_0})}+\dfrac{1}{1-\alpha \sigma_{H_0}}
  \Big)\\
\nonumber
\leq&
  {\alpha\sigma_{H_0} \|g(\theta_0)\|_2}
  \Big(
  \dfrac{1}{\alpha\sqrt{\chi\epsilon}(1-\sqrt{\alpha} \sigma_{H_0})}+\dfrac{1}{1-\alpha \sigma_{H_0}}
  \Big)\\
\label{boundness-xi-2}    
\overset{(\ref{bound-g-k})}\leq&
  { \dfrac{\sigma_{H_0}G R_{\max}}{1-\gamma}}
  \Big(
  \dfrac{1}{\sqrt{\chi\epsilon}(1-\sqrt{\alpha} \sigma_{H_0})}+\dfrac{\alpha}{1-\alpha \sigma_{H_0}}
  \Big).
\end{flalign}
It is noteworthy that to ensure the term $1-\alpha \sigma_{H_0}$ and $1-\sqrt{\alpha} \sigma_{H_0}$  keep positive, we require $\alpha<\min\{\frac{1}{ \sigma_{H_0}},\frac{1}{ \sigma^2_{H_0}}\}$.

 \underline{Boundedness of $I_1$ in Eq.(\ref{iteration-the-j})/(\ref{iteration-nabla-j}).}
Furthermore, for each $0\leq j\leq \widehat{\kappa}_{0}$, we have
\begin{flalign}
\nonumber
\|(I+\alpha H_0)^{k-j}\big(\widehat{\xi}_{j}+\xi_0\big)\|_2&\leq\| I+\alpha H_0\|^{\widehat{\kappa}_{0}}_{op}\|\widehat{\xi}_{j}+\xi_0\|_2\\
\nonumber
= &(1+\alpha \Lambda)^{\widehat{\kappa}_{0}}\|\widehat{\xi}_{j}+\xi_0\|_2\leq (1+\alpha \Lambda)^{\widehat{\kappa}_{0}}
\Big(
\|\widehat{\xi}_{j}\|_2+\|\xi_0\|_2
\Big)
\\
\nonumber
\overset{(\ref{boundness-xi-2})}\leq& \dfrac{1}{1-\sqrt{\alpha} \sigma_{H_0}}\bigg(
  { \dfrac{\sigma_{H_0}G R_{\max}}{1-\gamma}}
  \Big(
  \dfrac{1}{\sqrt{\chi\epsilon}(1-\sqrt{\alpha} \sigma_{H_0})}+\dfrac{\alpha}{1-\alpha \sigma_{H_0}}
  \Big)+\sigma
  \bigg)\\
\label{app-boundness-xi-1}
=: &C_1.
\end{flalign}
Finally, for each $0\leq j\leq \widehat{\kappa}_{0}$ and $k\leq \widehat{\kappa}_{0}$ we have
\begin{flalign}
\label{app-xi-mean-0}
\mathbb{E}\Big[(I+\alpha H_0)^{k-j}\big(\widehat{\xi}_{j}+\xi_0)\Big]=0.
\end{flalign}
Recall Hoeffding inequality, and the result of (\ref{app-boundness-xi-1}),(\ref{app-xi-mean-0}), for each $k\in[0,\widehat{\kappa}_{0}]$, for any $\delta\in(0,1)$, 
\begin{flalign}
\label{high-p-I-1}
\mathbb{P}\Bigg(
\Big\|\sum_{j=0}^{k}(I+\alpha H_0)^{k-j}\big(\widehat{\xi}_{j}+\xi_0\big)\Big\|_2\leq 2C_1\sqrt{\widehat{\kappa}_{0}\log{\dfrac{4}{\delta}}}\Bigg)\ge1-\delta.
\end{flalign}
\end{proof}
\begin{remark}
For a small $\alpha$, the term $C_1$ (\ref{app-boundness-xi-1}) can be rewritten as follows,
\begin{flalign}
C_{1}=  { \dfrac{\sigma_{H_0}G R_{\max}}{(1-\gamma)\sqrt{\chi\epsilon}}}
+\sigma+\mathcal{O}(\alpha).
\end{flalign}
\end{remark}

\subsection{Proof of Lemma \ref{high-pro-i-02-term-lem}}
\label{proof-ap-a-lemmapg-gap-nabla-j-hat}
\begin{proof}(of Lemma \ref{high-pro-i-02-term-lem}).
Recall $\widehat{\theta}_0=\theta_0$, then 
$\|\nabla{J}(\widehat{\theta}_0)\|\leq\epsilon$.
Thus, for each $k\in[0,\widehat{\kappa}_{0}]$, we have
\begin{flalign}
\label{boundness-I-2}
\big\|I_2\big\|_{2}=\Big\|\sum_{j=0}^{k}(I+\alpha H_0)^{j}\nabla{J}(\widehat{\theta}_0)\Big\|_{2}
\leq\sum_{j=0}^{k}\big(1+\alpha \Lambda)\big)^{j}\big\|\nabla{J}(\widehat{\theta}_0)\big\|_{2}
\overset{(\ref{app-condition-k-ieq})}\leq\dfrac{\epsilon}{1-\sqrt{\alpha}\sigma_{H_0}}.
\end{flalign}
This concludes the proof.
\end{proof}

\subsection{Proof of Lemma \ref{app-key-lem}}
\label{proof-of-emma-app-key-lem}

\begin{proof}(of Lemma \ref{app-key-lem})

\underline{Boundedness for the term $\widehat{\theta}_{k+1}-\widehat{\theta}_{0}$ defined in Eq.(\ref{iteration-the-j}) with high probability.}

Recall the term $\widehat{\theta}_{k+1}-\widehat{\theta}_{0}=\alpha I_1+\alpha I_2$ defined in (\ref{iteration-the-j}),
and results of (\ref{high-p-I-1}) and (\ref{boundness-I-2}) show the boundedness of $I_1$ and $I_2$ correspondingly. 
Then for each $k\in[0,\widehat{\kappa}_{0}]$, for any $\delta\in(0,1)$, we have
\begin{flalign}
\label{high-p-theta-gap}
\mathbb{P}\Bigg(
\big\|\widehat{\theta}_{k+1}-\widehat{\theta}_{0}\big\|_2\leq\alpha\bigg(2C_1\sqrt{ \widehat{\kappa}_{0}\log{\dfrac{4}{\delta}}}+
\dfrac{\epsilon}{1-\sqrt{\alpha}\sigma_{H_0}}
\bigg)
\Bigg)\ge1-\delta.
\end{flalign}

\underline{Boundedness for the term $\nabla \widehat{J}(\widehat{\theta}_{k+1})$ defined in Eq.(\ref{iteration-nabla-j}) with high probability.}

Firstly, we bound the term $(I+\alpha H_0)^{k}\nabla J(\theta_0)$ in Eq.(\ref{iteration-nabla-j}). 
Recall the condition of $\widehat{\kappa}_{0}$ defined in (\ref{hat-kapp-0}), the result of (\ref{app-A-17}), and the initial condition of $\|\nabla J(\theta_0)\big\|_2\leq\epsilon$, then we have
\begin{flalign}
\nonumber
\big\|(I+\alpha H_0)^{k}\nabla J(\theta_0)\big\|_2\leq\big\|I+\alpha H_0\|^{k}_{op}\|\nabla J(\theta_0)\|_2=(1+\alpha\Lambda)^{k}\|\nabla J(\theta_0)\big\|_2
\leq\dfrac{\epsilon}{1-\sqrt \alpha \sigma_{H_0}}.
\end{flalign}
Furthermore, from the result of (\ref{high-p-I-1}), for each $k\in[0,\widehat{\kappa}_{0}]$, the following holds with probability at least $1-\delta$,
$
\big\|\nabla \widehat{J}(\widehat{\theta}_{k+1})\big\|_2
=\Big\|(I+\alpha H_0)^{k}\nabla J(\theta_0)+{\alpha\sum_{j=0}^{k}(I+\alpha H_0)^{k-j}\big(\widehat{\xi}_{j}+\xi_0\big)}\Big\|_{2}
\leq\alpha2C_1\sqrt{ \widehat{\kappa}_{0}\log{\dfrac{4}{\delta}}}+
\dfrac{\epsilon}{1-\alpha \sigma_{H_0}}.
$
That is 
\begin{flalign}
\label{high-p-nabl-J-gap}
\mathbb{P}\Bigg(
\big\|\nabla \widehat{J}(\widehat{\theta}_{k+1})\big\|_2\leq
 2\alpha C_1\sqrt{ \widehat{\kappa}_{0}\log{\dfrac{4}{\delta}}}+
\dfrac{\epsilon}{1-\sqrt \alpha \sigma_{H_0}}
\Bigg)\ge1-\delta.
\end{flalign}
This concludes the proof.
\end{proof}
For simplify, we introduce two following notations $C_1,C_2$to short expressions,
\begin{flalign}
\label{c-2}
C_{2}=2C_1\sqrt{ \widehat{\kappa}_{0}\log{\dfrac{4}{\delta}}}+
\dfrac{\epsilon}{1-\sqrt \alpha\sigma_{H_0}},
C_{3}=2C_1\sqrt{ \widehat{\kappa}_{0}\log{\dfrac{4}{\delta}}}+
\dfrac{ 1}{1-\sqrt \alpha \sigma_{H_0}}.
\end{flalign}

\subsection{Proof of Lemma \ref{boundedness-j-the-k}}

Firstly,  let's bound on the term $\|\nabla A^{\pi_{\theta}}(s,a)\|_{2}=\|\nabla Q^{\pi_{\theta}}(s,a)-\nabla V^{\pi_{\theta}}(s)\|_{2}$. In fact,
\begin{flalign}
\label{app-A-bound-V}
\|\nabla V^{\pi_{\theta}}(s)\|_{2}=\Big\|\nabla \mathbb{E}_{\pi_\theta}[\sum_{t=0}^{\infty}\gamma^{t}r_{t+1}]\Big\|_{2}=\Big\|\sum_{a\in\mathcal{A}}\underbrace{\nabla \pi_{\theta}(a|s)}_{\overset{(\ref{def:F-G})}\leq U}\sum_{t=0}^{\infty}\gamma^{t}r_{t+1}\Big\|_{2}
\leq\sum_{a\in\mathcal{A}}\frac{R_{\max}}{1-\gamma}=\frac{UR_{\max}|\mathcal{A}|}{1-\gamma},
\end{flalign}
\begin{flalign}
\nonumber
\|\nabla Q^{\pi_{\theta}}(s,a)\|_{2}&=\Big\|\nabla\big(R(s,a)+\gamma\sum_{s^{'}\in\mathcal{S}}P(s^{'}|s,a)V^{\pi_{\theta}}(s)\big)\Big\|_{2}=\gamma\Big\|\sum_{s^{'}\in\mathcal{S}}P(s^{'}|s,a)\nabla V^{\pi_{\theta}}(s)\Big\|_{2}\\
\label{app-A-bound-Q}
&\leq\gamma\sum_{s^{'}\in\mathcal{S}}\|\nabla V^{\pi_{\theta}}(s)\|_{2}
\overset{(\ref{app-A-bound-V})}\leq\frac{\gamma U R_{\max}|\mathcal{A}||\mathcal{S}|}{1-\gamma}.
\end{flalign}
Thus, by the results of (\ref{app-A-bound-V}) and (\ref{app-A-bound-Q}), we have 
\begin{flalign}
\label{app-A-bound-A}
\Big\|\nabla A^{\pi_{\theta}}(s,a)\Big\|_{2}\leq\Big\|\nabla Q^{\pi_{\theta}}(s,a)\Big\|_{2}+\Big\|\nabla V^{\pi_{\theta}}(s)\Big\|_{2}
\leq (1+\gamma |\mathcal{S}|)\frac{UR_{\max}|\mathcal{A}|}{1-\gamma}.
\end{flalign}
Now, we consider the boundedness of $\Big\|\nabla J(\theta_{\star})-\nabla J(\theta_k)\Big\|_{2}$, in fact,
\begin{flalign} 
\nonumber
	\Big\|\nabla \Big(J(\theta_k)-J(\theta_{\star})\Big)\Big\|_{2}
	\overset{(\ref{kakade-2002})}=
	&\Big\|\nabla \int_{s\in\mathcal{S}}\int_{a\in\mathcal{A}}d_{\rho_0}^{\pi_{\theta_{\star}}}(s)\pi_{\theta_{\star}}(a|s)A^{\pi_{\theta}}(s,a)\text{d}s\text{d}a\Big\|_{2}\\
\nonumber
=&\Big\|\nabla \int_{s\in\mathcal{S}}\int_{a\in\mathcal{A}}\dfrac{d_{\rho_0}^{\pi_{\theta_{\star}}}(s)}{d_{\rho_0}^{\pi_{\theta}}(s)}d_{\rho_0}^{\pi_{\theta}}(s)\pi_{\theta_{\star}}(a|s)A^{\pi_{\theta}}(s,a)\text{d}s\text{d}a\Big\|_{2}\\
\nonumber
=&\Big\|  \int_{s\in\mathcal{S}}\int_{a\in\mathcal{A}}\dfrac{d_{\rho_0}^{\pi_{\theta_{\star}}}(s)}{d_{\rho_0}^{\pi_{\theta}}(s)}d_{\rho_0}^{\pi_{\theta}}(s)\underbrace{\pi_{\theta_{\star}}(a|s)}_{\leq1}\nabla A^{\pi_{\theta}}(s,a)\text{d}s\text{d}a\Big\|_{2}\\
\nonumber
\leq& \int_{s\in\mathcal{S}}\int_{a\in\mathcal{A}}\bigg(\dfrac{d_{\rho_0}^{\pi_{\theta_{\star}}}(s)}{d_{\rho_0}^{\pi_{\theta}}(s)} \big\|d_{\rho_0}^{\pi_{\theta}}(s)\big\|_2\big\|\nabla A^{\pi_{\theta}}(s,a)\big\|_{2}\bigg)\text{d}s\text{d}a\\
\label{app-A-15}
\leq&\int_{s\in\mathcal{S}}\int_{a\in\mathcal{A}}\underbrace{\max_{s\in\mathcal{S}}\dfrac{d_{\rho_0}^{\pi_{\theta_{\star}}}(s)}{d_{\rho_0}^{\pi_{\theta}}(s)}}_{=\Big\|\dfrac{{d_{\rho_0}^{\pi_{\theta_{\star}}}}}{d_{\rho_0}^{\pi_{\theta}}}\Big\|_{\infty}}\dfrac{1}{1-\gamma}\big\|\nabla A^{\pi_{\theta}}(s,a)\big\|_{2}\text{d}s\text{d}a\\
\label{app-A-16}
\overset{(\ref{app-A-bound-A})}\leq&\int_{s\in\mathcal{S}}\int_{a\in\mathcal{A}}\Big\|\dfrac{{d_{\rho_0}^{\pi_{\theta_{\star}}}}}{\rho_0}\Big\|_{\infty}(1+\gamma |\mathcal{S}|)\dfrac{UR_{\max}|\mathcal{A}|}{(1-\gamma)^2}\text{d}s\text{d}a\\
\label{app-A-17}
=&\Big\|\dfrac{{d_{\rho_0}^{\pi_{\theta_{\star}}}}}{\rho_0}\Big\|_{\infty}\dfrac{(1+\gamma |\mathcal{S}|)|\mathcal{A}|^2 |\mathcal{S}| UR_{\max}}{(1-\gamma)^2},
\end{flalign}
where Eq.(\ref{app-A-15}) holds since 
\begin{flalign}
\nonumber
\|d^{\pi_\theta}_{\rho_0}(s)\|_2=\|\mathbb{E}_{s_0\sim\rho_{0}(\cdot)}[d^{\pi_\theta}_{s_0}(s)]\|_{2}
=\|\mathbb{E}_{s_0\sim\rho_{0}(\cdot)}[\sum_{t=0}^{\infty}\gamma^{t}\underbrace{P^{\pi_\theta}(s_t=s|s_0)}_{\leq 1}]\|_{2}\leq\dfrac{1}{1-\gamma}.
\end{flalign}
Eq.(\ref{app-A-16}) holds since
$
d_{\rho_0}^{\pi_{\theta}}(s)\ge\rho_{0}(s),
$
thus, we have
\[
\Big\|\dfrac{{d_{\rho_0}^{\pi_{\theta_{\star}}}}}{d_{\rho_0}^{\pi_{\theta}}}\Big\|_{\infty}=\max_{s\in\mathcal{S}}\dfrac{d_{\rho_0}^{\pi_{\theta_{\star}}}(s)}{d_{\rho_0}^{\pi_{\theta}}(s)}\leq\max_{s\in\mathcal{S}}\dfrac{d_{\rho_0}^{\pi_{\theta_{\star}}}(s)}{{\rho_0(s)}}=\Big\|\dfrac{{d_{\rho_0}^{\pi_{\theta_{\star}}}}}{\rho_0}\Big\|_{\infty}.
\]

\subsection{Proof of Lemma \ref{app-key-lem-second-1}}

\textbf{Organization}
It is very technical to achieve the results of Lemma \ref{app-key-lem-second-1}, our outline our proof as follows:
in section \ref{bound-lemma-step-01}, we bound all the four terms of (\ref{iterdelta}); 
in section \ref{bound-lemma-step-02}, we bound the term $Z_j$ with high probability, 
which is a preliminary result for applying Azuma's Inequality to get a further result;
in section \ref{bound-lemma-step-03}, we bound the $\Delta_{k}=\nabla J(\theta_{k})-\nabla\widehat{J}(\widehat{\theta}_{k})$ with high probability;
finally, in section \ref{bound-lemma-step-04}, we bound $\theta_{k+1}-\widehat{\theta}_{k+1}$ with high probability.

\begin{proof}(of Lemma \ref{app-key-lem-second-1})
Before we present the details of the proof, we need some preliminary results.

\underline{Preliminary results}.
Under Assumption \ref{ass:On-policy-derivative}-\ref{ass:hessian-lipschitz}, 
we have 
\begin{flalign}
\label{app-h-k-e-k-lips}
\big\|\Delta H_{k}\big\|_{op}\leq \chi \big(\big\|\theta_{k+1}-\widehat{\theta}_{k+1}\big\|_2+\big\|\widehat{\theta}_{k+1}-\widehat{\theta}_0\big\|_2\big), ~~~\big\|e_{k}\big\|_2\leq \dfrac{\chi}{2}\big\|\theta_{k}-\theta_{k+1}\big\|_2.
\end{flalign}
We define two events $\mathcal{B}_{k}$ and $\mathcal{C}_{k}$ as follows,
\begin{flalign}
\label{event-1-k}
\mathcal{B}_{k}&=\bigcap_{j=0}^{k}\Big\{
j\leq k:
\big\|\widehat{\theta}_{j}-\widehat{\theta}_{0}\big\|_2\leq\alpha C_2~\text{and}~\big\|\nabla \widehat{J}(\widehat{\theta}_{j})\big\|_2\leq\alpha C_3
\Big\},\\
\label{event-2-k}
\mathcal{C}_{k}&=\bigcap_{j=0}^{k}\bigg\{j\leq k:\max\Big\{\big\|\Delta_j\big\|_2,\big\|\Delta_j\big\|^2_2\Big\}\leq C_4\sqrt{\alpha^3\log\dfrac{1}{\alpha}}\bigg\},
\end{flalign}
where $C_{2}$ and $C_3$ are defined in (\ref{c-2}), $C_4$ is a positive constant, and we will provide the rule of choosing $C_4$ in (\ref{condition:C-4}). 
Since $\Delta_0=0$, the event $\mathcal{C}_0\ne\varnothing$.
The chosen rule of $C_4$ ensures for each $k\in[0,\widehat{\kappa}_0]$, the event $\mathcal{C}_{k}\ne\varnothing$.

Furthermore,
for each $k\in[0,\widehat{\kappa}_{0}]$, the following equation occurs with probability at least $1-\delta$,
\begin{flalign}
\nonumber
\|\Delta H_{k}\|_{op}\overset{(\ref{app-h-k-e-k-lips})}\leq& \chi \big(\underbrace{\|\theta_{k+1}-\widehat{\theta}_{k+1}\|_2}_{\overset{(\ref{itertheta})}\leq\alpha\big\|\sum_{j=0}^{k}\Delta_{j}\big\|_2}+\underbrace{\|\widehat{\theta}_{k+1}-\widehat{\theta}_0\|_2}_{\overset{(\ref{event-1-k})}\leq \alpha C_2}\big)\\
\label{boundness-del-H-K}
\leq&\chi \alpha\Big(\sum_{j=0}^{k}\big\|\Delta_{j}\big\|_2+C_2\Big)\overset{(\ref{event-2-k})}\leq \chi \alpha\Big(C_4\widehat{\kappa}_{0}\sqrt{\alpha^3\log\dfrac{1}{\alpha}}+C_2\Big).
\end{flalign}

\subsubsection{Bounded of all the four terms of Eq.(\ref{iterdelta})}
\label{bound-lemma-step-01}

For each $k\in[0,\widehat{\kappa}_{0}]$, the following equation occurs with probability at least $1-\delta$:

\ding{182} for the first term of Eq.(\ref{iterdelta}),
\begin{flalign}
\label{app-a-bound-1}
\big\|(I+\alpha H_0)\Delta_{k}\big\|_2\leq\big\|I+\alpha H_0\big\|_{op}\big\|\Delta_{k}\big\|_2\overset{(\ref{event-2-k})}\leq (1+\alpha\Lambda)\sqrt{\alpha^3\log\dfrac{1}{\alpha}}C_4
=:\sqrt{\alpha^3\log\dfrac{1}{\alpha}} B_1,
\end{flalign}
where $B_1=(1+\alpha\Lambda)C_4$.

\ding{183} for the second term of Eq.(\ref{iterdelta}),
\begin{flalign}
\nonumber
\big\|\alpha\Delta H_{k}[\Delta_{k}+\nabla\widehat{J}(\widehat{\theta}_{k})]\big\|_2=&\alpha\big\|\Delta H_{k} \nabla J(\theta_{k})\big\|_2\leq \alpha\big\|\Delta H_{k}\big\|_{op}\big\| \nabla J(\theta_{k})\big\|_2\\
\nonumber
\overset{(\ref{app-A-17}),(\ref{boundness-del-H-K})}\leq& \alpha^2\chi\Big(C_4\widehat{\kappa}_{0}\sqrt{\alpha^3\log\dfrac{1}{\alpha}}+C_2\Big)\Big\|\dfrac{{d_{\rho_0}^{\pi_{\theta_{\star}}}}}{\rho_0}\Big\|_{\infty}\dfrac{(1+\gamma |\mathcal{S}|)|\mathcal{A}|^2 |\mathcal{S}| UR_{\max}}{(1-\gamma)^2}\\
\label{app-a-bound-2}
=:&\alpha^2 B_2,
\end{flalign}
where $B_2=\chi\big(C_4\widehat{\kappa}_{0}\sqrt{\alpha^3\log\frac{1}{\alpha}}+C_2\big)\Big\|\frac{{d_{\rho_0}^{\pi_{\theta_{\star}}}}}{\rho_0}\Big\|_{\infty}\frac{(1+\gamma |\mathcal{S}|)|\mathcal{A}|^2 |\mathcal{S}| UR_{\max}}{(1-\gamma)^2}$.

\ding{184} for the third term of Eq.(\ref{iterdelta}),
\begin{flalign}
\nonumber
&\|\alpha\Delta H_{k}(\widehat{\xi}_{k}+\xi_0)\|_2\\
\nonumber
\overset{(\ref{boundness-xi-2}),(\ref{boundness-del-H-K})}
\leq&\alpha^{2} \chi\Big(\sqrt{\alpha^3\log\dfrac{1}{\alpha}}\widehat{\kappa}_{0}C_4+ C_2\Big)
\bigg(
  { \dfrac{\sigma_{H_0}G R_{\max}}{1-\gamma}}
  \Big(
  \dfrac{1}{\sqrt{\chi\epsilon}(1-\sqrt{\alpha} \sigma_{H_0})}+\dfrac{\alpha}{1-\alpha \sigma_{H_0}}
  \Big)+\sigma
  \bigg)
\\
\label{app-a-bound-3}
=:&\alpha^{2} B_3,
\end{flalign}
where $B_3=\chi\big(\sqrt{\alpha^3\log\frac{1}{\alpha}}\widehat{\kappa}_{0}C_4+ C_2\big)\Big(
  { \dfrac{\sigma_{H_0}G R_{\max}}{1-\gamma}}
  \big(
  \frac{1}{\sqrt{\chi\epsilon}(1-\sqrt{\alpha} \sigma_{H_0})}+\frac{\alpha}{1-\alpha \sigma_{H_0}}
  \big)+\sigma
  \Big)$.

\ding{185} for the forth term of Eq.(\ref{iterdelta}),
\begin{flalign}
\label{app-a-bound-4}
\|e_{k}\|_2=\|\theta_{k+1}-\theta_{k}\|_{2}\leq\alpha\dfrac{\chi}{2}\|g(\theta_{k})\|_{2}
\overset{(\ref{bound-g-k})}\leq\dfrac{1}{2} \alpha\chi
\Big(
\dfrac{GR_{\max}}{1-\gamma}
\Big)
=:\alpha B_4,
\end{flalign}
where $B_4=\chi
\big(
\frac{GR_{\max}}{2(1-\gamma)}
\big)$.

From the results of (\ref{app-a-bound-1})-(\ref{app-a-bound-4}), we have
\begin{flalign}
\nonumber
\mathbb{E}\Big[\|\Delta_{k+1}\|_2^{2}\Big|\mathcal{F}_{k}\Big]\overset{(\ref{iterdelta})}\leq&\|(I+\alpha H_0)\Delta_{k}\|^{2}_2+2\alpha\|(I+\alpha H_0)\Delta_{k}\|_{2}(\alpha B_2+\alpha B_3 + B_4)
\\
\label{bound-del-1}
&+\alpha^4 B^{2}_2 +\alpha^4 B^2_3+\alpha^2 B^{2}_4+2(\alpha^4 B_2B_3+\alpha^3 B_2B_4+\alpha^3B_3B_4).
\end{flalign}
For simplify, we introduce two following notations to short expressions, 
\begin{flalign}
\label{def:d-1-2}
D_1=\alpha B_2+\alpha B_3 + B_4,~~~D_2=\alpha^2 B^{2}_2 +\alpha^2 B^2_3+B^{2}_4+2(\alpha^2 B_2B_3+\alpha B_2B_4+\alpha B_3B_4),
\end{flalign}
then we can rewrite (\ref{bound-del-1})
as follows,
\begin{flalign}
\nonumber
\mathbb{E}\Big[\big\|\Delta_{k+1}\big\|_2^{2}|\mathcal{F}_{k}\Big]&\leq\big\|(I+\alpha H_0)\Delta_{k}\big\|^{2}_2+2\alpha D_1 \big\|(I+\alpha H_0)\Delta_{k}\big\|_{2}+\alpha^{2} D_2\\
\label{app-up-del-1}
\overset{(\ref{app-a-bound-1})}\leq&(1+\alpha\Lambda)^{2}\big\|\Delta_{k}\big\|_2^{2}+2\alpha^{\frac{5}{2}}\sqrt{\log\dfrac{1}{\alpha} }D_1B_1+\alpha^2 D_2.
\end{flalign}
Rearranging Eq.(\ref{app-up-del-1}), we have
\begin{flalign}
\nonumber
&\mathbb{E}\Bigg[\big\|\Delta_{k+1}\big\|_2^{2}+\frac{2\alpha^{\frac{3}{2}}\sqrt{\log\frac{1}{\alpha} }D_1B_1+\alpha D_2}{\alpha \Lambda^2+2\Lambda}\Bigg|\mathcal{F}_{k}\Bigg]\\
\label{app-up-dell-re-1}
\leq&(1+\alpha\Lambda)^{2}\Bigg(\big\|\Delta_{k}\big\|_2^{2}+\frac{2\alpha^{\frac{3}{2}}\sqrt{\log\frac{1}{\alpha} }D_1B_1+\alpha D_2}{\alpha \Lambda^2+2\Lambda}\Bigg).
\end{flalign}

\subsubsection{Boundedness of $Z_j=(1+\alpha\Lambda)^{-2j}\Big(\|\Delta_{j}\|_2^{2}+\frac{2\alpha^{\frac{3}{2}}\sqrt{\log\frac{1}{\alpha} }D_1B_1+\alpha D_2}{\alpha \Lambda^2+2\Lambda}\Big)$.}

\label{bound-lemma-step-02}

We will use Azuma's inequality, i.e., Lemma \ref{lem:Azuma-Inequality}
to achieve the boundedness of $Z_j$, which require us to show: (i) $\{Z_j\}$ is a super-martingale, (ii) the boundedness $|Z_{j+1}-Z_{j}|$ almost surely.

Let $\mathcal{F}_{k}=\sigma\{{\xi}_{0},\widehat{\xi}_{1},\cdots,\widehat{\xi}_{k}\}$ be a filtration generated by all the information from time $0$ to $k$.
Then $\{Z_j\}$ is a super-martingale with respect to the filtration $\{\mathcal{F}_j\}$. In fact,
for each $j\in[0,\widehat{\kappa}_{0}]$, we have
\begin{flalign}
\nonumber
&\mathbb{E}\Big[Z_{j}\Big|\mathcal{F}_{j-1}\Big]\\
\nonumber
=&(1+\alpha\Lambda)^{-2j}\bigg(\mathbb{E}\Big[\|\Delta_{j}\|_2^{2}\Big]+\frac{2\alpha^{\frac{3}{2}}\sqrt{\log\frac{1}{\alpha} }D_1B_1+\alpha D_2}{\alpha \Lambda^2+2\Lambda}\bigg)\\
\nonumber
\overset{(\ref{app-up-del-1})}\leq&(1+\alpha\Lambda)^{-2j}\bigg((1+\alpha\Lambda)^2\|\Delta_{j-1}\|_2^{2}+2\alpha^{\frac{5}{2}}\sqrt{\log\frac{1}{\alpha} }D_1B_1+\alpha^2 D_2+\frac{2\alpha^{\frac{3}{2}}\sqrt{\log\frac{1}{\alpha} }D_1B_1+\alpha D_2}{\alpha \Lambda^2+2\Lambda}\bigg)\\
\nonumber
\leq&(1+\alpha\Lambda)^{-2(j-1)}\bigg(\|\Delta_{j-1}\|_2^{2}+\underbrace{\frac{2\alpha^{\frac{5}{2}}\sqrt{\log\frac{1}{\alpha} }D_1B_1+\alpha^2 D_2}{(1+\alpha\Lambda)^2}+\frac{2\alpha^{\frac{3}{2}}\sqrt{\log\frac{1}{\alpha} }D_1B_1+\alpha D_2}{(\alpha \Lambda^2+2\Lambda)(1+\alpha\Lambda)^2}}_{=\dfrac{2\alpha^{\frac{3}{2}}\sqrt{\log\dfrac{1}{\alpha} }D_1B_1+\alpha D_2}{\alpha \Lambda^2+2\Lambda}}\bigg)\\
\label{app-con-cc-4}
=&Z_{j-1},
\end{flalign}
which implies $Z_{j}$ is a super-martingale.

We introduce $\beta$ to short the expression $Z_j$,
\[
\beta=\frac{2\sqrt{\alpha\log\frac{1}{\alpha} }D_1B_1+ D_2}{\Lambda^2+2\Lambda}
\]
i.e., $Z_j=(1+\alpha\Lambda)^{-2j}\big(\|\Delta_{j}\|_2^{2}+\alpha\beta\big)$.

Furthermore, let us bound the term $\big|Z_{j}-Z_{j-1}\big|$.
In fact, since $Z_{j}$ is a super-martingale, then
\begin{flalign}
\label{app-iner-z-gap}
&\big|Z_{j}-Z_{j-1}\big|\\
\nonumber
\leq&\big|Z_{j}-\mathbb{E}\big[Z_{j}\big|\mathcal{F}_{j-1}\big]\big|
=(1+\alpha\Lambda)^{-2j}\big| \|\Delta^{2}_{j}\|_{2}-\mathbb{E}\big[\|\Delta^{2}_{j}\|_{2}|\mathcal{F}_{j-1}\big]\big|.
\end{flalign}
The result of (\ref{app-iner-z-gap}) shows that to bound the term $\big|Z_{j}-Z_{j-1}\big|$,
we need to bound the term $\big| \|\Delta^{2}_{j}\|_{2}-\mathbb{E}\big[\Delta^{2}_{j}\|_{2}|\mathcal{F}_{j-1}\big]\big|$.

Recall the filtration $\mathcal{F}_{j}=\sigma\{{\xi}_{0},\widehat{\xi}_{1},\cdots,\widehat{\xi}_{j}\}$, and $\Delta_{j}$ defined in (\ref{iterdelta}), where the first two terms are deterministic, and only the third and fourth term
are random. Then the following holds almost surely,
\begin{flalign}
\nonumber
\Big| \|\Delta_{j}\|^{2}_{2}-\mathbb{E}\Big[\|\Delta_{j}\|^{2}_{2}|\mathcal{F}_{j-1}\Big]\Big|
\leq&4\alpha\big\|\big(I+\alpha H_0\big)\Delta_{j}\big\|_{2}\big(\alpha B_3 + B_4\big)
\\
\nonumber
&+2\alpha^4 B^2_3+2\alpha^2 B^{2}_4+4\big(\alpha^4 B_2B_3+\alpha^3B_2B_3+\alpha^3B_3B_4\big)\\
\nonumber
\overset{(\ref{app-a-bound-1})}\leq&4\alpha\sqrt{\alpha^3\log\dfrac{1}{\alpha}}B_1\big(\alpha B_3 + B_4\big)
\\
\nonumber
&+2\alpha^4 B^2_3+2\alpha^2 B^{2}_4+4\big(\alpha^4 B_2B_3+\alpha^3B_2B_3+\alpha^3B_3B_4\big)\\
\label{c-j-new}
=:&c_{j}.
\end{flalign}
To short the expression, we introduce two notations as follows,
\begin{flalign}
\nonumber
E_1&=4B_1(\alpha B_3 + B_4),\\
\label{def:constant-E}
E_2&=2\alpha^2 B^2_3+2 B^{2}_4+4(\alpha^2 B_2B_3+\alpha B_2B_3+\alpha B_3B_4).
\end{flalign}
From the results of (\ref{app-iner-z-gap}) and (\ref{c-j-new}), we have the boundedness of $\big|Z_{j}-Z_{j-1}\big|$ as follows,
\[
\big|Z_{j}-Z_{j-1}\big|\leq \alpha^{\frac{5}{2}}\sqrt{\log\dfrac{1}{\alpha}} E_1 +\alpha^{2} E_2 =c_j.
\]

Finally, from Azuma's inequality, i.e., Lemma \ref{lem:Azuma-Inequality}, the following holds, for any $\delta>0$, we have
\begin{flalign}
\nonumber
\mathbb{P}\big(Z_{\widehat{\kappa}_0}-Z_0\ge\delta\big)\leq\exp
\bigg(
-\dfrac{2\delta^2}{\sum_{j= 0}^{\widehat{\kappa}_0} c_{j}^{2}}
\bigg),
\end{flalign} 
which implies the following holds with probability less than $\alpha^2$
\begin{flalign}
\label{high-probba-z-1}
Z_{\widehat{\kappa}_0}-Z_0\ge\sqrt{\log\dfrac{1}{\alpha}\sum_{j= 0}^{\widehat{\kappa}_0} c_{j}^{2}}.
\end{flalign}
Recall
$
\widehat{\kappa}_{0}=\bigg\lfloor \frac{\log\Big(\frac{1}{1-\sqrt{\alpha}\sigma_{H_0}}\Big)}{\log(1+\alpha \sqrt{\chi\epsilon})}\bigg\rfloor
$, and Taylor's expression of $\log(1+x)$,
\[
\log(1+x)=x-\dfrac{x^2}{2}+\dfrac{x^3}{3}-\dfrac{x^4}{4}+\cdots+(-1)^{n-1}\dfrac{x^n}{n}+\cdots,~~x\in(-1,1),
\]
then for a enough small $\alpha$, $\widehat{\kappa}_{0}$ is upper-bounded as follows,
\begin{flalign}
\label{estimate-ka-0}
\widehat{\kappa}_{0}=\Bigg\lfloor \dfrac{\log\Big(\dfrac{1}{1-\sqrt{\alpha}\sigma_{H_0}}\Big)}{\log(1+\alpha \sqrt{\chi\epsilon})}\Bigg \rfloor=\mathcal{O}\Big(\dfrac{\sqrt{\alpha} \sigma_{H_0}}{\alpha\Lambda}\Big)+1\leq \mathcal{O}\Big(\dfrac{\sigma_{H_0}}{\sqrt{\chi\epsilon\alpha}}\Big).
\end{flalign}
From the result of (\ref{high-probba-z-1}), for each $k\in[1,\widehat{\kappa}_{0}]$, we have
\begin{flalign}
\label{eq-high-p-z-k-z-001}
\alpha^2&\ge\mathbb{P}\Bigg(Z_{k}-Z_0\ge \sqrt{\log\dfrac{1}{\alpha}\sum_{j= 0}^{\widehat{\kappa}_0} c_{j}^{2}}\Bigg)
=\mathbb{P}\Bigg(Z_{k}\ge Z_0+\sqrt{\log\dfrac{1}{\alpha}\sum_{j= 0}^{\widehat{\kappa}_0} c_{j}^{2}}\Bigg)
\\
\label{eq-high-p-z-k-z-0}
&=\mathbb{P}\Bigg(Z_{k}\ge \alpha\beta+\mathcal{O}\bigg(
\alpha^{\frac{7}{4}}\sqrt{\log\dfrac{1}{\alpha}}E_2\sqrt{\dfrac{\sigma_{H_0}}{\sqrt{\chi\epsilon}}}
\bigg)\Bigg),
\end{flalign}
where Eq.(\ref{eq-high-p-z-k-z-0}) holds from (\ref{eq-high-p-z-k-z-001}) due to the following three aspects:

(i) $Z_0=\|\Delta_{0}\|_2^{2}+ \alpha\beta$ and $\Delta_0=\nabla J(\theta_{0})-\nabla\widehat{J}(\widehat{\theta}_{0})\overset{(\ref{nab-j-re})}=0$, which implies $Z_0=\alpha\beta$;

(ii) recall $c_j=\alpha^{\frac{5}{2}}\sqrt{\log\frac{1}{\alpha}} E_1 +\alpha^{2} E_2$,
which implies for an enough small $\alpha$, we can estimate $c_j$ as follows,
$
c^2_j=\mathcal{O}\Big(
\alpha^4 E^2_2
\Big);
$

(iii)furthermore, since $\widehat{\kappa}_{0}$ is upper bounded by $ \mathcal{O}\Big(\dfrac{\sigma_{H_0}}{\sqrt{\chi\epsilon\alpha}}\Big)$,
then we have
\[
\sqrt{\log\dfrac{1}{\alpha}\sum_{j= 0}^{\widehat{\kappa}_0} c_{j}^{2}}=\mathcal{O}
\bigg(
\alpha^{\frac{7}{4}}\sqrt{\log\frac{1}{\alpha}}E_2\sqrt{\frac{\sigma_{H_0}}{\sqrt{\chi\epsilon}}}
\bigg).
\]

Rewrite (\ref{eq-high-p-z-k-z-0}), we present the boundedness of $Z_{k}$ with high probability as follows,
\[
\mathbb{P}\Bigg(Z_{k}\leq \alpha\beta+\mathcal{O}\bigg(
\alpha^{\frac{7}{4}}\sqrt{\log\dfrac{1}{\alpha}}E_2\sqrt{\dfrac{\sigma_{H_0}}{\sqrt{\chi\epsilon}}}
\bigg)\Bigg)\ge 1-\alpha^2.
\]

\subsubsection{Boundedness of $\Delta_{k}=\nabla J(\theta_{k})-\nabla\widehat{J}(\widehat{\theta}_{k})$.}
\label{bound-lemma-step-03}

Recall $Z_k=(1+\alpha\Lambda)^{-2k}\big(\|\Delta_{k}\|_2^{2}+\alpha\beta\big)$, the event 
\[Z_{k}\ge \alpha\beta+\mathcal{O}\bigg(
\alpha^{\frac{7}{4}}\sqrt{\log\dfrac{1}{\alpha}}E_2\sqrt{\dfrac{\sigma_{H_0}}{\sqrt{\chi\epsilon}}}
\bigg)\]
is equivalent to 
\begin{flalign}
\label{pro-ine-01}
\|\Delta_{k}\|_2^{2}\ge(1+\alpha\Lambda)^{2k}\cdot
\mathcal{O}
\bigg(
\alpha^{\frac{7}{4}}\sqrt{\log\dfrac{1}{\alpha}}E_2\sqrt{\dfrac{\sigma_{H_0}}{\sqrt{\chi\epsilon}}}
\bigg)+\Big((1+\alpha\Lambda)^{2k}-1\Big)\alpha\beta.
\end{flalign}
Since $\widehat{\kappa}_0$: ${(1+\alpha \Lambda)^{\widehat{\kappa}_0}}\leq\dfrac{1}{1-\sqrt{\alpha} \sigma_{H_0}}$, which implies the first term of (\ref{pro-ine-01}) is upper-bounded as follows,
\begin{flalign}
\label{app-Del-in-1}
(1+\alpha\Lambda)^{2k}\cdot
\mathcal{O}
\bigg(
\alpha^{\frac{7}{4}}\sqrt{\log\dfrac{1}{\alpha}}E_2\sqrt{\dfrac{\sigma_{H_0}}{\sqrt{\chi\epsilon}}}
\bigg)
\leq&
\Big(\dfrac{1}{1-\sqrt{\alpha} \sigma_{H_0}}\Big)^2\cdot
\mathcal{O}
\bigg(
\alpha^{\frac{7}{4}}\sqrt{\log\dfrac{1}{\alpha}}E_2\sqrt{\dfrac{\sigma_{H_0}}{\sqrt{\chi\epsilon}}}
\bigg)
.
\end{flalign}
The second term of (\ref{pro-ine-01}) is upper-bounded as follows: for a small $\alpha$, 
for each $k\in[0,\widehat{\kappa}_0]$,
\begin{flalign}
\nonumber
\Big((1+\alpha\Lambda)^{2k}-1\Big)\alpha\beta&=\bigg(\underbrace{\sum_{j=0}^{2k}\binom {2k}{j}(\alpha\Lambda)^{j}}_{\text{Newton's~binomial~expression~of}~(1+\alpha\Lambda)^{2k}}-1\bigg)\alpha\beta\\
\nonumber
&=\Big(2k\alpha\Lambda+\frac{2k(2k-1)(\alpha\Lambda)^2}{2}+\cdots+(\alpha\Lambda)^{2k}\Big)\alpha\beta\\
\label{app-Del-in-2}
&=\mathcal{O}\Big(2k\alpha^{2}\Lambda\beta\Big)\overset{(\ref{estimate-ka-0})}\leq\mathcal{O}\Big(\dfrac{2\alpha^{\frac{3}{2}}\Lambda\beta \sigma_{H_0}}{\sqrt{\chi\epsilon}}\Big).
\end{flalign}
Combining above (\ref{app-Del-in-1}), (\ref{app-Del-in-2}), we can bound the right equation of (\ref{pro-ine-01}) as follows,
\begin{flalign}
\nonumber
&(1+\alpha\Lambda)^{2k}\cdot
\mathcal{O}
\bigg(
\alpha^{\frac{7}{4}}\sqrt{\log\dfrac{1}{\alpha}}E_2\sqrt{\dfrac{\sigma_{H_0}}{\sqrt{\chi\epsilon}}}
\bigg)+\Big((1+\alpha\Lambda)^{2k}-1\Big)\alpha\beta\\
\leq&
\nonumber
\mathcal{O}\bigg(
\alpha^{\frac{7}{4}}\sqrt{\log\dfrac{1}{\alpha}}E_2\sqrt{\dfrac{\sigma_{H_0}}{\sqrt{\chi\epsilon}}}
\bigg)+\mathcal{O}\Big(\dfrac{2\alpha^{\frac{3}{2}}\Lambda\beta \sigma_{H_0}}{\sqrt{\chi\epsilon}}\Big)
+\Big(\dfrac{1}{1-\sqrt{\alpha} \sigma_{H_0}}\Big)^2\\
\nonumber
\leq&\mathcal{O}\Bigg(\max\bigg\{\dfrac{2\alpha^{\frac{3}{2}}\Lambda\beta \sigma_{H_0}}{\sqrt{\chi\epsilon}},
\alpha^{\frac{7}{4}}\sqrt{\log\dfrac{1}{\alpha}}E_2\sqrt{\dfrac{\sigma_{H_0}}{\sqrt{\chi\epsilon}}}
\bigg\}\Bigg)
\\
\label{condition:C-4}
\leq&C_4\sqrt{\alpha^3\log\dfrac{1}{\alpha}}.
\end{flalign}
The last (\ref{condition:C-4}) present the condition of the parameter $C_4$.

Now, we define two events $\widetilde{\mathcal{E}}_1$ and $\widetilde{\mathcal{E}}_2$ as follows,
\begin{flalign}
\nonumber
\widetilde{\mathcal{E}}_1=&\bigg\{
\|\Delta_{k}\|_2^{2}\ge(1+\alpha\Lambda)^{2k}\cdot
\mathcal{O}
\bigg(
\alpha^{\frac{7}{4}}\sqrt{\log\dfrac{1}{\alpha}}E_2\sqrt{\dfrac{\sigma_{H_0}}{\sqrt{\chi\epsilon}}}
\bigg)+\Big((1+\alpha\Lambda)^{2k}-1\Big)\alpha\beta
\bigg\}\\
\nonumber
\widetilde{\mathcal{E}}_2=&\Bigg\{ \|\Delta_{k}\|_2^{2}\ge C_4\sqrt{\alpha^3\log\dfrac{1}{\alpha}}
\Bigg\}.
\end{flalign}
The results of (\ref{pro-ine-01}) and (\ref{condition:C-4}) imply the event $\widetilde{\mathcal{E}}_1$ contains $\widetilde{\mathcal{E}}_2$, i.e., $\widetilde{\mathcal{E}}_2\subset \widetilde{\mathcal{E}}_1$. 
Then
\begin{flalign}
\nonumber
\mathbb{P}(\widetilde{\mathcal{E}}_2)&=\mathbb{P}
\bigg(
 \|\Delta_{k}\|_2^{2}\ge C_4\sqrt{\alpha^3\log\dfrac{1}{\alpha}}
\bigg)\\
\label{app-pro4-pro-in-01}
&\leq
\mathbb{P}(\widetilde{\mathcal{E}}_1)
\overset{(\textbf{a})}
=\mathbb{P}\Bigg(Z_{k}\ge \alpha\beta+\mathcal{O}\bigg(
\alpha^{\frac{7}{4}}\sqrt{\log\dfrac{1}{\alpha}}E_2\sqrt{\dfrac{\sigma_{H_0}}{\sqrt{\chi\epsilon}}}
\bigg)\Bigg)
\overset{(\ref{eq-high-p-z-k-z-0})}\leq \alpha^2,
\end{flalign}
where (\textbf{a}) of (\ref{app-pro4-pro-in-01}) holds since the event $\widetilde{\mathcal{E}}_1$ is equivalent to
$\Big\{
Z_{k}\ge \alpha\beta+\mathcal{O}\Big(
\alpha^{\frac{7}{4}}\sqrt{\log\frac{1}{\alpha}}E_2\sqrt{\frac{\sigma_{H_0}}{\sqrt{\chi\epsilon}}}
\Big)
\Big\},
$
we have provided the detail of this conclusion in (\ref{pro-ine-01}).

Furthermore, for each $k\in[0,\widehat{\kappa}_0]$, we have
\begin{flalign}
\nonumber
\overline{\mathcal{C}_{k}}\overset{(\ref{event-2-k})}\subset&
\overline{
\bigcap_{j=0}^{k}\bigg\{j\leq k:\big\|\Delta_j\big\|^2_2\leq C_4\sqrt{\alpha^3\log\dfrac{1}{\alpha}}\bigg\}
}
=
\bigcup_{j=0}^{k}\overline{\bigg\{j\leq k:\big\|\Delta_j\big\|^2_2\leq C_4\sqrt{\alpha^3\log\dfrac{1}{\alpha}}\bigg\}
}
\\
\nonumber
=&
\underbrace{\bigcup_{j=0}^{k-1}\overline{\bigg\{j\leq k:\big\|\Delta_j\big\|^2_2\leq C_4\sqrt{\alpha^3\log\dfrac{1}{\alpha}}\bigg\}
}}_{=\overline{\mathcal{C}_{k-1}}}\bigcup\bigg\{\|\Delta_k\|^2_2\ge C_4\sqrt{\alpha^3\log\dfrac{1}{\alpha}}\bigg\}
\\
\nonumber
=&\overline{\mathcal{C}_{k-1}}\cup\widetilde{\mathcal{E}}_2,
\end{flalign}
which implies for each $k\in[0,\widehat{\kappa}_0]$,
\begin{flalign}
\nonumber
\mathbb{P}(\overline{\mathcal{C}_{k}})\leq \mathbb{P}(\overline{\mathcal{C}_{k-1}})+\mathbb{P}(\widetilde{\mathcal{E}}_2)\leq \mathbb{P}(\overline{\mathcal{C}_{k-1}})+\alpha^2.
\end{flalign}
Summing the above equation from $k=1$ to $\widehat{\kappa}_0$, we have
\begin{flalign}
\nonumber
\mathbb{P}(\overline{\mathcal{C}_{\widehat{\kappa}_0}})=&\underbrace{ \mathbb{P}(\overline{\mathcal{C}_{0}})}_{=0}+
\sum_{j=1}^{\widehat{\kappa}_0}  \Big(
\mathbb{P}(\overline{\mathcal{C}_{j}})- \mathbb{P}(\overline{\mathcal{C}_{j-1}})
\Big)\\
\label{app-propbalility-of-c-k}
\leq& \widehat{\kappa}_0\alpha^2\leq\bigg\lfloor \frac{\log\Big(\frac{1}{1-\sqrt{\alpha}\sigma_{H_0}}\Big)}{\log(1+\alpha \sqrt{\chi\epsilon})}\bigg\rfloor\alpha^2=\dfrac{\sigma_{H_0}}{\sqrt{\chi\epsilon}}\alpha^{\frac{3}{2}}+o(\alpha^{\frac{3}{2}}),
\end{flalign}
where $ \mathbb{P}(\overline{\mathcal{C}_{0}})=0$ since the event $\overline{\mathcal{C}_{0}}=\big\{\Delta_0=0>C_4\sqrt{\alpha^3\log\frac{1}{\alpha}}\big\}$ can not occur.
The result of (\ref{app-propbalility-of-c-k}) shows that 
\[
\mathbb{P}\Bigg(\max \Big\{\|\Delta_k\|_2,
\|\Delta_k\|^2_2 \Big\}\ge C_4\sqrt{\alpha^3\log\dfrac{1}{\alpha}}\Bigg)\leq\dfrac{\sigma_{H_0}}{\sqrt{\chi\epsilon}}\alpha^{\frac{3}{2}}+o(\alpha^{\frac{3}{2}}).
\]
\subsubsection{
Boundedness of $\theta_{k+1}-\widehat{\theta}_{k+1}$ (\ref{itertheta}).
}
\label{bound-lemma-step-04}

From the previous result, the following happens at less than $1-\dfrac{\sigma_{H_0}}{\sqrt{\chi\epsilon}}\alpha^{\frac{3}{2}}$
\begin{flalign}
\nonumber
\big\|\theta_{k+1}-\widehat{\theta}_{k+1}\big\|_{2}\overset{(\ref{itertheta})}\leq\alpha\sum_{j=0}^{k}\|\Delta_{j}\|_2
\leq& \alpha\widehat{\kappa}_0C_4\sqrt{\alpha^3\log\frac{1}{\alpha}}
=\alpha\sqrt{\alpha^3\log\frac{1}{\alpha}}
\bigg\lfloor \dfrac{\log\Big(\frac{1}{1-\sqrt{\alpha}\sigma_{H_0}}\Big)}{\log(1+\alpha \sqrt{\chi\epsilon})}\bigg \rfloor C_4
\\
\label{app-a-gap-theta-hat-1}
=&\alpha^2\sqrt{\log\dfrac{1}{\alpha}}\dfrac{\sigma_{H_0}}{\sqrt{\chi\epsilon}}C_4+o\Big(\alpha^2\sqrt{\log\dfrac{1}{\alpha}}\Big).
\end{flalign}
This concludes the proof.
\end{proof}
\subsection{Proof of Proposition \ref{saddle-point-case}}
\label{app-proof-pro-sada-impro}
Recall Assumption \ref{ass:hessian-lipschitz}: $H(\theta)=:\nabla^{2}J(\theta)$ is $\chi$-Hessian-Lipschitz, then for any $\theta$, $\theta_0$, we have
\begin{flalign}
\label{third-app-eq}
J(\theta)\ge J(\theta_0)+\nabla J(\theta_0)^{\top}(\theta-\theta_0)+\dfrac{1}{2}(\theta-\theta_0)^{\top}H_0(\theta-\theta_0)-\dfrac{\chi}{6}\|\theta_0-\theta\|_{2}^{3}.
\end{flalign}
Let $\widehat{\phi}=\widehat{\theta}_{k+1}-\theta_{0}=\widehat{\theta}_{k+1}-\widehat{\theta}_{0}$ and $\phi=\theta_{k+1}-\widehat{\theta}_{k+1}$, after some careful calculations, we can rewrite above equation (\ref{third-app-eq}) as follows,
\begin{flalign}
\nonumber
J(\theta_{k+1})-J(\theta_{0})\ge J_1+ J_2,
\end{flalign}
where
$
J_1=\nabla J(\theta_0)^{\top}\widehat{\phi}+\dfrac{1}{2}\widehat{\phi}^{\top}H_0\widehat{\phi},
~~J_2=\nabla J(\theta_0)^{\top}\phi+
\widehat{\phi}^{\top}H_0\phi+\dfrac{1}{2}{\phi}^{\top}H_0\phi-\dfrac{\chi}{6}\|\widehat{\phi}+\phi\|_{2}^{3}.
$

\underline{Firstly, we show the lower-boundedness of the expectation of $J_1$.}

Since $J_1=\nabla J(\theta_0)^{\top}\widehat{\phi}+\dfrac{1}{2}\widehat{\phi}^{\top}H_0\widehat{\phi}$, 
we bound $\nabla J(\theta_0)^{\top}\widehat{\phi}$ at first, then bound $\dfrac{1}{2}\widehat{\phi}^{\top}H_0\widehat{\phi}$.

Recall the result of (\ref{iteration-the-j}), since $\widehat{\theta}_0=\theta_0$, then we have
\begin{flalign}
\nonumber
&\nabla J(\theta_0)^{\top}\widehat{\phi}=\nabla J(\theta_0)^{\top}(\widehat{\theta}_{k+1}-\theta_{0})\\
\label{j-1-expression}
=&\alpha\sum_{j=0}^{k}\nabla J(\widehat{\theta}_0)^{\top}(I+\alpha H_0)^{j}\nabla{J}(\widehat{\theta}_0)+\alpha\sum_{j=0}^{k}\nabla J(\widehat{\theta}_0)^{\top}(I+\alpha H_0)^{k-j}\big(\widehat{\xi}_{j}+\xi_0\big).
\end{flalign}
Furthermore, there exists an orthogonal matrix $Q\in\mathbb{R}^{p\times p}$ (i.e., $Q^{\top}Q=I$), s.t.,
$
Q^{\top}H_0 Q=
\begin{pmatrix}
\lambda_1 & 0 & \cdots  & 0\\
0 & \lambda_2 &  \cdots & 0\\
\vdots & \vdots &  \ddots & \vdots \\
0 & 0 &  \cdots & \lambda_p,
\end{pmatrix}
\Rightarrow
\nonumber
 Q^{\top}(I+\alpha H_0 )^{j}Q=
\begin{pmatrix}
(1+\alpha\lambda_1)^j & 0 & \cdots  & 0\\
0 & (1+\alpha\lambda_2)^j &  \cdots & 0\\
\vdots & \vdots &  \ddots & \vdots \\
0 & 0 & 0 & (1+\alpha\lambda_p)^j
\end{pmatrix}.
$
Let $Q=(u_1,u_2,\cdots,u_p)$, then we have $(I+\alpha H_0)^{j}=\sum_{i=1}^{p}(1+\alpha\lambda_i)^{j}u_{i}u^{\top}_{i}$.  
In fact, with some simple linear algebra, we know  $u_i$ is the unit eigenvector with respect to the eigenvalue $\lambda_i$ correspondingly.
Since then, we can rewrite the first term of (\ref{j-1-expression}) as follows,
\begin{flalign}
\label{def:J-nabla-e-i}
\nabla J(\widehat{\theta}_0)^{\top}(I+\alpha H_0)^{j}\nabla{J}(\widehat{\theta}_0)
=\sum_{i=1}^{p}(1+\alpha\lambda_i)^{j}\nabla J(\widehat{\theta}_0)^{\top}u_{i}u^{\top}_{i}\nabla J(\widehat{\theta}_0)
=\sum_{i=1}^{p}(1+\alpha\lambda_i)^{j}e^2_i,
\end{flalign}
where $e_{i}=:u^{\top}_{i}\nabla J(\widehat{\theta}_0)=u^{\top}_{i}\nabla J(\theta_0)=u^{\top}_{i}\mathbb{E}[g(\theta_0)]=\mathbb{E}[u^{\top}_{i}g(\theta_0)]$, since we set the initial $\widehat{\theta}_0=\theta_0$.

Recall  $H_0 u_p=\lambda_p u_p$, i.e., $u_p=\lambda^{-1}_{p}H_0 u_p$, and $\|u_p\|_2=1$, which implies $\lambda_{p}\leq \|H_0\|_{op}$.
Recall 
$g(\theta_k)=:g(\theta_k|\tau_k)=\sum_{t=0}^{h}\nabla\log\pi_{{\theta}}(a_{t}|s_{t})R(\tau_k)|_{\theta=\theta_k}$, then we have
\begin{flalign}
\nonumber
\mathbb{E}[e_p^2]&=\mathbb{E}\big[(u_{p}^{\top}g(\theta_0))^2\big]
=\mathbb{E}\Big[\Big(u_{p}^{\top}\sum_{t=0}^{h}\nabla\log\pi_{{\theta}}(a_{t}|s_{t})R(\tau_k)|_{\theta=\theta_0}\Big)^2\Big]\\
\nonumber
&\ge \frac{R_{\min}^{2}}{(1-\gamma)^2}\mathbb{E}
\Big[\Big(u_{p}^{\top}\sum_{t=0}^{h}\nabla_{\theta}\log\pi_{{\theta}}(a_{t}|s_{t})\Big)^2
\Big]\\
\nonumber
&\ge \frac{R_{\min}^{2}}{(1-\gamma)^2}\mathbb{E}\bigg(
\Big[\sum_{t=0}^{h}(u_{p}^{\top}\nabla_{\theta}\log\pi_{{\theta}}(a_{t}|s_{t}))^{2}\Big]\\
\label{app-c0991}
&~~~~~~~~~~~~~~~~~~+2\mathbb{E}\Big[\sum_{0\leq i<j\leq h}u_{p}^{\top}\nabla_{\theta}\log\pi_{{\theta}}(a_{i}|s_{i})u_{p}^{\top}\nabla_{\theta}\log\pi_{{\theta}}(a_{j}|s_{j})
\Big]\bigg)\\
\nonumber
&= \frac{R_{\min}^{2}}{(1-\gamma)^2}\mathbb{E}\bigg(
\Big[\sum_{t=0}^{h} u_{p}^{\top}\nabla_{\theta}\log\pi_{{\theta}}(a_{t}|s_{t})\nabla^{\top}_{\theta}\log\pi_{{\theta}}(a_{t}|s_{t})u_p\Big]\\
\nonumber
&~~~~~~~~~~~~~~~~~~+2\mathbb{E}\Big[\sum_{0\leq i<j\leq h}u_{p}^{\top}\nabla_{\theta}\log\pi_{{\theta}}(a_{i}|s_{i})u_{p}^{\top}\nabla_{\theta}\log\pi_{{\theta}}(a_{j}|s_{j})
\Big]\bigg)\\
\nonumber
&\ge\frac{R_{\min}^{2}h\omega}{(1-\gamma)^2}
+2\frac{R_{\min}^{2}}{(1-\gamma)^2}\mathbb{E}\Big[\sum_{0\leq i<j\leq h}(\lambda^{-1}_{p}H_0 u_p)^{\top}\nabla_{\theta}\log\pi_{{\theta}}(a_{i}|s_{i}){(\lambda^{-1}_{p}H_0 u_p)}^{\top}\nabla_{\theta}\log\pi_{{\theta}}(a_{j}|s_{j})
\Big]
\\
\nonumber
&\ge\frac{R_{\min}^{2}h\omega}{(1-\gamma)^2}
+\frac{2R_{\min}^{2}}{(1-\gamma)^2\|H_0\|^{2}_{op}}\mathbb{E}\Big[\sum_{0\leq i<j\leq h}(H_0 u_p)^{\top}\nabla_{\theta}\log\pi_{{\theta}}(a_{i}|s_{i}){(H_0 u_p)}^{\top}\nabla_{\theta}\log\pi_{{\theta}}(a_{j}|s_{j})
\Big]\\
\label{app-c0992}
&=\frac{R_{\min}^{2}h\omega}{(1-\gamma)^2}
+\frac{2R_{\min}^{2}\lambda^{2}_{p}}{(1-\gamma)^2\|H_0\|^{2}_{op}}\mathbb{E}\Big[\sum_{0\leq i<j\leq h}\nabla^{\top}_{\theta}\log\pi_{{\theta}}(a_{i}|s_{i})\nabla_{\theta}\log\pi_{{\theta}}(a_{j}|s_{j})
\Big],
\end{flalign}
where Eq.(\ref{app-c0991}) holds since a simple fact: 
$(\sum_{t=0}^{h}x_t)^{2}= \sum_{t=0}^{h}x^{2}_{t}+2\sum_{0\leq i<j\leq h}x_{i}x_{j}$.
\begin{discussion}
\label{app-discussion}
Now, we discuss the lower boundedness of (\ref{app-c0992}), we introduce a notation $c_0$ as flows,
\[c_0=:\mathbb{E}\Big[\sum_{0\leq i<j\leq h}\nabla^{\top}_{\theta}\log\pi_{{\theta}}(a_{i}|s_{i})\nabla_{\theta}\log\pi_{{\theta}}(a_{j}|s_{j}),\Big]\]
if $c_0\ge0$, the result of (\ref{app-c0992}) shows that 
\[\mathbb{E}[e_p^2]=\mathbb{E}\big[(u_{p}^{\top}g(\theta_0))^2\big]=\mathbb{E}[\langle g(\tau|\theta),u_p \rangle^2]\ge\frac{R_{\min}^{2}h\omega}{(1-\gamma)^2};\]
if $c_0<0$, since $\lambda_{p}\ge\sqrt{\chi\epsilon}$, if we require (\ref{app-c0992}) keep positive, then we have a small $\epsilon$ s.t.,
\[
\epsilon\leq\dfrac{h\omega\|H_0\|_{op}^{2}}{\chi|c_0|}.\]
Then, for a small enough $\epsilon$, we have the lower boundedness of the term $\mathbb{E}[\langle g(\tau|\theta),u_p \rangle^2]$ as follows,
\begin{flalign}
\label{app-e-it-constant}
\mathbb{E}[\langle g(\tau|\theta),u_p \rangle^2]\ge
\min\Big\{
\frac{R_{\min}^{2}h\omega}{(1-\gamma)^2}, \frac{R_{\min}^{2}h\omega}{(1-\gamma)^2}
+\frac{2R_{\min}^{2}\lambda^{2}_{p}}{(1-\gamma)^2\|H_0\|^{2}_{op}}c_0
\Big\}=:\iota^2.
\end{flalign}
\end{discussion}

Now, we consider $\nabla J(\theta_0)^{\top}\widehat{\phi}$, for an enough small $0<\alpha<\max\{\frac{1}{|\lambda_1|},\frac{1}{\lambda_p}\}$, we have
\begin{flalign}
\nonumber
\mathbb{E}[\nabla J(\theta_0)^{\top}\widehat{\phi}]\overset{(\ref{def:J-nabla-e-i})}=&\alpha\sum_{j=0}^{T}\sum_{i=1}^{p}(1+\alpha\lambda_i)^j \mathbb{E}[e_i^{2}]+\alpha\sum_{j=0}^{T}\nabla J(\widehat{\theta}_0)^{\top}(I+\alpha H_0)^{k-j}{\mathbb{E}\big[\big(\widehat{\xi}_{j}+\xi_0\big)\big]}\\
\label{app-lower-bound-j-1}
\ge&\alpha\sum_{j=0}^{T}(1+\alpha\lambda_p)^j \mathbb{E}[e_p^{2}]\ge \alpha(1+\alpha\sqrt{\chi\epsilon})\iota^2.
\end{flalign}
For the term  $\dfrac{1}{2}\widehat{\phi}^{\top}H_0\widehat{\phi}$, we have,
\begin{flalign}
 \dfrac{1}{2}|\widehat{\phi}^{\top}H_0\widehat{\phi}|\leq \dfrac{1}{2}\Lambda\|\widehat{\phi}\|_2^2\leq\Lambda\alpha^2C_2^2
\end{flalign}

\underline{Secondly, we bound the expectation of $J_2$.}

We define an event $\mathcal{D}_{k}$ as follows,
\[
\mathcal{D}_{k}=\bigcap_{j=0}^{k}\Big\{
\|\theta_{j}-\widehat{\theta}_{j}\|_{2}\leq \alpha^2\sqrt{\log\dfrac{1}{\alpha}}\dfrac{\sigma_{H_0}}{\sqrt{\chi\epsilon}}C_4,~\text{and}~
\|\widehat{\theta}_{j}-\widehat{\theta}_{0}\big\|_2\leq\alpha C_2
\Big\},
\]
from the result of Lemma \ref{app-key-lem} and Lemma \ref{app-key-lem-second-1}, for each $k\in[0,\widehat{\kappa}_0]$, we have
\[
\mathbb{P}(\mathcal{D}_{k})\ge1-\min\Big\{\delta,
\dfrac{\sigma_{H_0}}{\sqrt{\chi\epsilon}}\alpha^{\frac{3}{2}}
\Big\}.
\]
Recall
$
J_2=\nabla J(\theta_0)^{\top}\phi+
\widehat{\phi}^{\top}H_0\phi+\dfrac{1}{2}{\phi}^{\top}H_0\phi-\dfrac{\chi}{6}\|\widehat{\phi}+\phi\|_{2}^{3}
$, we need to bound all the terms of $J_2$ with high probability,
In fact,
\begin{flalign}
\label{j-2-1}
\big|\nabla J(\theta_0)^{\top}\phi\big|\leq \epsilon\big\|\theta_{T}-\widehat{\theta}_{T}\big\|_2\overset{(\ref{app-a-gap-theta-hat-1})}\leq
\alpha^2\sqrt{\log\dfrac{1}{\alpha}}\sqrt{\dfrac{\epsilon}{\chi}}\sigma_{H_0}C_4+o\bigg(\sqrt{\dfrac{\epsilon}{\chi}}\alpha^2\sqrt{\log\dfrac{1}{\alpha}}
\bigg)
\end{flalign}
\begin{flalign}
\nonumber
\big|\widehat{\phi}^{\top}H_0\phi\big|
\leq
\big\|\widehat{\phi}\big\|_2\big\|H_0\big\|_{op}\big\|\phi\big\|_2\overset{(\ref{high-p-theta-gap}),(\ref{app-a-gap-theta-hat-1})}\leq&\Lambda\alpha\sqrt{\alpha^3\log\frac{1}{\alpha}}
\bigg \lfloor\dfrac{\log\Big(\frac{1}{1-\sqrt{\alpha}\sigma_{H_0}}\Big)}{\log(1+\alpha \sqrt{\chi\epsilon})}\bigg \rfloor C_4\alpha C_2\\
\label{j-2-2}
=&\alpha^3\sqrt{\log\dfrac{1}{\alpha}}\dfrac{\sigma_{H_0}}{\sqrt{\chi\epsilon}}\Lambda C_2C_4 +
o
\Big(
\alpha^3\sqrt{\log\dfrac{1}{\alpha}}\cdot\dfrac{1}{\sqrt{\chi\epsilon}}
\Big),
\end{flalign}
\begin{flalign}
\label{j-2-3}
\big|\dfrac{1}{2}{\phi}^{\top}H_0\phi\big |\leq\frac{1}{2}\big\|{\phi}\big \|_2\big\|H_0\big\|_{op}\big\|\phi\big\|_2\overset{(\ref{app-a-gap-theta-hat-1})}\leq
\dfrac{1}{2\chi\epsilon}
\Lambda^2 \alpha^4\log\dfrac{1}{\alpha}\sigma^2_{H_0}C^2_4+o\bigg({\dfrac{1}{\chi\epsilon}}\alpha^4{\log\dfrac{1}{\alpha}}\bigg)
\end{flalign}
\begin{flalign}
\label{j-2-4}
\dfrac{\chi}{6}\|\widehat{\phi}+\phi\|_{2}^{3}\leq\dfrac{\chi}{6}
\Bigg(
\sqrt{\alpha^3\log\frac{1}{\alpha}}
\bigg\lfloor \dfrac{\log\Big(\frac{1}{1-\sqrt{\alpha}\sigma_{H_0}}\Big)}{\log(1+\alpha \sqrt{\chi\epsilon})}\bigg \rfloor C_4+\alpha C_2
\Bigg)^3=\dfrac{\chi}{6}\alpha^3C_4^3 +o(\alpha^3).
\end{flalign}
From the results of (\ref{j-2-1})-(\ref{j-2-4}), for a enough small $\alpha$, we have
\[
J_2\bm{1}_{\mathcal{D}_k}
\leq
\alpha^2\sqrt{\log\frac{1}{\alpha}}\sqrt{\frac{\epsilon}{\chi}}\sigma_{H_0}C_4+o\Big(\sqrt{\frac{\epsilon}{\chi}}\alpha^2\sqrt{\log\frac{1}{\alpha}}
\Big),
\]
which implies
\begin{flalign}
\label{app-ex-J2-1}
\mathbb{E}[J_2 \bm{1}_{\mathcal{D}_k}]\leq
\alpha^2\sqrt{\log\dfrac{1}{\alpha}}\sqrt{\dfrac{\epsilon}{\chi}}\sigma_{H_0}C_4+o\bigg(\sqrt{\dfrac{\epsilon}{\chi}}\alpha^2\sqrt{\log\dfrac{1}{\alpha}}
\bigg).
\end{flalign}

Furthermore, from the result of (\ref{iteration-the-j}), (\ref{itertheta}),
it is easy to show $\|\theta_{k}-\widehat{\theta}_{k}\|_{2}\leq\mathcal{O}(1)$,
$\|\theta_{k}-\widehat{\theta}_{0}\|_{2}\leq\mathcal{O}(1)$, which implies
\begin{flalign}
\label{app-ex-J2-2}
\mathbb{E}[J_2 \bm{1}_{\overline{\mathcal{D}_k}}]\leq\min\Big\{\delta,
\dfrac{\sigma_{H_0}}{\sqrt{\chi\epsilon}}\alpha^{\frac{3}{2}}
\Big\}.
\end{flalign}
By the results of (\ref{app-ex-J2-1}) and (\ref{app-ex-J2-2}), the lower boundedness of $\mathbb{E}[J_2]$ reaches
\[
\mathbb{E}[J_2]\ge
-\alpha^2\sqrt{\log\dfrac{1}{\alpha}}\sqrt{\dfrac{\epsilon}{\chi}}\sigma_{H_0}C_4-\min\Big\{\delta,
\dfrac{\sigma_{H_0}}{\sqrt{\chi\epsilon}}\alpha^{\frac{3}{2}}
\Big\}=\mathcal{O}
\Big(
-\min\Big\{\delta,
\dfrac{\sigma_{H_0}}{\sqrt{\chi\epsilon}}\alpha^{\frac{3}{2}}
\Big\}
\Big)
.
\]
Combining above result with (\ref{app-lower-bound-j-1}), we have
\begin{flalign}
\nonumber
\mathbb{E}[J(\theta_{T})]-J(\theta_{0})&\ge\mathbb{E}[J_1]+\mathbb{E}[J_2]\\
\nonumber
&\ge \alpha(1+\alpha\sqrt{\chi\epsilon})\iota^2-\Lambda\alpha^2C_2^2
-
\mathcal{O}
\Big(
\min\Big\{\delta,
\dfrac{\sigma_{H_0}}{\sqrt{\chi\epsilon}}\alpha^{\frac{3}{2}}
\Big\}
\Big)\\
\label{last-eq}
&\ge\alpha^{2} \iota^2\sqrt{\chi\epsilon},
\end{flalign}
the last (\ref{last-eq}) holds since we chose a proper step-size satifies
\begin{flalign}
\label{step-size-con}
\alpha\iota^2-\Lambda\alpha^2C_2^2
-
\mathcal{O}
\Big(
\min\Big\{\delta,
\frac{\sigma_{H_0}}{\sqrt{\chi\epsilon}}\alpha^{\frac{3}{2}}
\Big\}
\Big)>0,
\end{flalign}
for a small $\alpha$, Eq.(\ref{step-size-con}) always exiits.
Concretely, if $\delta>\frac{\sigma_{H_0}}{\sqrt{\chi\epsilon}}\alpha^{\frac{3}{2}}$, then we can chose step as follows,
\begin{flalign}
\label{app-condi-001}
\sqrt{\alpha}<\dfrac{\frac{\sigma_{H_0}}{\sqrt{\chi\epsilon}}
+
\sqrt{
(\frac{\sigma_{H_0}}{\sqrt{\chi\epsilon}})^2+4\Lambda C_2^{2}\iota^2
}
}{a\Lambda C_2^{2}};
\end{flalign}
otherwise, if  $\delta\leq\frac{\sigma_{H_0}}{\sqrt{\chi\epsilon}}\alpha^{\frac{3}{2}}$, we can chose step as follows,
\[
\alpha\iota^2-\Lambda\alpha^2C_2^2
-\delta\ge\alpha\iota^2-\Lambda\alpha^2C_2^2
-\frac{\sigma_{H_0}}{\sqrt{\chi\epsilon}}\alpha^{\frac{3}{2}}>0,
\]
which implies same condition as Eq.(\ref{app-condi-001}).